\documentclass{article}

\usepackage[preprint]{neurips_2022}

\usepackage[utf8]{inputenc} 
\usepackage[T1]{fontenc}    
\usepackage{hyperref}       
\usepackage{url}            
\usepackage{booktabs}       
\usepackage{amsfonts}       
\usepackage{nicefrac}       
\usepackage{microtype}      
\usepackage{xcolor}         

\usepackage{enumitem}       
\usepackage{graphicx}
\usepackage{subfigure}
\usepackage{float}
\usepackage{siunitx}
\usepackage{caption}

\title{Driving in Real Life with Inverse Reinforcement Learning}

\author{%
  Tung Phan-Minh \\
   \And
   Forbes Howington \\
   \And
   Ting-Sheng Chu \\
   \And
   Sang Uk Lee \\
   \And
   Momchil S.~Tomov \\
   \And
   Nanxiang Li \\
   \And
   Caglayan Dicle \\
   \And
   Samuel Findler \\
   \And
   Francisco Suarez-Ruiz \\
   \And
   Robert Beaudoin \\
   \And
   Bo Yang \\
   \And
   Sammy Omari \\
   \And
   Eric M.~Wolff \\
  Motional \\
  \texttt{\{tung.phan, forbes.howington, ting-sheng.chu, eric.wolff\}@motional.com}
}

\begin{document}

\maketitle

\begin{abstract}
In this paper, we introduce the first learning-based planner to drive a car in dense, urban traffic using Inverse Reinforcement Learning (IRL).
Our planner, DriveIRL, generates a diverse set of trajectory proposals, filters these trajectories with a lightweight and interpretable safety filter, and then uses a learned model to score each remaining trajectory.
The best trajectory is then tracked by the low-level controller of our self-driving vehicle.
We train our trajectory scoring model on a $500$+ hour real-world dataset of expert driving demonstrations in Las Vegas within the maximum entropy IRL framework.
DriveIRL's benefits include: a simple design due to only learning the trajectory scoring function, relatively interpretable features, and strong real-world performance.
We validated DriveIRL on the Las Vegas Strip and demonstrated fully autonomous driving in heavy traffic, including scenarios involving cut-ins, abrupt braking by the lead vehicle, and hotel pickup/dropoff zones.
Our dataset will be made public to help further research in this area.
\end{abstract}


\section{Introduction}
Self-driving cars have been the focus of significant research and development over the past decade.
Some companies are tantalizingly close to deploying commercial self-driving taxi services that would make urban transportation cheaper and safer.
Progress in self-driving cars has been largely driven by new datasets~\citep{nuscenes2019,sun2020waymo-open-dataset,chang2019argoverse,geiger2012kitti} that helped fuel dramatic improvements in machine learning approaches to object detection ~\citep{zhou2018voxelnet,lang2019pointpillars} and motion forecasting ~\citep{cui2019multimodal,chai2019multipath,phan2020covernet}.
However, the critical motion planning and decision-making algorithms that ultimately determine driving behavior have yet to see similar benefits from machine learning.

Classical planning and decision-making algorithms for self-driving cars rely heavily on hand-engineered components~\citep{paden2016survey-planning-self-driving}.
Developers will typically hand-tune the scoring function that determines which behaviors are desirable.
Manually adjusting features and weights can be a painstaking process -- improving performance in one area often causes unintended regressions elsewhere.
Our planner avoids the need to manually craft detailed features or tune weights by learning these components from expert demonstrations using a maximum entropy IRL framework.

Our DriveIRL system works by \emph{generating}, \emph{checking}, and \emph{scoring} trajectories for our vehicle.
We use simple and interpretable modules to do the relatively easy tasks of generating a diverse set of ego trajectories and checking that they are safe.
Careful construction of the proposed trajectories ensures that they a) are dynamically feasible, b) follow the route, c) satisfy assumptions from the vehicle controller, and d) are diverse.
We then apply a lightweight \emph{safety filter} that ensures that each trajectory satisfies a recursive safety guarantee: if we execute the first part of the trajectory, there exists a safe continuation of that trajectory which avoids collision.
The learning component of our model focuses entirely on appropriately scoring these trajectories based on the expert demonstrations.
Our design directs the model capacity towards hard-to-specify nuances in behavior (e.g., speed profiles, clearances) instead of also creating ``nice'' trajectories and avoiding obviously unsafe behavior. 

DriveIRL achieves strong real-world driving performance on the Las Vegas Strip.
The Strip is a major thoroughfare in Las Vegas which connects many of the major hotels and casinos.
Challenges include dense traffic, aggressive cut-ins, erratic drivers, and busy passenger pickup/dropoff zones near the hotels.
We deployed DriveIRL on a self-driving car and drove fully autonomously on the Strip in these scenarios, showing the practical utility of our approach.

Our main contributions towards learning-based planning for self-driving cars are:
\begin{itemize}[noitemsep,nolistsep]
    \item The first learning-based planner to drive a car in dense, urban traffic using IRL.
    \item A simple yet powerful modeling framework that focuses learning on the aspect of driving that is most challenging to specify.
    \item Detailed evaluation of our planner on a real-world dataset that we will make public.
\end{itemize}

\begin{figure}[t]
    \centering
    \includegraphics[width=\textwidth]{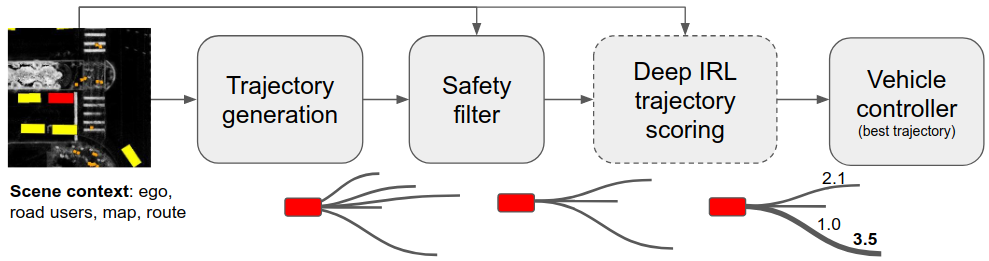}
    \vspace{-5mm}
    \caption{DriveIRL architecture. The learned scoring component is indicated with a dotted boundary.}
    \label{fig:model}
\end{figure}


\section{Related work}
\noindent\textbf{Classical planning}:
Traditional approaches formulate the planning problem as search over an appropriately constructed graph ~\citep[e.g., A*, RRT*, PRM*;][]{lavalle2006planning-algos, paden2016survey-planning-self-driving} or trajectory optimization~\citep{paden2016survey-planning-self-driving}.
These methods often have strong theoretical guarantees on convergence to an optimal solution and are relatively easy to interpret.
However, the cost function that defines desired behavior is often hand-engineered, and in practice requires painstaking tuning to produce appropriate behavior.

\noindent\textbf{Imitation learning (IL)}:
IL methods attempt to directly imitate the actions of an expert driver.
It has seen applications to self-driving cars starting with the pioneering work of ALVINN~\citep{pomerleau1988alvinn}.
More recently, an end-to-end driving policy was learned from camera images to control actions for lane keeping~\citep{bojarski2016nvidia-end-to-end}.

A fundamental issue with IL is that there is a distribution shift from training to deployment, as small errors lead to the model operating outside of its training data, which then leads to larger errors.
ChauffeurNet~\citep{bansal2019chauffeurnet} uses behavioral cloning with extensive data augmentation to mitigate the distribution shift issue, and UrbanDriver~\citep{scheel2021urban-driver} uses an offline policy gradient method with closed-loop rollouts during training to automatically create appropriate data augmentation.
TrafficSim~\citep{suo2021trafficsim} similarly uses closed-loop training, but with a focus on creating traffic simulation.
While data augmentation improves our performance, it is not as critical since our trajectory generation mechanism pulls the car towards the lane center, reducing divergence.

Closely related is work by~\citet{zeng2019neural-planner-costmap} which learns a costmap over the environment to score a set of procedurally generated trajectories.
Our approach improves on trajectory generation since we ensure map compliance, as well as on scoring flexibility since we do not impose the assumption of an additive costmap.
Furthermore, we demonstrate our model on a vehicle in dense urban traffic.

Another similar approach is that of~\citet{vitelli2021safety-net}, where a hybrid model with a learned planner and an interpretable fallback layer drive in San Francisco.
Our IRL-based model is simpler and less reliant on a fallback layer.
Furthermore, the recursive check of our safety filter is less conservative.

\noindent\textbf{Reinforcement learning (RL)}:
RL approaches learn a driving policy by optimizing a reward function.
The standard approach requires a simulator~\citep[e.g., CARLA;][]{dosovitskiy2017carla} to update the environment that the driving policy interacts with.
There have been a variety of approaches that have shown strong performance in simulation~\citep{chen2020learning-by-cheating,chen2021world-on-rails}.

Real-world applications of RL for self-driving cars have been rarer, likely due to the difficulty in modeling the environment and specifying the reward function.
An early notable example is~\citet{riedmiller2007drive-in-20-min}, where they learn a steering policy for a real car.
More recently, lane following was demonstrated using deep RL~\citep{kendall2019drive-in-a-day}.
This approach controlled both speed and steering on a real car.
We contrast the rural driving evaluations above with our experiments in busy Las Vegas.

\noindent\textbf{Inverse reinforcement learning (IRL)}:
IRL methods assume that the expert is optimizing an unknown cost function, which is learned from expert demonstrations.
An early application of IRL to self-driving cars was for parking lot navigation~\citep{abbeel2004apprenticeship-car}.
The method learned multiple different driving styles from a handful of demonstrations.
However, the environment was static and the formulation assumes a linear combination of carefully handcrafted features.

Our approach is based on the popular maximum entropy formulation of IRL~\citep{ziebart2008max-entropy-irl}, which avoids ambiguities inherent in matching feature expectations.
The maximum entropy IRL approach was extended to deep learning in~\citet{wulfmeier2015deep-max-entropy-irl}, which avoided the need for laborious hand-engineering of features, and applied to simple benchmarks.
The work of \citet{huang2021irl} is the most related to our approach, but their model only learns a handful of feature weights while still assuming a linear combination of handcrafted features.
In addition, their method is validated on a highway driving dataset and not on a real vehicle.


\section{Inverse Reinforcement Learning Planner}
In this section, we describe our Inverse Reinforcement Learning (IRL) Planner as shown in Fig.~\ref{fig:model}.
Our system consists of three main stages: trajectory generation (Sec.~\ref{subsec:traj-gen}), safety filtering (Sec.~\ref{subsec:safety-filter}), and trajectory scoring (Sec.~\ref{subsec:traj-score}).
We rely on simple and reliable hand-engineered modules for trajectory generation and safety, and focus on learning how to score trajectories.

\subsection{Input and output}
\noindent\textbf{Input}:
We encode the environment (or scene) around our self-driving car using a mid-level representation.
We assume that the ego is localized within a high-definition map and that objects are detected and tracked by a Perception system.
Other road users (e.g., cars, bicyclists, and pedestrians) are represented by object type, an oriented bounding box, and speed.
The high-definition map  provides lane center-lines, road boundaries, traffic light locations, pedestrian crosswalks, speed limits, and other semantic information.
We also provide a route, which indicates the lanes that the ego should traverse to make progress towards its goal.

We refer to the \emph{scene context} at a given timestamp as a) the ego dynamic state $\mathcal{S}$ (speed, acceleration, steering), b) the other road users $\mathcal{U}$ (type, oriented bounding box, speed) , c) the map $\mathcal{M}$, and d) the ego's desired route $\mathcal{R}$.
The model receives the \emph{scene context} at the current timestamp as well as a specified number of previous timestamps (e.g., the past 1 second) as history $\mathcal{H}$.

\noindent\textbf{Output}:
Our planner generates multiple ego trajectories and scores each one according to how closely it matches what an expert would do given the scene context.
A trajectory is a discrete sequence of future states of the ego, where we assume that there is a fixed timestep between all states.
Let $s_t = (x, y, \theta, v)$ represent a state at time $t$, with position ($x$, $y$), heading $\theta$, and speed $v$.
All values are with respect to the ego's geometric center in a fixed coordinate frame.
The trajectory $\tau = [s_1, \ldots, s_T]$, where $T$ is the planned time horizon, that is ranked the best among a set of trajectories $\mathcal{T}$, is used as a reference for the vehicle's tracking and actuator controller.

\subsection{Trajectory generation}
\label{subsec:traj-gen}
The trajectory generation module uses the scene context to synthesize a diverse set of possible future motions for the ego.
Important considerations for the ego's trajectory are that it a) is dynamically feasible, and b) satisfies all requirements of the low-level tracking and actuator control (i.e., levels of continuity, minimum turn radius, minimum acceleration from a stop).
Secondary considerations are that the trajectory is compliant with the map (e.g., it stays on the road).
While these considerations do not preclude using a learned trajectory generation module, we found that a hand-engineered trajectory generator best satisfied the considerations above.

\begin{figure*}
    \centering
    \subfigure
    {
        \includegraphics[width=.17\columnwidth]{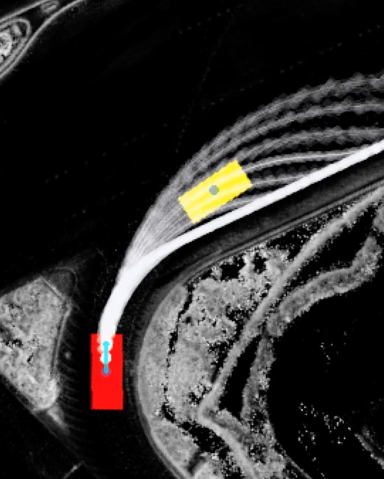}
    }
    \subfigure
    {
        \includegraphics[width=.17\columnwidth]{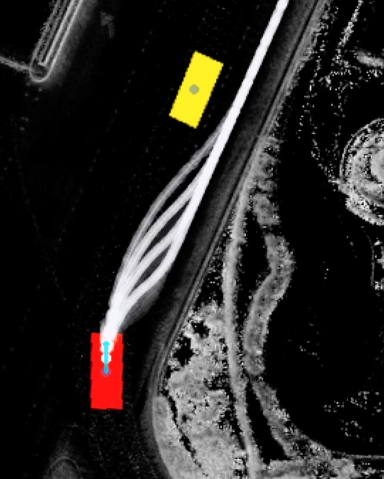}
    }
    \subfigure
    {
        \includegraphics[width=.17\columnwidth]{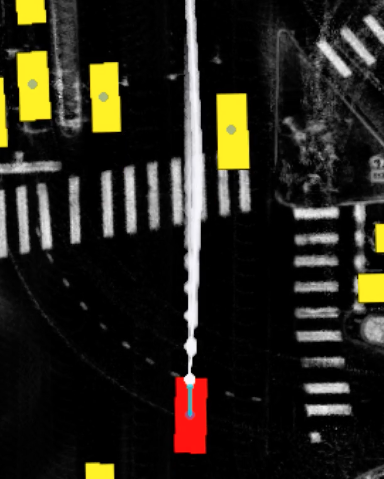}
    }
    \subfigure
    {
        \includegraphics[width=.17\columnwidth]{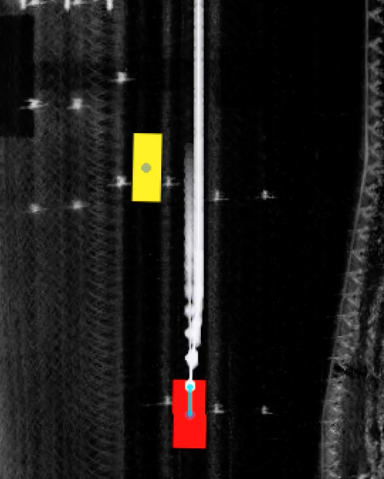}
    }
    \caption{Proposed trajectories for the ego (red rectangle). Each trajectory is shown in translucent white dot-line. Overlap is due to multiple acceleration profiles. All trajectories return to the route.}
    \label{fig:trajecotry_set}
\end{figure*}

The trajectory generator uses i) the current ego state $\mathcal{S}$, ii) the route $\mathcal{R}$, and iii) the map $\mathcal{M}$ to create a diverse set of ego trajectories $\mathcal{T}$, namely $\mathcal{(S, R, M)}\mapsto \mathcal{T}$.
The generator integrates a desired acceleration profile along the route ahead of the ego.
In our experiments, we specified a range of constant acceleration profiles ranging from a hard brake ($\SI{-5.0}{\meter \per \second^2}$) to a moderate acceleration ($\SI{1.5}{\meter \per \second^2}$).
As the ego will not always be on the lane center-line (due to vehicle controller tracking errors), we smoothly connect the initial ego pose with the route with Dubins paths~\cite{lavalle2006planning-algos} where turning radii are a fixed set of parameters.
In a typical scene, the trajectory generator usually creates $50$-$150$ trajectories depending on the ego state and route.
Some examples are shown in the Fig \ref{fig:trajecotry_set}.
Appendix~\ref{app:trajectory-quality} provides results to validate that the generated trajectories are of good quality. 

\subsection{Safety filter}
\label{subsec:safety-filter}

\begin{figure}[h]
    \centering
    \includegraphics[width=\textwidth]{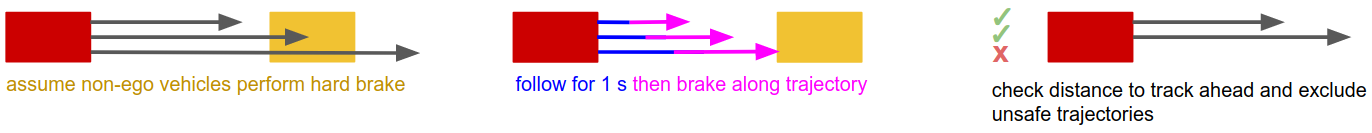}
    \vspace{-5mm}
    \caption{Safety filter. \textit{Left}: toy scenario with three trajectories (ego is in red, vehicle ahead is in yellow). \textit{Middle}: modified trajectories. \textit{Right}: unsafe trajectories excluded from trajectory set.}
    \label{fig:safety-filter}
\end{figure}

Before scoring candidate trajectories, we apply an interpretable safety filter (Fig~\ref{fig:safety-filter}) to guarantee basic safety (i.e., no collisions).
It consists of:

\begin{itemize}[noitemsep,nolistsep]
\item a set of world assumptions used to predict the behavior of the non-ego road users,
\item a set of trajectory modifiers which are applied to the ego trajectory, and
\item a set of safety checks which the modified ego trajectory needs to pass.
\end{itemize}

For a candidate trajectory to be considered safe, it must pass all safety checks, under the given trajectory modifications and assumptions about the other road users.
See~\ref{app:safety-filter-details} for details.

Our safety filter is similar in spirit to the fallback layer proposed by~\citet{vitelli2021safety-net}, except that 1) it directly filters the proposed trajectories, rather than projecting the output trajectory to an ad-hoc trajectory set, and 2) the trajectory modifier effectively implements a recursive safety guarantee with minimal assumptions and checks, without compromising comfort.

\subsection{Trajectory scoring with maximum entropy IRL}
\label{subsec:traj-score}
Appropriately scoring trajectories is the core challenge of our planning approach.
This difficulty is because proper driving behavior is heavily influenced by the environment around us, including other road user behavior and goals, of which we only have a partial understanding.

Trajectories are scored by a deep neural network trained with a maximum entropy IRL loss~\citep{ziebart2008max-entropy-irl}.
We use expert demonstrations collected from a skilled human driving our vehicle.
The loss favors trajectories that most closely match the expert demonstration $\tau^{\star}$ in feature space.
In particular, let $r(\tau)$ represent the reward of the trajectory $\tau \in \mathcal{T}$, the probability of a trajectory $\tau^{\ast}$ being selected according to the maximum entropy principle is
$P(\tau^{\star}) = \frac{\exp{r(\tau^{\star})}}{\sum\limits_{\tau} \exp{r(\tau)}}$.

The negative log-likelihood loss (NLL) on a dataset $D$ is defined as
$\ell(D) = -\sum\limits_{d \in D}{\log{P(\tau^{\star}(d))}}$ where $\tau^{\star}(d)$ is the demonstrated trajectory on the token $d \in D$.
To address data imbalance issues, we augment NLL with focal loss~\citep{lin2017focal} (with a $\gamma$ of $2.0$)

\begin{equation}
    \label{eq:loss-function}
    \ell(D) = -\sum\limits_{d \in D}{(1-P(\tau^{\star}(d)))^{\gamma}\log{P(\tau^{\star}(d))}} .
\end{equation}

\noindent\textbf{Features}:
We compute features for each proposed trajectory to use as inputs to our neural network.
These features can be based on any combination of a proposed trajectory $\tau$, ego state
$\mathcal{S}$, other road users $\mathcal{U}$, the map $\mathcal{M}$, route $\mathcal{R}$, and
history $\mathcal{H}$, meaning that $F_i: (\tau, \mathcal{S}, \mathcal{U}, \mathcal{M}, \mathcal{R},
\mathcal{H} ) \mapsto f_i \in \mathbb{R}^{k_{i}}$, where $F_i$ is the feature extraction function corresponding to feature $i$ and $k_i$ is its dimension.

\begin{itemize}[noitemsep,nolistsep]
    \item \textit{Time-to-collision (TTC)}: the minimum number of seconds before the ego would collide with another road user in the (predicted) future. Evaluated at multiple points.
    \item \textit{ACCInfo}: the ego speed, the distance to the road user ahead, the speed of the road user ahead, and the relative speed of the road user ahead. Evaluated at multiple points.
    \item \textit{MaxJerk}: the maximum jerk ($\SI{}{\meter \per \second^3}$) along the trajectory. 
    \item \textit{MaxLateralAccel}: the max lateral acceleration ($\SI{}{\meter \per \second^2}$) along the trajectory.
    \item \textit{PastCoupling}: concatenation of the future trajectory and the one second of past ego poses to model learn to maintain the coherence between the past, present, and future trajectories.
    \item \textit{SpeedLimit}: how closely the trajectory obeys the speed limit. Evaluated at multiple points.
\end{itemize}

More implementation details can be found in the Appendix~\ref{app:features}.

\noindent\textbf{Motion prediction}:
Some of the features computed for each proposed trajectory require an estimate of where other road users will be in the future, such as \textit{Time-to-collision (TTC)} and \textit{ACCInfo}.
We use an Intelligent Driver Model (IDM)~\citep{treiber2000congested-idm} as our prediction model for other cars, with a conservative acceleration value to avoid assuming that stationary vehicles will speed up.
We use a constant velocity model for pedestrians and for vehicles without a nearby lane. 

\noindent\textbf{Model architecture}:
To score a trajectory, we adopt an architecture in which the extracted features are processed separately before interacting with one another through a masked self-attention mechanism.

\begin{figure}[h]
    \centering
    \includegraphics[width=\textwidth]{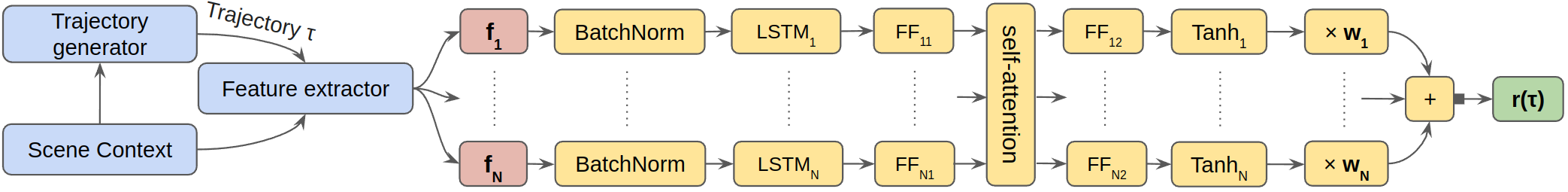}
    \vspace{-2mm}
    \caption{Detailed trajectory scoring architecture.}
\end{figure}

Under this architecture, each input feature $f_i$, as a temporal sequence of related vehicle-environment interaction data, is first normalized through an application of a BatchNorm1D layer before being fed to an LSTM module with one layer and a hidden size of $20$.
The output of the LSTM becomes the input to a feed-forward module and then a self-attention mechanism with two heads and an embedding dimension of $120$. Here we employ zero-masking of the queries to encode position.
By taking into account other features through self-attention, the model produces for each feature a
``corrected'' output embedding that can now be passed to a feed-forward network which converts it
into a scalar and then a $\tanh$ activation to produce a feature score $y_i$.
The final score for the trajectory is the sum of these feature scores after they are multiplied by the corresponding learnable feature weight parameters $w_i$: $r(\tau) = \sum\limits_{i} w_i y_i$.
In total, our base (best) model has $\approx88,700$ trainable parameters.


\section{Experiments}
\label{sec:experiments}
The proposed inverse reinforcement leaning planner was evaluated on a large-scale dataset and the
results are presented in the following.
The dataset we used is described in Sec \ref{subsec:dataset}.
The metrics for comparison is explained in Sec \ref{subsec:metrics}.
Various model ablation studies and the comparison with baseline are shown in Sec \ref{subsec:model-ablations} and \ref{subsec:baseline}.
We demonstrated both simulation results and the real-world driving tests in Sec \ref{subsec:sim-result} and \ref{subsec:realworld-driving} respectively.

\subsection{Dataset}
\label{subsec:dataset}
We created a self-driving car dataset that captures real-world urban driving in the center of Las Vegas.
Our dataset is a part of the nuPlan~\citep{caesar2021nuplan} dataset that will be made public.
It includes object annotations and high-definition maps.
Vehicles, pedestrians, and bicyclists are automatically annotated using an offline perception system (similar in spirit to \citet{qi2021offboard-auto-label}) and viewed as ground truth.
We performed filtering and extracted 182,032 scenarios, each 11 seconds in duration (1 second past, 10 seconds future), for a total of approximately 556 hours.
Our main interest was to learn good adaptive cruise control (ACC) behavior.
Thus, we filtered out scenarios where the ego made lane changes or deviated far from the lane.
After filtering, we performed a 3:1:1 split for train, val, and test sets.
Tab.~\ref{tab:dataset-distribution} shows a detailed distribution of our dataset by scenario tags.
The tags in the table are not mutually exclusive and a scenario can belong to multiple tags.
More detailed definitions for the scenario tags are in Appendix~\ref{app:scenario_tags}.

\begin{table}
  \caption{A detailed distribution of our dataset ($182,032$ total $11$-second scenarios).}
  \label{tab:dataset-distribution}
  \centering
  \begin{tabular}{ccccccccc}
    \toprule
    Tags             & Straight       & Right      & Left       & Stopped          & Slow      & Intersection     & Close & ASV \\
    \midrule
    Scenarios        & $163,079$      & $1,463$    & $2,578$    & $53,566$         & $37,414$ & $28,197$          & $11,232$ & $15,329$ \\
    Ratio            & $89.6$ $\%$    & $0.8$ $\%$ & $1.4$ $\%$ & $29.4$ $\%$      & $20.6$ $\%$    & $15.5$ $\%$ & $6.2$ $\%$ & $8.4$ $\%$ \\
    \bottomrule
  \end{tabular}
\end{table}

\subsection{Metrics}
\label{subsec:metrics}
We evaluate our model using a variety of metrics to give a full picture of driving.
To approximate real-world conditions, we perform a closed-loop replay for each scenario for a duration of 10 seconds.
We initialize the ego at the start of the scene, compute a planned trajectory, move along that trajectory for one step, replay the other agents, and repeat.
Then, we compute metrics on the resulting executed trajectory as averages over the full scene duration.

\noindent\textbf{Metric computation}:
Evaluation was done with a time step size of $0.2$ seconds and a total duration of $10$ seconds.  
The model was given $1$ second of ground truth past prior to the start of the scene.  
Other road users were updated by replaying their positions from the database.

\noindent\textbf{Metric categories}:
We have four high level categories of metrics that contain ``low level'' metrics and high level summary metrics that act as a score for the category.
The categories for our metrics are Safety, Comfort, Progress, and $\ell_2$ (with a yaw penalty of $2.5$).
Further details about our metric categories and the low level computations that make them up can be found in Appendix~\ref{app:metricbreakdown}.

\noindent\textbf{Metric limitations}:
Currently, there are two major limitations to our metrics evaluation.

\textit{The use of replay for other agents.} 
Since other vehicles do not react to the ego (e.g., if we drive slower than the expert in the data), the overall ``Safety'' score is a lower bound on safety.

\textit{No controller or vehicle dynamic simulation for the Ego.}
We currently ``teleport'' the ego along its trajectory, causing jerk to be erroneously high in some cases.
Similar to safety, this makes our ``Comfort'' score a lower bound on what we actually observe in real world driving.

\subsection{Model ablations}
\label{subsec:model-ablations}
In our experiments, we use a batch size of $64$ and an Adam optimizer with an initial learning rate of $10^{-3}$. 
Additionally, we use a ``cosine annealing with warm restarts'' scheduler, which gradually lowers the learning rate to a minimum of $10^{-4}$ and resets it every seven epochs.  
All models are trained over 20 epochs on eight AWS g4dn-metal instances with eight 16 GB NVIDIA Tesla T4 GPUs each. 
Because closed-loop simulation is computationally expensive, we randomly sampled $1,000$ scenarios from our evaluation set for ablation studies and $3,000$ scenarios from the test set for the final performance evaluation against other baselines.
Training and closed-loop metrics evaluation takes about an hour per epoch.

\noindent\textbf{Feature importance}:
To understand the importance of each hand-engineered feature and the main contribution of each, an ablation study for features is conducted and summarized in Tab.~\ref{tab:feature-importance}. 
The relative importance of each feature is shown by dropping one of them out at a time. 
We claim that all the features are important because the \textit{Base} model which includes all features got the highest scores across all high-level metrics and had lowest Collision rate. 
Even though the $\ell_2$ error is a bit higher compared to \textit{No MaxJerk}, the $0.089$ $m$ difference is not significant in the qualitative results. 
The results also demonstrated the importance of \textit{PastCoupling} feature in ensuring Comfort.
The experiment also showed that \textit{TTC} feature contributes significantly to reducing the collision rate.

\begin{table}
  \caption{Ablation study on the importance of each feature. The row indicates the feature removed from the baseline model. See Sec.~\ref{subsec:traj-score} for definitions.}
  \centering
  \begin{tabular}{lcccccc}
    \toprule
    Model   &Safety & Comfort & Progress & $\ell_2$ (w/ yaw)  & Collision   & Tailgate    \\
    \midrule
    Base (ours)    & $\textbf{0.925}$   & $\textbf{0.840}$  & $\textbf{0.988}$  & $2.290$ & $\textbf{0.001}$ & $0.015$ \\
    No TTC    & $0.865$   & $0.790$  & $0.958$  & $2.679$ & $0.055$ & $0.070$ \\
    No ACCInfo    & $0.862$   & $0.817$  & $0.959$  & $2.832$ & $0.004$ & $0.035$ \\
    No MaxJerk    & $0.863$   & $0.825$  & $0.960$  & $\textbf{2.201}$ & $\textbf{0.001}$ & $0.029$ \\
    No MaxLatAccel    & $0.917$   & $0.821$  & $0.982$  & $2.299$ & $0.005$ & $\textbf{0.011}$ \\
    No PastCoupling    & $0.901$   & $0.697$  & $0.979$  & $2.905$ & $0.005$ & $\textbf{0.011}$ \\
    No SpeedLimit    & $0.881$   & $0.809$  & $0.987$  & $2.483$ & $0.002$ & $0.028$ \\
    \bottomrule
  \end{tabular}
  \label{tab:feature-importance}
\end{table}

\noindent\textbf{Data augmentation}:
Data augmentation is important to ensure that our model can learn how to recover from errors.
Since the reference trajectory is never followed perfectly by the vehicle, errors can accumulate.
We perturb the ego's initial state during training to reduce the sensitivity to such errors.
For our low noise baseline, we use zero-mean Gaussian data augmentation for longitudinal offset ($\SI{1.2}{\meter}$ std), lateral offset ($\SI{0.8}{\meter}$ std), heading offset ($\SI{0.1}{\radian}$ std), and velocity ($\SI{0.1}{\meter \per \second}$ std).
For the high noise ablation, we respectively use $\SI{2.5}{\meter}$ std, $\SI{1.5}{\meter}$ std, $\SI{0.3}{\radian}$ std, and $\SI{0.2}{\meter \per \second}$ std.
We clamp velocity to avoid negative values.
Several example images are shown in \ref{section:da}.

\begin{table}
  \caption{Comparison between different augmentation schemes.}
  \label{tab:data-augmentation}
  \centering
  \begin{tabular}{lcccccc}
    \toprule
    Model   & Safety & Comfort & Progress & $\ell_2$ (w/ yaw)  & Collision   & Tailgate    \\
    \midrule
    Base (low noise)    & $\textbf{0.925}$   & $0.840$             & $\textbf{0.988}$  & $2.290$ & $\textbf{0.001}$ & $\textbf{0.015}$ \\
    No noise   & $0.850$   & $\textbf{0.850}$             & $0.956$  & $\textbf{2.289}$ & $0.005$ & $0.055$\\
    High noise  & $0.917$   & $0.845$             & $0.986$  & $2.525$ & $0.002$ & $0.016$\\
    Low past + present        & $0.921$   & $0.817$             & $0.982$  & $2.302$ & $0.005$ & $0.019$ \\
    \bottomrule
  \end{tabular}
\end{table}

\noindent\textbf{Architecture}:
We perform several ablations on the model architecture before selecting an architecture in which the extracted features are processed separately before interacting with one another through a masked self-attention mechanism.
We show in Tab.~\ref{tab:model-architecture} that the other two extremes, namely, concatenating all input features and using them as one monolithic feature in a single feedforward network or siloing all input features (not allowing any interaction through attention or otherwise) have resulted in inferior performance.
It is also seen that input normalization and attention input masking are beneficial.

\begin{table}
  \caption{Comparison between different model architectures.}
  \label{tab:model-architecture}
  \centering
  \begin{tabular}{lcccccc}
    \toprule
    Model   & Safety & Comfort & Progress & $\ell_2$ (w/ yaw)  & Collision   & Tailgate    \\
    \midrule
    Base (ours)    & $\textbf{0.925}$   & $\textbf{0.840}$             & $\textbf{0.988}$  & $2.290$ & $\textbf{0.001}$ & $0.015$ \\
    Monofeature FC  & $0.816$   & $0.715$             & $0.922$  & $3.084$ & $0.009$ & $0.020$\\
    Siloed FC   & $0.900$   & $0.784$             & $0.964$  & $2.394$ & $0.018$ & $0.021$\\
    No input norm   & $0.880$   & $0.817$             & $0.957$  & $2.496$ & $0.003$ & $0.016$\\
        Attention no masking   & $0.910$   & $0.835$             & $0.983$  & $\textbf{2.285}$ & $0.004$ & $\textbf{0.012}$\\
    \bottomrule
  \end{tabular}
\end{table}

\noindent\textbf{Loss}:
Tab.~\ref{tab:loss-function} shows that it is better to maximize the probability the projection of the ground truth onto the trajectory set (the best approximation in average $\ell_2$ norm) instead of the ground truth itself.
This makes sense because the ground truth does not come from the same distribution as the proposals and is not available at inference time.
Filtering possibly unsafe trajectories from the set before finding the ground truth projection is also crucial to obtaining a safe model.
Doing the projection using the average $\ell_2$ norm instead of an $\ell_2$ norm with a yaw error penalty also seems favorable.
Lastly from the same table, we can see that using focal loss as in Equation~\ref{eq:loss-function} improves performance.
Another experiment in Appendix~\ref{app:data_curation} that compares focal loss against training on a better balanced dataset also shows that using focal loss is actually more effective for DriveIRL.

\begin{table}
  \caption{Comparison between different loss functions.}
  \label{tab:loss-function}
  \centering
  \begin{tabular}{lcccccc}
    \toprule
    Metric   & Safety & Comfort & Progress & $\ell_2$ (w/ yaw)  & Collision   & Tailgate    \\
    \midrule
    Base (ours)    & $0.925$   & $\textbf{0.840}$             & $0.988$  & $\textbf{2.290}$ & $\textbf{0.001}$ & $\textbf{0.015}$ \\
    GT as demo   & $0.910$   & $0.553$             & $\textbf{0.992}$  & $2.870$ & $0.012$ & $0.029$\\
    Possibly unsafe demo  & $0.873$   & $0.823$             & $0.977$  & $2.518$ & $0.036$ & $0.049$\\
    Demo w/ weighted yaw        & $\textbf{0.932}$   & $0.839$             & $0.984$  & $2.530$ & $0.005$ & $0.018$ \\
    Without focal loss        & $0.910$   & $0.831$             & $0.976$  & $2.393$ & $0.004$ & $0.018$ \\
    \bottomrule
  \end{tabular}
\end{table}

\subsection{Baselines}
\label{subsec:baseline}
In this section, we evaluate our model on a test dataset and compare it with an Intelligent Driver Model (IDM) \citep{treiber2000congested-idm} and a constant speed (CS) lane follow model. 
The IDM baseline is a reasonable choice because it is a well-known version of an expert planner that focuses on adaptive cruise control. 
Meanwhile, the CS lane follow model is a simple lower-bound.
The results are shown in Tab.~\ref{tab:baselines}.
Our base model plus safety filter outperforms others in all safety related metrics, and that shows the safety filter protects the vehicle on several collision cases our model cannot handle perfectly.
Without the safety filter, our base model still outperforms the IDM baseline.
The IDM model has significantly higher $\ell_2$ error, indicating that the IDM model does not drive like a human expert.
Furthermore, our base model also has higher scores in all safety related metrics in both high- and low-level scores like collision rate and tailgate rate. 

\begin{table}
  \caption{Baselines on the test set. IDM = Intelligent Driver Model. CS = constant speed.}
  \centering
  \begin{tabular}{lcccccc}
    \toprule
    Metric   & Safety & Comfort & Progress & $\ell_2$ (w/ yaw)  & Collision   & Tailgate    \\
    \midrule
    Expert     & $1.000$   & $0.984$             & $1.000$  & $0.000$ & $0.000$ & $0.000$ \\
    \midrule
    Base + Safety (ours)   & $\textbf{0.930}$   & $0.815$             & $0.971$  & $\textbf{2.204}$ & $\textbf{0.001}$ & $\textbf{0.006}$ \\
    Base (ours)    & $0.916$   & $0.830$             & $0.986$  & $2.351$ & $0.003$ & $0.017$ \\
    IDM    & $0.891$   & $0.898$             & $\textbf{0.987}$  & $4.478$ & $0.005$ & $0.019$ \\
    CS lane follow    & $0.669$   & $\textbf{0.992}$             & $0.902$  & $3.963$ & $0.161$ & $0.166$ \\
     \bottomrule
  \end{tabular}
  \label{tab:baselines}
\end{table}

\subsection{Simulation results}
\label{subsec:sim-result}
\begin{figure*}
    \centering
    \subfigure[Ego starting from a stop.]
    {
        \frame{\includegraphics[width=.45\columnwidth]{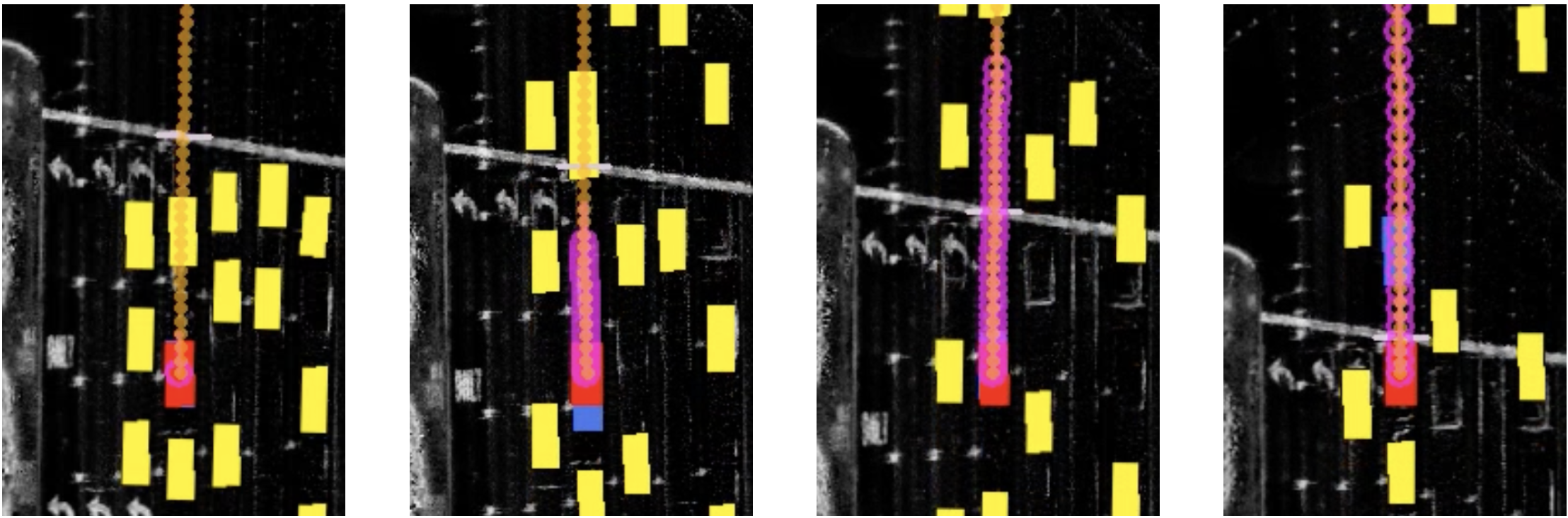}}
    }
    \subfigure[Ego stopping for the lead vehicle.]
    {
        \frame{\includegraphics[width=.45\columnwidth]{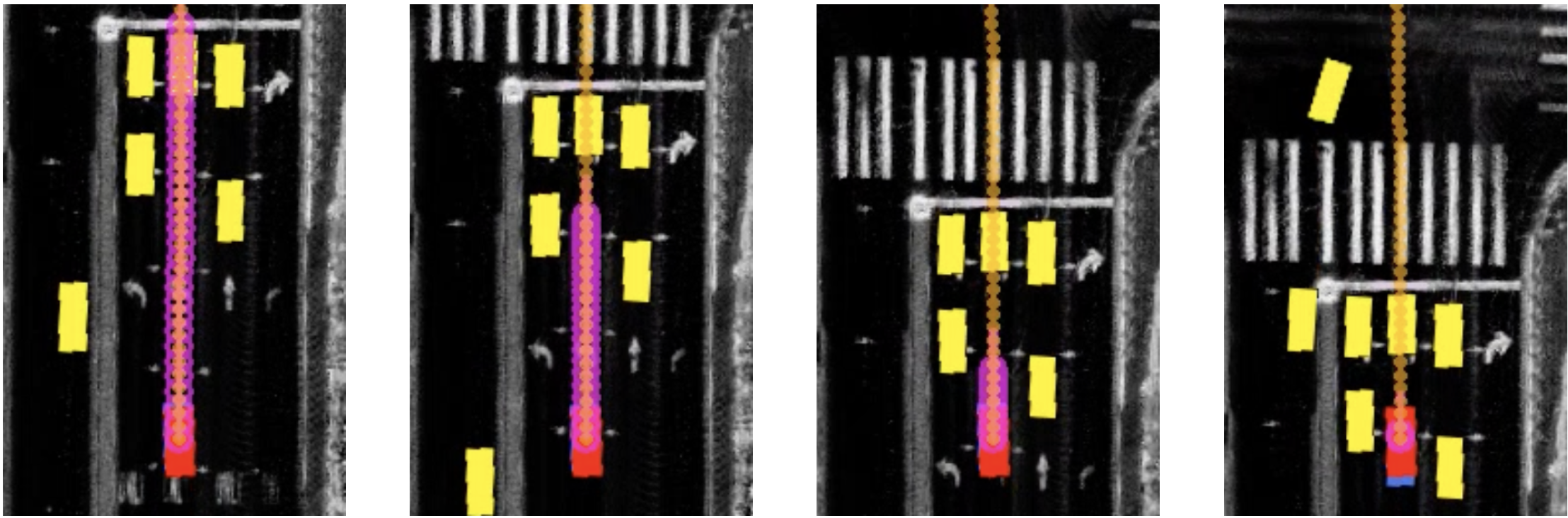}}
    }
    \subfigure[Adaptive cruise control.]
    {
        \frame{\includegraphics[width=.45\columnwidth]{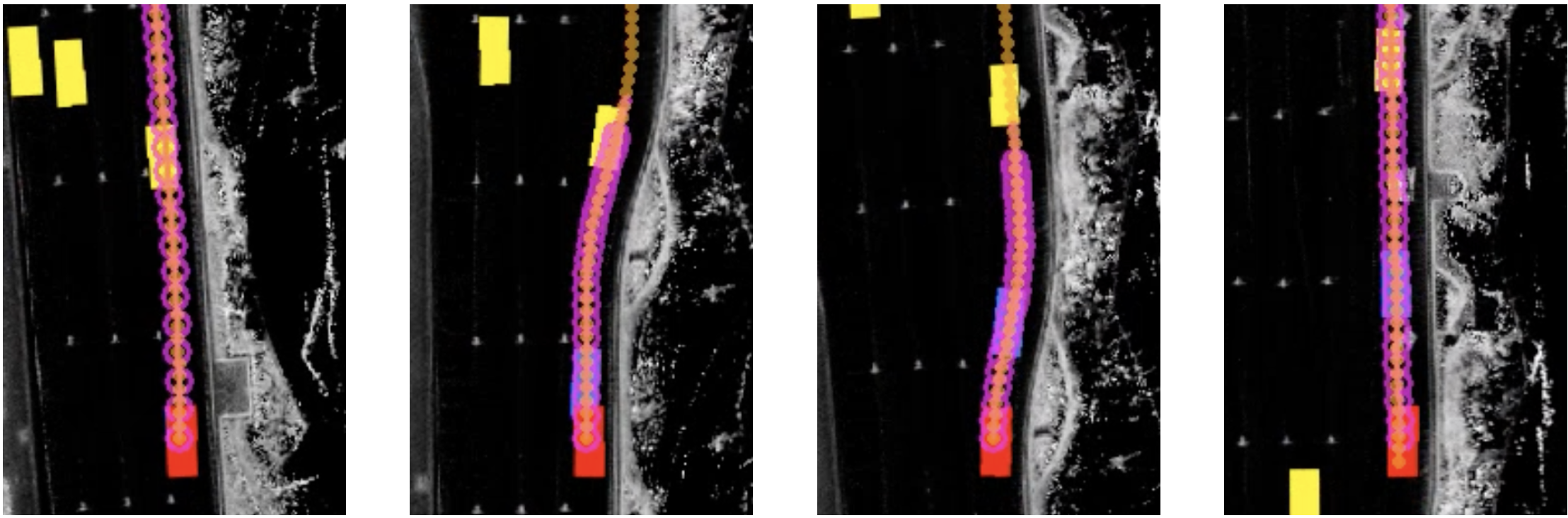}}
    }
    \subfigure[A moderate cut-in.]
    {
        \frame{\includegraphics[width=.45\columnwidth]{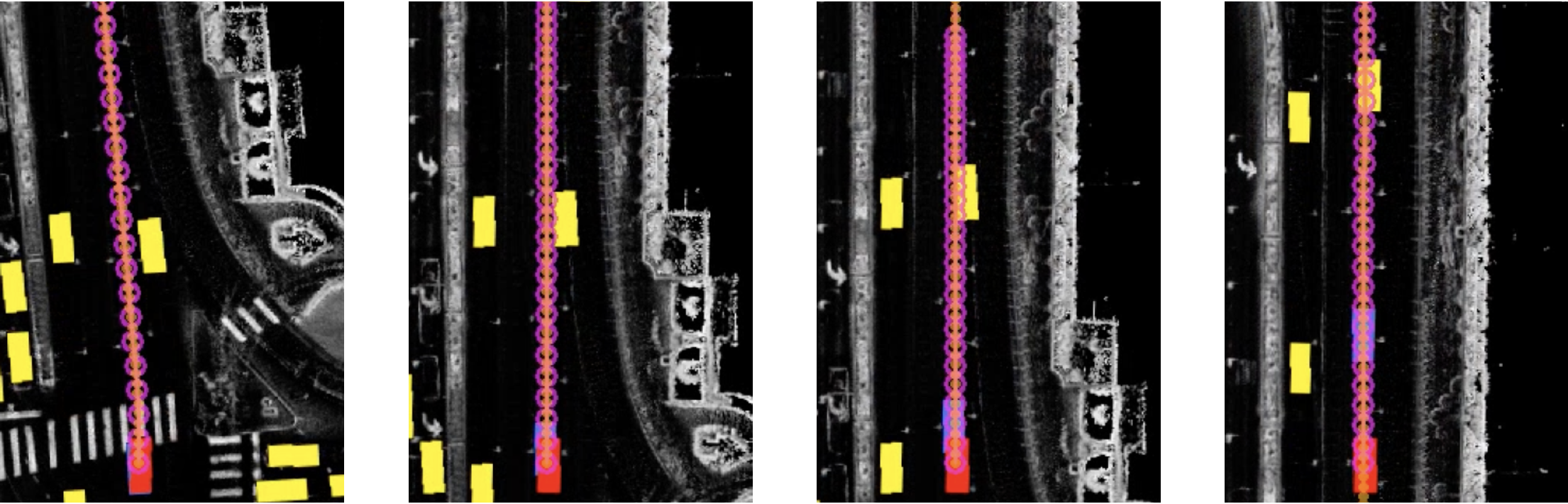}}
    }
    \vspace{-2mm}
    \caption{Qualitative driving performance in common scenarios. Full-size images are in~\ref{app:simulation_videos}.}
    \label{fig:sim-scenarios}
\end{figure*}

Fig.~\ref{fig:sim-scenarios} exhibits some qualitative closed-loop simulation results of our planner driving in typical scenarios.
These scenarios are shown as a sequence of snapshots along the closed-loop rollout.
The ego vehicle is shown as a red rectangle, the expert vehicle is in blue, and other vehicles are in yellow.
The orange line is the planned route (along the lane centerline) and the purple circles represent the planned trajectory for the next $6$ seconds.
Our planner shows good performance over a range of scenarios, exhibiting reasonable and consistent behavior over $10$ second rollouts.

\subsection{Real-world driving}
\label{subsec:realworld-driving}
Prior to deploying on public roads, DriveIRL was rigorously tested in both simulation and on private, closed-course routes.
The simulation tests consist of the same Las Vegas Strip route that was our deployment goal, and involve a high-fidelity dynamics model for the ego vehicle and numerous actors exhibiting a wide variety of behaviors.
When deployed on the Strip, the vehicle was piloted by a vehicle operator who was trained to take over for unsafe behavior and situations outside of our operating domain, including construction zones, bus stops, and yielding for emergency vehicles.

On the Strip, our planner handled challenging scenarios such as heavy traffic, aggressive cut-ins, unpredictable drivers, and busy passenger pick-up/drop-off zones near the hotels and casinos.
Without the safety filter, the vehicle remained in autonomous mode for 8.8 miles of the 11-mile route.
Overrides occurred for mandatory takeover regions and twice for undesired behavior.
With the safety filter, the vehicle remained in autonomous mode for 6.9 of 8.5 miles, with takeovers only occurring due to mandatory takeover regions.

\begin{figure*}
    \centering
    \subfigure{\includegraphics[width=.325\columnwidth]{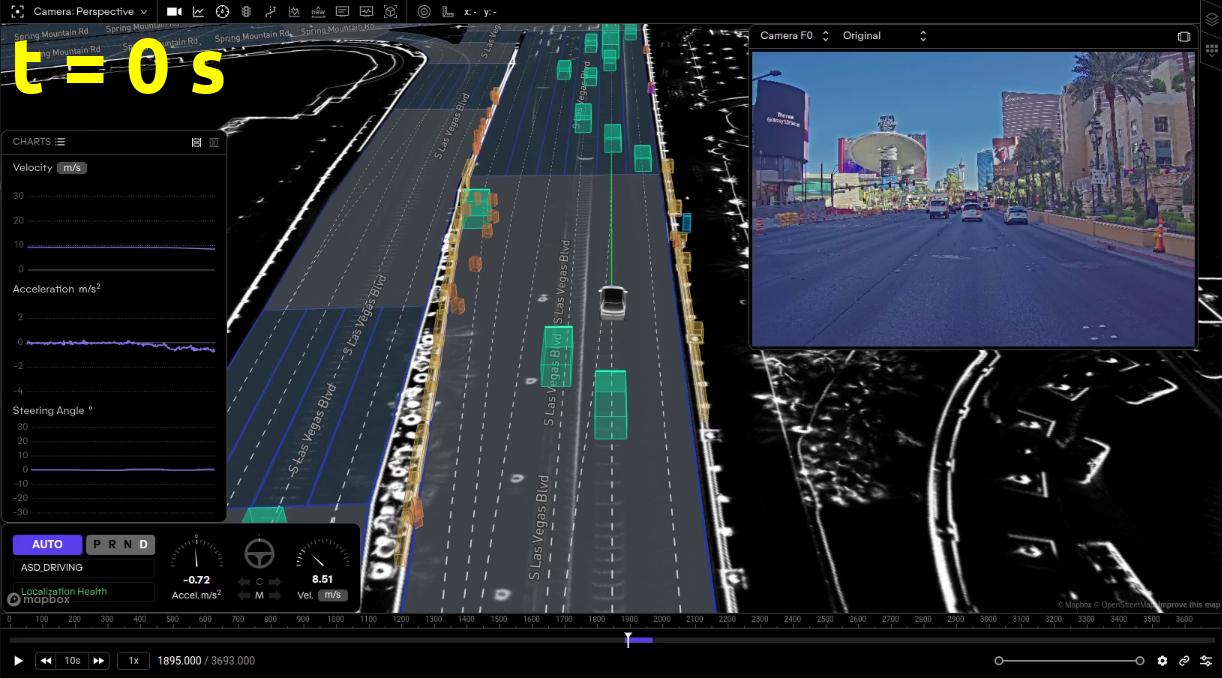}}
    \subfigure{\includegraphics[width=.325\columnwidth]{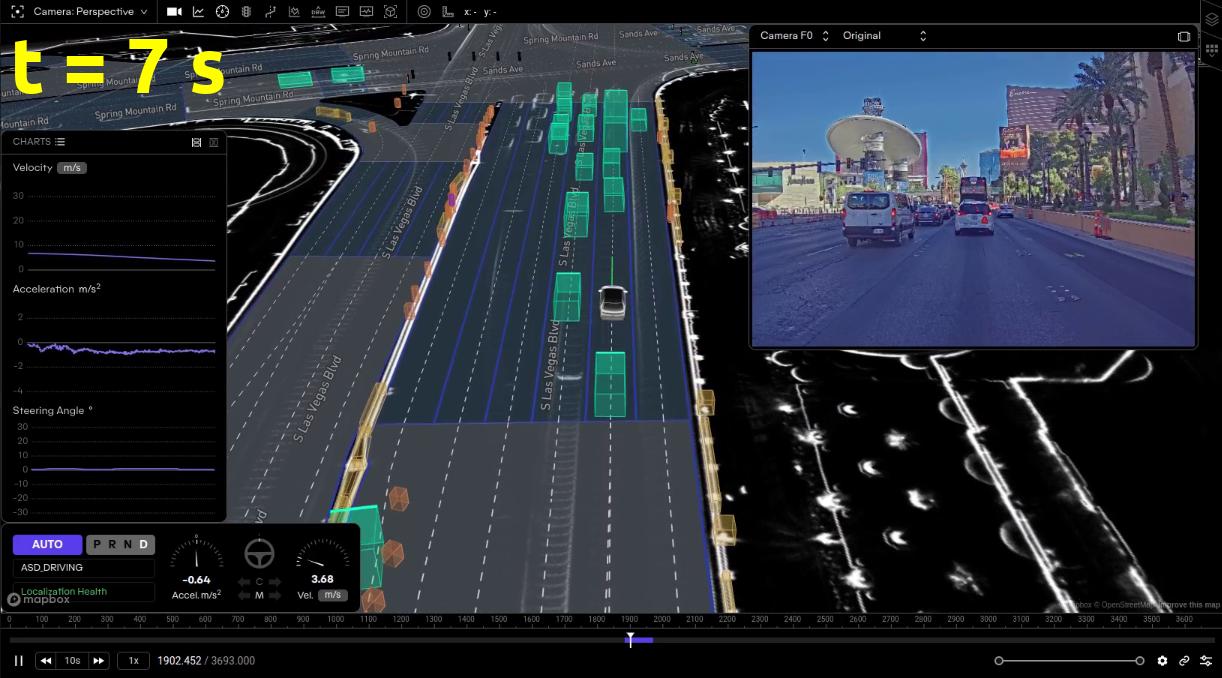}}
    \subfigure{\includegraphics[width=.325\columnwidth]{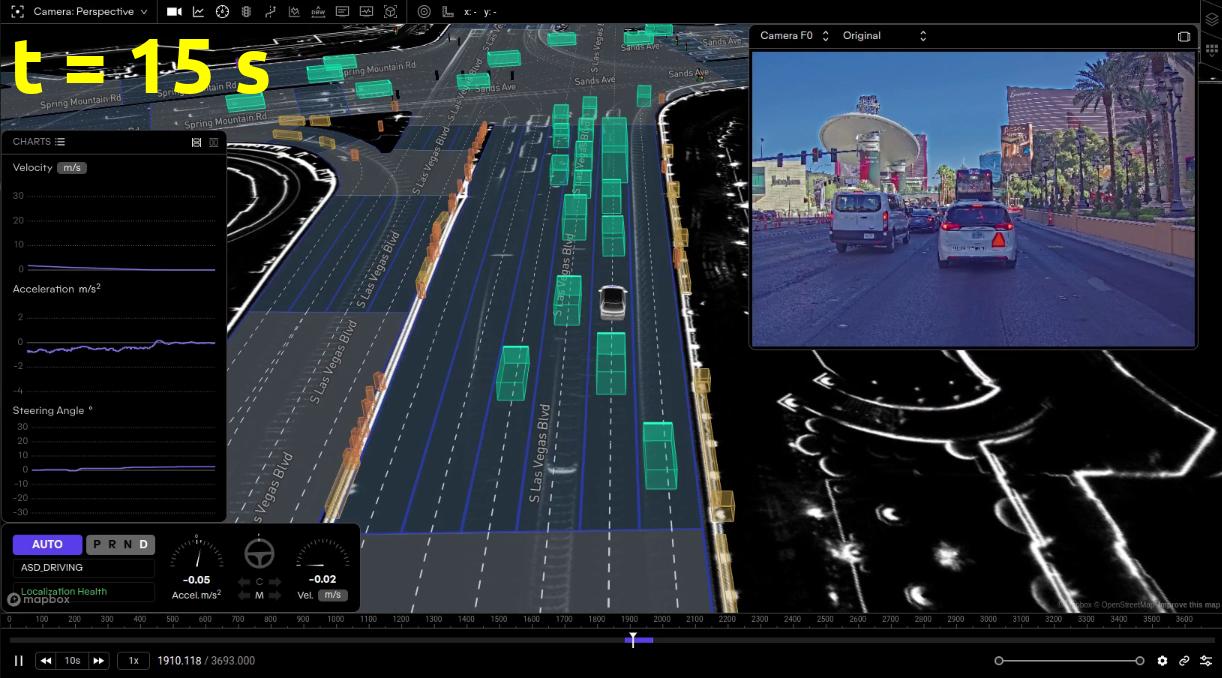}}
    \caption{Smoothly stopping behind a vehicle in dense traffic on the Las Vegas Strip.}
    \label{fig:stopping-02}
\end{figure*}

Fig.~\ref{fig:stopping-02} shows a typical maneuver where our self-driving vehicle smoothly stops for a vehicle ahead while surrounded by multiple vehicles.
In Sec.~\ref{app:real-clips}, we have included video clips with more real-world driving maneuvers.
These videos are grouped in categories such as cut-ins, driving around passenger pickup/dropoff zones, driving with a vehicle ahead and stopping behind a vehicle.


\section{Conclusions}
We introduced DriveIRL: the first learning-based planner to control a car in dense, urban traffic using inverse reinforcement learning.
By designing an architecture split into ego trajectory \emph{generation}, \emph{checking}, and \emph{scoring}, we were able to leverage simple and reliable methods for the relatively easy tasks of trajectory generation and safety checking.
This architecture allowed the trajectory scoring component of our system to focus on learning nuanced driving behavior important for good performance in dense traffic.
We demonstrated our planner on the busy Las Vegas Strip, where it showed strong real-world performance on challenging scenarios such as cut-ins, abrupt braking, and cluttered hotel pickup/dropoff zones.


\begin{ack}
We would like to thank our colleagues Juraj Kabzan, Dimitris Geromichalos, Christopher Eriksen, Whye Kit Fong, Gordon Gustafson, Qiang Xu, Elena Corina Grigore, and Sunaya Bajracharya for their help and support throughout this project.
\end{ack}

{\small
\bibliographystyle{unsrtnat}
\bibliography{bibliography}
}

\newpage
\appendix

\section{Supplemental Material}

\subsection{Trajectory quality}
\label{app:trajectory-quality}
We evaluate the quality of the trajectory generation by projecting the expert (ego) trajectory on the closest (in $\ell_2$) trajectory from the trajectory generator (Expert Projection).
We compare this with the actual expert trajectory (Expert) and the trajectory chosen by our best model (Base) on 6-second scenarios in Tab.~\ref{tab:trajectory-quality}.
Note the use of shorter scenarios to match the length of the (open-loop) trajectory.
The low $\ell_2$ for the Expert Projection indicates that there is enough diversity in the trajectory set to closely match the expert.
These results also indicate that, given the trajectory set, our model is near ceiling performance in terms of the high-level metrics, but it could improve on $\ell_2$.

\begin{table}[hb]
  \caption{Trajectory set quality.}
  \centering
  \begin{tabular}{lcccccc}
    \toprule
    Metric   & Safety & Comfort & Progress & $\ell_2$ (w/ yaw)  & Collision   & Tailgate    \\
    \midrule
    Base (ours)    & $0.968$   & $0.851$             & $0.989$  & $1.216$ & $0.000$ & $0.004$ \\
    Expert    & $1.000$   & $0.962$             & $1.000$  & $0.000$ & $0.000$ & $0.000$ \\
    Expert Projection    & $0.984$   & $0.880$             & $0.996$  & $0.309$ & $0.000$ & $0.000$ \\
     \bottomrule
  \end{tabular}
  \label{tab:trajectory-quality}
\end{table}

\subsection{Safety filter details}
\label{app:safety-filter-details}
For each candidate trajectory, we check that the distance to the track ahead never falls below $\SI{1.5}{\meter}$, assuming that all non-ego vehicles perform a hard brake at $\SI{3.5}{\meter \per \second^2}$.
We perform the check on modified ego trajectories, where each trajectory is followed for the first $\SI{1}{\second}$, after which firm braking is applied along the trajectory at $\SI{2.5}{\meter \per \second^2}$, subject to a $\SI{3.5}{\meter \per \second^3}$ jerk limitation.
In effect, this implements a kind of recursive safety guarantee: if we follow the trajectory for $\SI{1}{\second}$ and then brake, are we still safe?

\subsection{Feature implementation details}
\label{app:features}
We compute features for each proposed trajectory to use as inputs to our neural network.
These features are represented as $F^i(\tau, \mathcal{S}, \mathcal{U}, \mathcal{M}, \mathcal{R}, \mathcal{H} ) = f^i \in \mathbb{R}^{k_i}$, where $F^i$ is the feature extraction function corresponding to feature $i$ and $k_i$ is its dimension, and the output feature vector is $f^i$.
A feature vector $f^i$ can be either a sequence of feature tuples $[f_1, \ldots, f_T]$ where $f_t$ is a feature tuple at timestamp $t$, or just a single feature tuple $[f_w]$ which encodes the characteristics of the whole proposed trajectory regardless of timestamps.

\begin{itemize}
    \item \textit{Time-to-collision (TTC)}: The number of seconds before (predicted) collision with other agents. Because the computation time for this feature is expensive, we sub-sample the trajectory waypoints at $[0.2, 0.4, 0.6, 1.0, 2.0, 4.0]$ seconds. The TTC saturation time is set to $4.0$ seconds, meaning agents that the ego is not projected to collide into within $4.0$ seconds are effectively ignored. Namely, the feature function is: $F_{TTC}(\tau, \mathcal{S}, \mathcal{U}, \mathcal{M}) \rightarrow [f_{1\_collide}, \ldots, f_{T\_collide}]$ where $f_{t\_collide}$ denotes the minimum time to collision at timestamp $t$.
    
    \item \textit{ACCInfo}: The information needed for adaptive cruise control. This feature complements the time-to-collision feature. When ego is slowly moving forward, the time-to-collision might not be very meaningful as its value may still be large even when the ego is already very close to the vehicle in front, which is not desirable. \textit{ACCInfo} directly addresses this issue by feeding the distance information to the model along with some other information. The feature function is $F_{ACCInfo}(\tau, \mathcal{S}, \mathcal{U}, \mathcal{M}) \rightarrow [f_{1\_acc}, \ldots, f_{T\_acc}]$ where $f_{t\_acc}$ is $(d_{ahead}, b_{ahead}, v_{ego}, v_{ahead}, v_{ego} - v_{ahead})_t$ where $d_{ahead}$ is the distance to the track ahead, $b_{ahead}$ is a Boolean flag indicating if there is a track in front given a tunable distance, which by default is $\SI{20}{\meter}$, $v_{ego}$ and $v_{ahead}$ are the speed of the ego and the track ahead respectively, and lastly the relative speed between the ego and the track ahead is added.
    
    \item \textit{MaxJerk}: The maximum jerk ($\SI{}{\meter \per \second^3}$) along the trajectory. This feature does not just compute the jerk value at the current time given the past ego states. Rather, it computes all jerk values along the proposed trajectory, and then takes a maximum among these values as the single max jerk value $j_{\max}$. This feature function can be represented as: $F_{MaxJerk}(\tau, \mathcal{H}) \rightarrow [f_{w\_jerk}, j_{\max}]$. The way that we represent the jerk value in this feature tuple $f_{w\_jerk}$ is through a series of Boolean flags as this representation is easier for the model to consume, similar to the idea of one-hot encoding. Given a range of interest for the jerk value, a set of thresholds are assigned among the range with a given resolution. The range in our experiments used is $[0, 10]$ with a fixed step size of $\SI{0.5}{\meter \per \second^3}$. If the jerk value is smaller than a threshold, the corresponding flag will be set to $1$; otherwise $0$, meaning smaller max jerk value gets more flags set to $1$. We also append the original jerk value  $j_{\max}$ at the end of this feature tuple for the model to consume.
    
    \item \textit{MaxLateralAccel}: The max lateral acceleration ($\SI{}{\meter \per \second^2}$) along the trajectory. Similar to the MaxJerk feature, the MaxLateralAccel feature computes the maximum acceleration in a lateral direction w.r.t. the ego, representing another measurement of comfort. The same quantization representation as used for MaxJerk above is used in this feature as well, namely $F_{MaxLateralAccel}(\tau, \mathcal{H}) \rightarrow [f_{w\_lat}, a^{\textit{lat}}_{\max}]$ where $f_{w\_lat}$ is a series of Boolean flags. The threshold values range from $0$ to $\SI{5}{\meter \per \second^2}$ in step sizes of $\SI{0.2}{\meter \per \second^2}$. Note that original value of the maximum lateral acceleration $a^{\textit{lat}}_{\max}$ is appended at the end as well.
    
    \item \textit{PastCoupling}: The concatenation of the proposed trajectory and the past ego states. As another complementary comfort feature, we found that a feature that represents all information across all time stamps is also helpful in terms of comfort level because it makes model learn the level of coherence between the past, present, and future trajectory. 
    Specifically, $F_{PastCoupling}(\tau, \mathcal{H}) \rightarrow [f_{1\_coup}, \ldots, f_{T\_coup}]$ where $f_{t\_coup}$ is a five-tuple $(x, y, \theta, v, a)_t$ at the timestamp $t$, and $v$ and $a$ stand for the ego speed and acceleration.
    
    \item \textit{SpeedLimit}: Whether the ego drives under or close to the speed limit.
    The speed limit of the road block can be queried from our map data given the ego position. 
    For each waypoint in our proposed trajectory, two values are computed, a relative speed based on the difference between the speed limit and the current speed, and a Boolean flag $b_{limit}$ to indicate if the ego speed is over the speed limit, meaning that the feature function is: $F_{SpeedLimit}(\tau, \mathcal{S}, \mathcal{M}) \rightarrow [f_{1\_limit}, \ldots, f_{T\_limit}]$ where $f_{t\_limit}$ is a two tuple $((v_{ego} - v_{limit})/v_{limit}, b_{limit})_t$.
    To make the relative speed more uniform between the high speed case and the low speed case, the relative speed is also normalized by the value of speed limit itself.
    
    
\end{itemize}

\subsection{Scenario tags}
\label{app:scenario_tags}
A detailed distribution of our dataset is shown in Tab.~\ref{tab:dataset-distribution}.
The scenario tags in the table are not mutually exclusive and a sample can belong to multiple tags.

We classified a scenario as \textit{Straight} if the ego drove straight throughout the scenario and the changes in the initial and final heading angles of ego were less than $\SI{0.1}{\radian}$.
A scenario was classified as a \textit{Right} or \textit{Left} turn if the ego made a turn in the scenario and the change in the heading angles was greater than $\frac{\pi}{3}$ $\SI{}{\radian}$.
Due to the geometry of the Las Vegas Strip, the ego mostly drove straight.
Turns were made occasionally to enter or leave pick-up or drop-off areas.
As the definitions in \textit{Straight} and \textit{Turn} tags indicate, these tags do not form a perfect partition.
Thus, the number of scenarios for these tags do not add up to the total.

\textit{Stopped} means the ego did not move through out the scenario.
This is largely because the ego was in the middle of a traffic jam or waiting for pedestrians.
We classified a scenario as \textit{Slow} if the ego was driving slower than $\SI{2.64}{\meter \per \second}$.
In our dataset, ego drove relatively slowly.
For example, we had $33,646$ scenarios where ego drove faster than $\SI{9.64}{\meter \per \second}$.
This is because of the following reasons: i) the speed limit of the Strip was $\SI{15}{\meter \per \second}$ and ii) there was a lot of traffic in the Strip.
The heavy traffic makes our dataset low-speed but challenging.

Whenever the ego went through an intersection area during a scenario, it was tagged as \textit{Intersections} regardless of the traffic light status.

There are also tags that capture the interactions of the ego and the leading vehicle, namely \textit{Close} and \textit{Approaching Stopped Vehicle (ASV)}.
A scenario was classified as \textit{Close} if the ego had time gap less than $\SI{1.7}{\second}$, calculated by distance to the vehicle ahead divided by ego speed.
\textit{ASV} was when the ego was moving toward a leading stopped vehicle and got closer than $\SI{10}{\meter}$.

\subsection{Metrics breakdown}
\label{app:metricbreakdown}
The following is a break down of our metric categories and how their respective metrics are computed:
\begin{itemize}
    \item \textit{Safety}: is the most important category and it is composed of the following items:
        \begin{itemize} 
            \item \textit{collision rate}: 
            Calculated as whether the ego hit any vehicle in \textit{front} of it.
            \footnote{We use front collision rather than total collision, because we are replaying the other agents, which leads to many false positive rear collisions that are not caused by the ego's behavior.  
            Specifically, if the ego stops but the expert did not then cars that were following the expert may then proceed to collide with the ego vehicle.}  
            \item \textit{ego not off-road}:  
            Measures whether or not the ego vehicle goes off the road at any point of the scene.
            \item \textit{ego minimum time to collision > $\SI{0.95}{\second}$
            \footnote{All thresholds for metrics were derived from examining expert driving and were adjusted based on feedback of what worked and what didn't during deployment in the Las Vegas Strip.}}:
            Measures whether or not the time to collision of the ego was always above the set threshold
            \item \textit{tailgate rate, i.e. ego distance to lead vehicle < $\SI{1.5}{\meter}$ }: 
            Measures whether the ego was appropriately far from lead vehicles.  
            Note that this metric is especially important during slow scenarios where the time to collision is not very informative.
        \end{itemize}
    \item \textit{Comfort}: 
    is based on acceleration, jerk and yaw rate and is considering passing based on whether they are within the following thresholds: $\SI{-4.05}{\meter \per \second^2}$ < \textit{ego longitudinal acceleration} < $\SI{2.40}{\meter \per \second^2}$, \textit{ego lateral acceleration} < $\SI{4.89}{\meter \per \second^2}$, \textit{ego absolute yaw rate} < $\SI{0.95}{\radian \per \second}$, \textit{ego absolute yaw acceleration}  < $\SI{1.93}{\radian \per \second^2}$, \textit{ego absolute longitudinal jerk}  < $\SI{4.13}{\meter \per \second^3}$, and \textit{ego absolute jerk magnitude}  < $\SI{8.37}{\meter\per\second^3}$.
    \item \textit{Progress}: 
    is made up of only two components, 
    (1) whether or not the ego has made positive distance to the goal of more then $\SI{1}{\meter}$ when projected onto the route given to the ego and 
    (2) whether or not the ego has diverged from the route, measured as whether the maximum distance from the ego to the route is greater than $\SI{4}{\meter}$.
    \item $\ell_2$: 
    These metrics measure the distance between the resulting trajectory of the ego and the path the expert actually drove on the scenario.  
    They have no high level summary.
    Some examples of $\ell_2$ metrics that have been useful for us include \textit{ego average l2 with yaw} and \textit{ego average cross-track (i.e. longitudinal) l2}.
    For $\ell_2$ with yaw this is calculated as standard $\ell_2$ between the trajectories with an adjustment of $2.5$ times the principal value of the heading differences between the trajectories.
\end{itemize}

\subsection{Data augmentation examples}
\label{section:da}
\begin{figure*}
    \centering
    \subfigure{\includegraphics[width=.15\columnwidth]{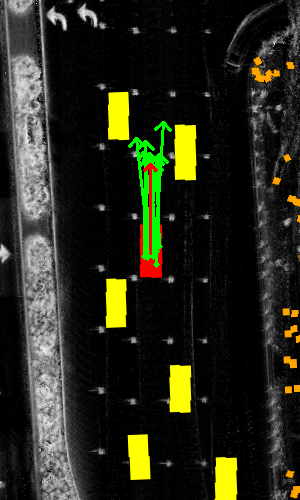}}
    \subfigure{\includegraphics[width=.15\columnwidth]{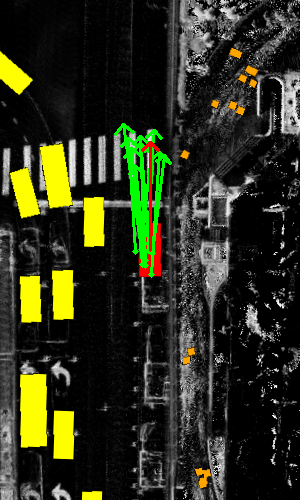}}
    \subfigure{\includegraphics[width=.15\columnwidth]{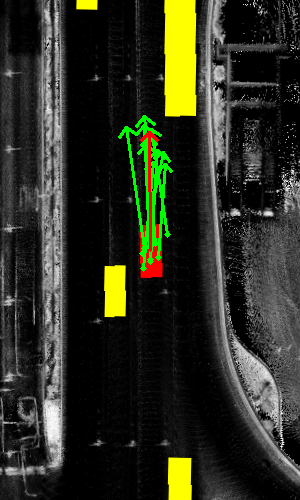}}
    \subfigure{\includegraphics[width=.15\columnwidth]{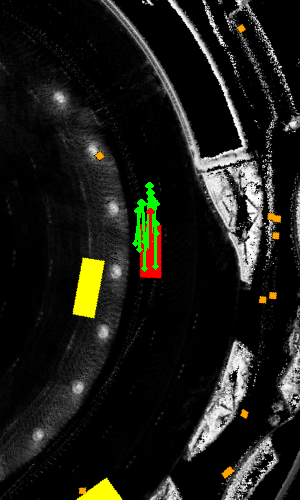}}
    \subfigure{\includegraphics[width=.15\columnwidth]{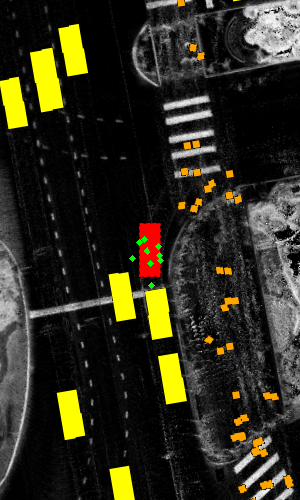}}
    \caption{Data augmentation examples.}
    \label{fig:data-curation}
\end{figure*}

Several examples of data augmentation are shown in Fig.~\ref{fig:data-curation}.
The original ego pose and velocity are shown as a red rectangle and arrow respectively while the speed scales with the length of arrow.
The augmented ego center positions and velocities are shown as green points and arrows respectively.
We plot $10$ random results for demonstration.

\subsection{Data balancing}
\label{app:data_curation}
As shown in Tab.~\ref{tab:dataset-distribution}, the distribution of our dataset is unbalanced.
For example, there are a lot more \textit{Straight} scenarios compared to \textit{Right} or \textit{Left} turn scenarios.
In addition, the interaction between the ego and the leading vehicle is crucial in ACC.
This interaction is captured in \textit{Close} and \textit{ASV} scenarios, but they account for only $6.2\%$ and $8.4\%$, respectively.
We performed balancing on our train set to make sure our model learns rare but important scenarios well.
To be more specific, we took an upsampling approach and increased the ratio of rare scenarios by duplicating them multiple times in the train set.
Even though the scenarios were copied, our data augmentation gave these scenarios some variation.
Tab.~\ref{tab:balanced-dataset-distribution} shows the distribution of our balanced train set.
Compared to Tab.~\ref{tab:dataset-distribution}, the ratios of rare scenarios such as \textit{Right}, \textit{Left}, \textit{Close} and \textit{ASV} increased whereas those of frequent scenarios such as \textit{Stopped} and \textit{Straight} decrease.
We performed balancing only on the train set and did not modify the validation and test sets.

\begin{table}
  \caption{A detailed distribution of our balanced train set ($125,875$ total scenarios).}
  \label{tab:balanced-dataset-distribution}
  \centering
  \begin{tabular}{ccccccccc}
    \toprule
    Tags             & Straight       & Right      & Left        & Stopped          & Slow      & Intersection     & Close & ASV \\
    \midrule
    Scenarios        & $103,411$      & $3,340$    & $4,464$     & $32,273$         & $26,339$ & $24,784$          & $12,600$ & $12,287$ \\
    Ratio            & $82.2\%$    & $2.7\%$ & $3.5\%$  & $25.6\%$      & $20.9\%$    & $19.7\%$ & $10.0\%$ & $9.8\%$ \\
    \bottomrule
  \end{tabular}
\end{table}

The first two rows of Tab.~\ref{tab:data-balancing} enables a comparison between using the original (unbalanced) train set and the balanced one.
The next two rows are for comparing two train sets when trained without local loss.
The balanced train set did not help much when applied on our base model.
However, the balanced train set showed improvement when applied on the model without focal loss.
This indicates using focal loss has a similar effect as using a balanced train set as expected.
In fact, by comparing the first (with focal loss + unbalanced train set) and fourth (without focal loss + balanced train set) rows, we see that using focal loss and a balanced train set have comparable performances.
Tab.~\ref{tab:data-balancing-detailed} compares the $\ell_2$ (with yaw) metrics of each scenario tags for the three models of \textit{Without focal loss}, \textit{Base (ours)}, and \textit{Balanced train set + without focal loss}.
We can see that the base model that used focal loss achieved better $\ell_2$ (with yaw) for rare scenarios such as \textit{Right}, \textit{Left}, and \textit{Close}.
This suggests using the focal loss is more effective in resolving the dataset imbalance issue.

\begin{table}
  \caption{Data balancing.}
  \label{tab:data-balancing}
  \centering
  \begin{tabular}{lcccccc}
    \toprule
    Metric   & Safety & Comfort & Progress & $\ell_2$ (with yaw)  & Collision rate   & Tailgate rate    \\
    \midrule
    Base (ours)    & $\textbf{0.925}$   & $\textbf{0.840}$             & $\textbf{0.988}$  & $2.290$ & $\textbf{0.001}$ & $\textbf{0.015}$ \\
    Balanced set \\ + base   & $0.913$   & $0.828$             & $0.984$  & $\textbf{2.209}$ & $0.005$ & $0.022$ \\
    \midrule
    Without focal        & $0.910$   & $0.831$             & $0.976$  & $2.393$ & $0.004$ & $0.018$ \\
    Balanced set \\ + without focal    & $\textbf{0.922}$   & $\textbf{0.851}$             & $\textbf{0.981}$  & $\textbf{2.280}$ & $\textbf{0.002}$ & $\textbf{0.014}$ \\
    \bottomrule
  \end{tabular}
\end{table}

\begin{table}
  \caption{$\ell_2$ (with yaw) metric for each scenario tag.}
  \label{tab:data-balancing-detailed}
  \centering
  \begin{tabular}{lccccccccc}
    \toprule
    Tags                           & All & Straight & Right    & Left    & Stopped   & Slow    & Intersec. & Close   & ASV    \\
    \midrule
    Base (ours)                    & $2.29$       & $2.15$  & $\textbf{4.50}$  & $\textbf{4.24}$ & $0.37$   & $2.22$ & $\textbf{3.54}$      & $\textbf{3.33}$ & $1.73$ \\
    Without focal             & $2.39$       & $2.24$  & $5.76$  & $5.09$ & $0.50$   & $\textbf{2.12}$ & $3.69$      & $3.52$ & $1.74$ \\
    Balanced set \\ + without focal    & $\textbf{2.28}$   & $\textbf{2.12}$             & $5.16$  & $5.07$ & $\textbf{0.34}$ & $2.12$ & $3.67$ & $3.54$ & $\textbf{1.57}$ \\
    \bottomrule
  \end{tabular}
\end{table}

\subsection{Simulation video clips}
\label{app:simulation_videos}
\begin{figure*}
    \centering
    \subfigure{\includegraphics[width=.23\columnwidth]{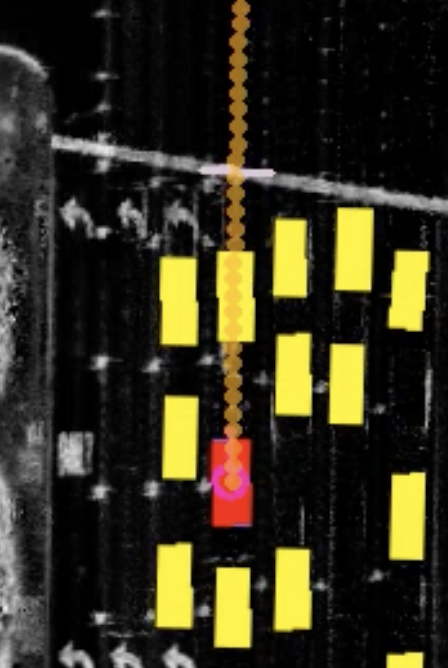}}
    \subfigure{\includegraphics[width=.23\columnwidth]{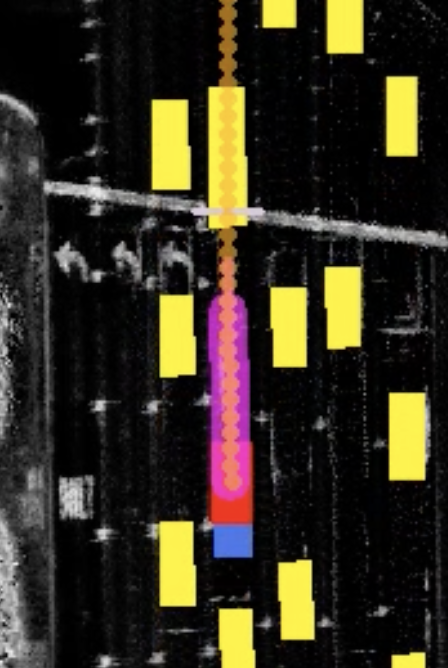}}
    \subfigure{\includegraphics[width=.23\columnwidth]{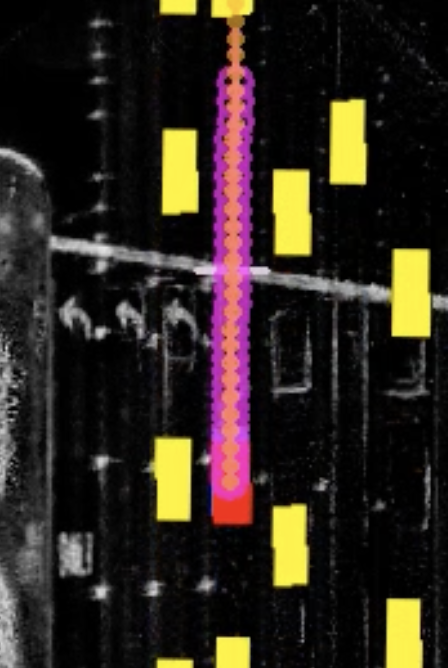}}
    \subfigure{\includegraphics[width=.23\columnwidth]{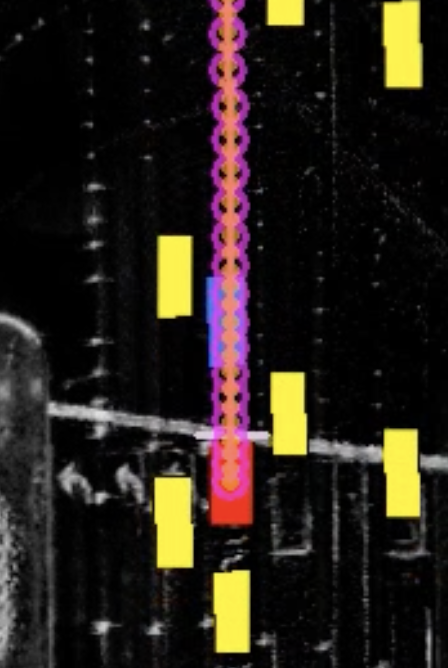}}
    \caption{The ego vehicle starts to move along side the vehicle ahead. Video clip: sim-start-from-stop1.mp4}
    \label{fig:sim-start-from-stop1}
\end{figure*}

\begin{figure*}
    \centering
    \subfigure{\includegraphics[width=.23\columnwidth]{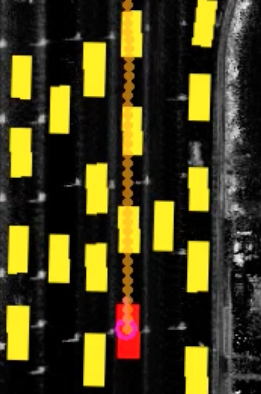}}
    \subfigure{\includegraphics[width=.23\columnwidth]{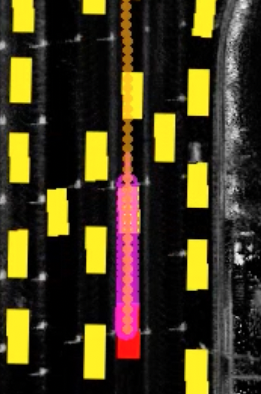}}
    \subfigure{\includegraphics[width=.23\columnwidth]{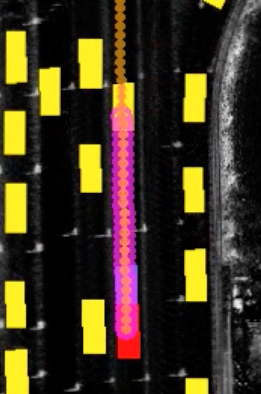}}
    \subfigure{\includegraphics[width=.23\columnwidth]{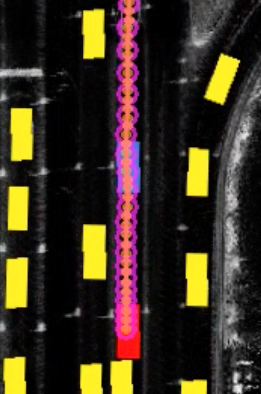}}
    \caption{Another case where the ego vehicle starts to move along side the vehicle ahead. Video clip: sim-start-from-stop2.mp4}
    \label{fig:sim-start-from-stop2}
\end{figure*}

\begin{figure*}
    \centering
    \subfigure{\includegraphics[width=.23\columnwidth]{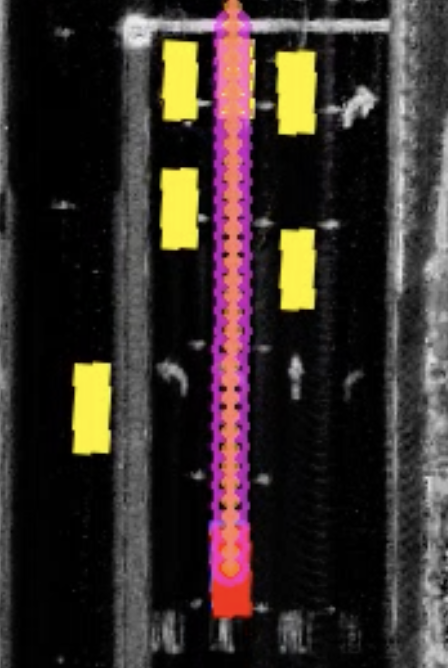}}
    \subfigure{\includegraphics[width=.23\columnwidth]{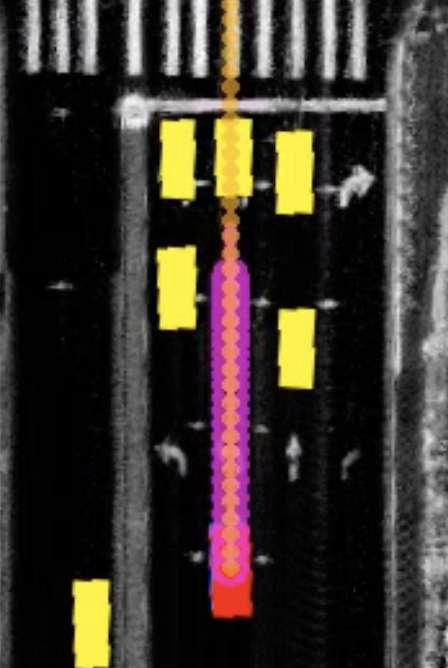}}
    \subfigure{\includegraphics[width=.23\columnwidth]{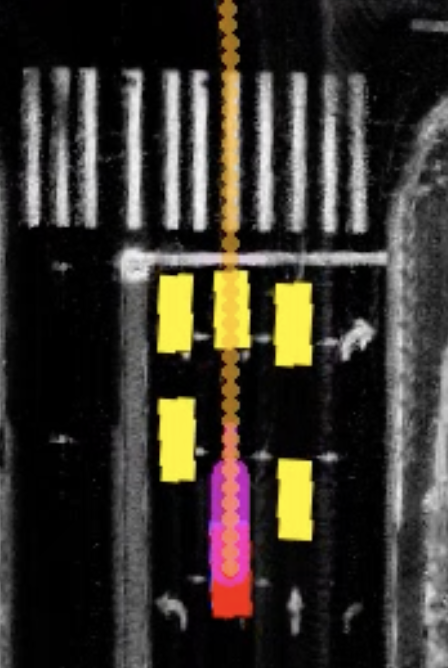}}
    \subfigure{\includegraphics[width=.23\columnwidth]{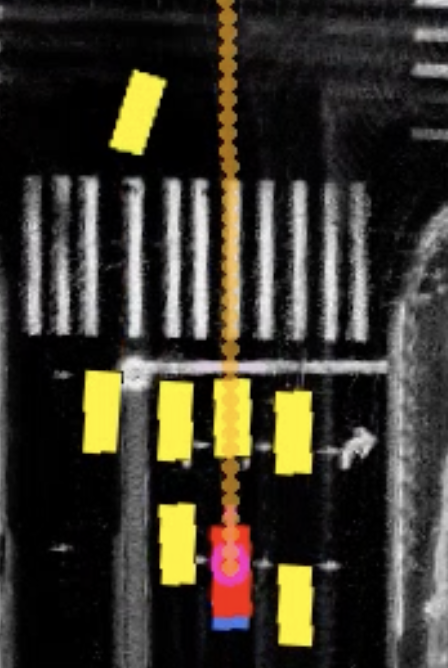}}
    \caption{The ego vehicle smoothly stops behind a vehicle. Video clip: sim-stop-for-vehicle1.mp4}
    \label{fig:sim-stop-for-vehicle1}
\end{figure*}

\begin{figure*}
    \centering
    \subfigure{\includegraphics[width=.23\columnwidth]{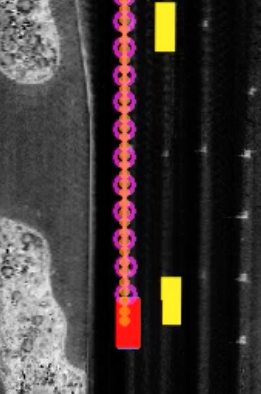}}
    \subfigure{\includegraphics[width=.23\columnwidth]{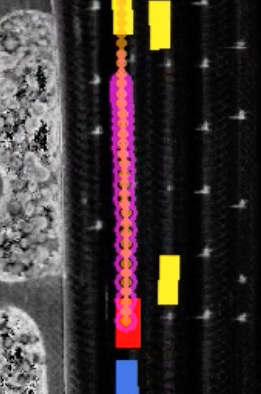}}
    \subfigure{\includegraphics[width=.23\columnwidth]{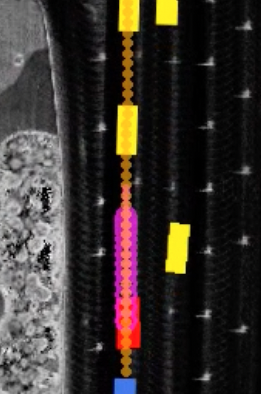}}
    \subfigure{\includegraphics[width=.23\columnwidth]{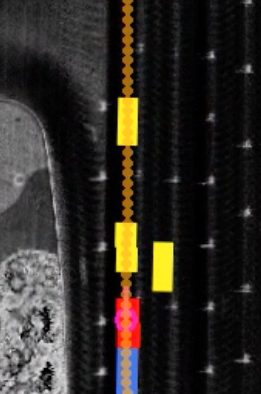}}
    \caption{The ego vehicle smoothly stops, from a higher velocity, behind a vehicle. Video clip: sim-stop-for-vehicle2.mp4}
    \label{fig:sim-stop-for-vehicle2}
\end{figure*}

\begin{figure*}
    \centering
    \subfigure{\includegraphics[width=.23\columnwidth]{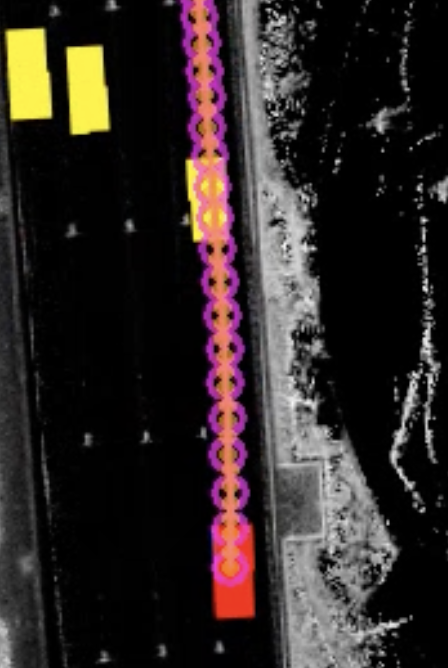}}
    \subfigure{\includegraphics[width=.23\columnwidth]{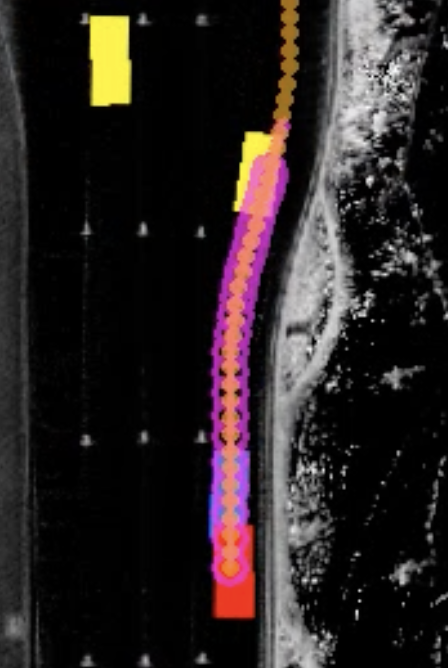}}
    \subfigure{ \includegraphics[width=.23\columnwidth]{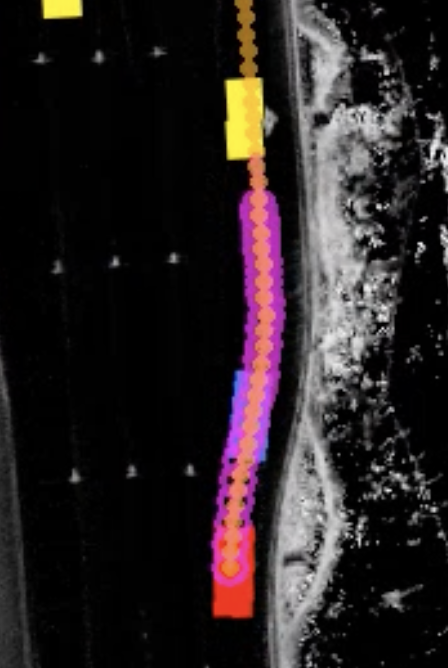}}
    \subfigure{\includegraphics[width=.23\columnwidth]{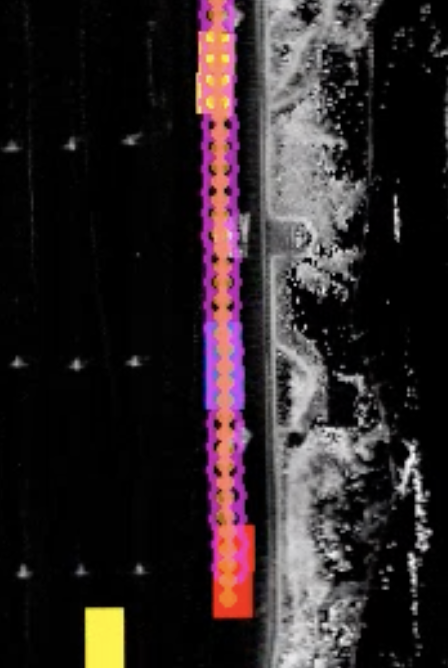}}
    \caption{The ego vehicle performs adaptive cruise control and slows down due to the behavior of the lead vehicle. Video clip: sim-acc1.mp4}
    \label{fig:sim-acc1}
\end{figure*}

\begin{figure*}
    \centering
    \subfigure{\includegraphics[width=.23\columnwidth]{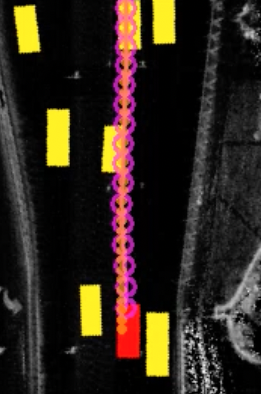}}
    \subfigure{\includegraphics[width=.23\columnwidth]{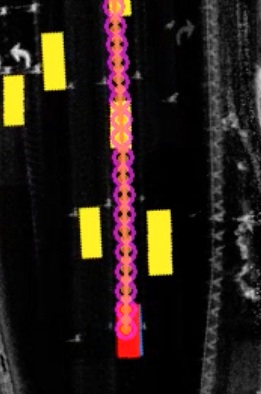}}
    \subfigure{\includegraphics[width=.23\columnwidth]{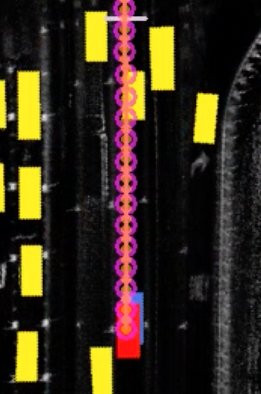}}
    \subfigure{\includegraphics[width=.23\columnwidth]{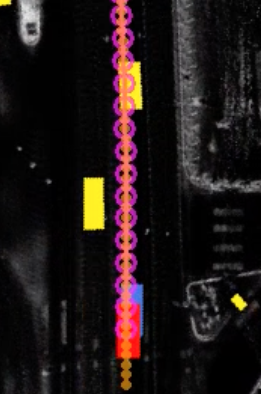}}
    \caption{The ego vehicle performs adaptive cruise control to maintain a safe distance from the lead vehicle. Video clip: sim-acc2.mp4}
    \label{fig:sim-acc2}
\end{figure*}

\begin{figure*}
    \centering
    \subfigure{\includegraphics[width=.23\columnwidth]{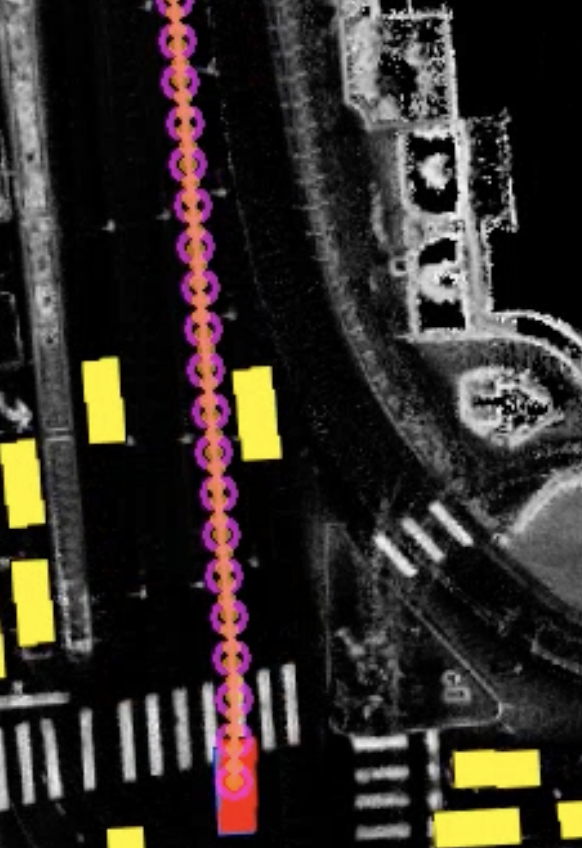}}
    \subfigure{\includegraphics[width=.23\columnwidth]{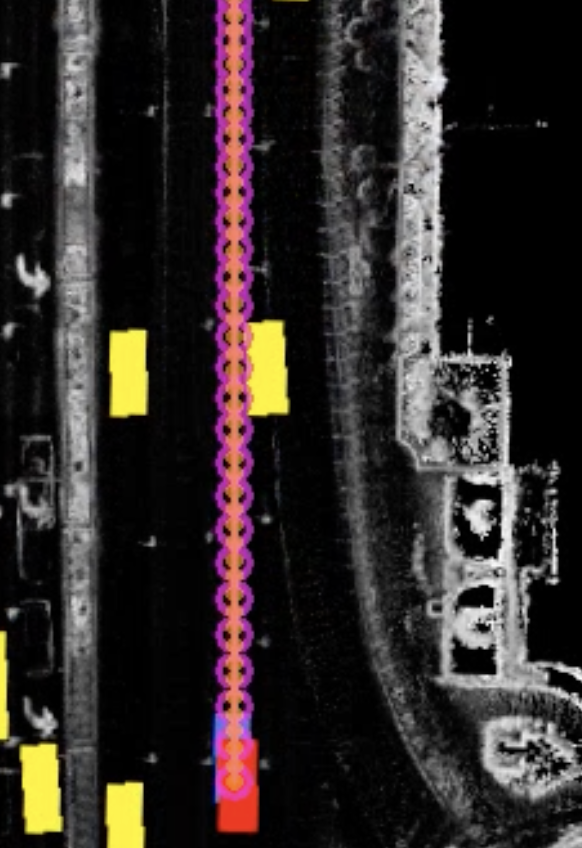}}
    \subfigure{\includegraphics[width=.23\columnwidth]{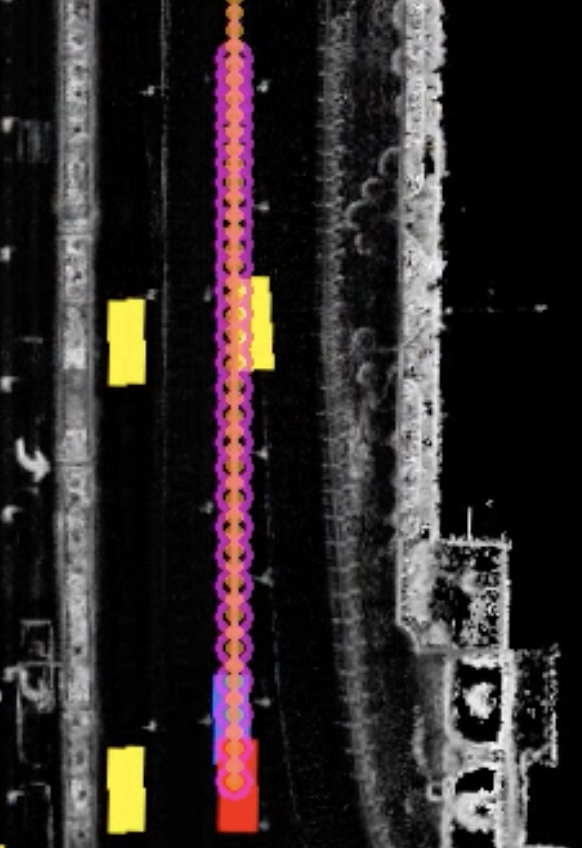}}
    \subfigure{\includegraphics[width=.23\columnwidth]{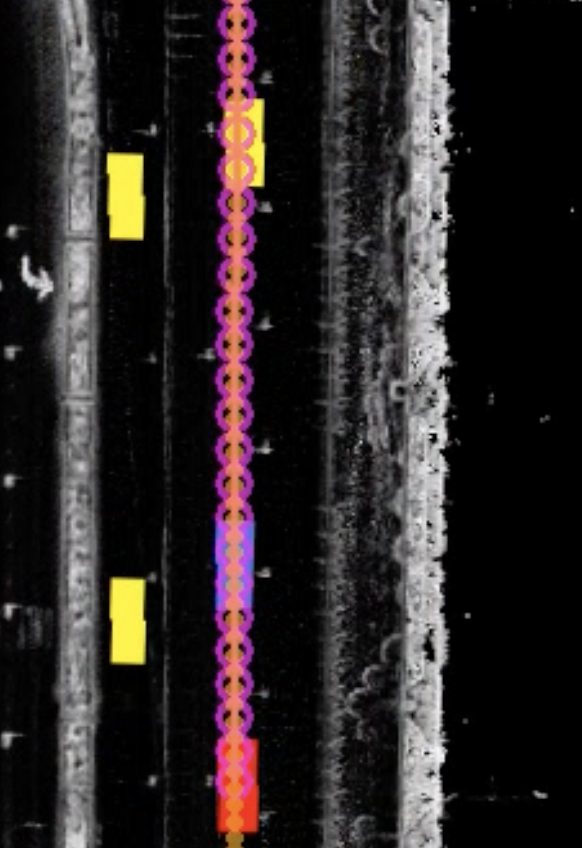}}
    \caption{The ego vehicle handles a moderate cut-in maneuver. Video clip: sim-cut-ins1.mp4}
    \label{fig:sim-cut-in1}
\end{figure*}

\begin{figure*}
    \centering
    \subfigure{\includegraphics[width=.23\columnwidth]{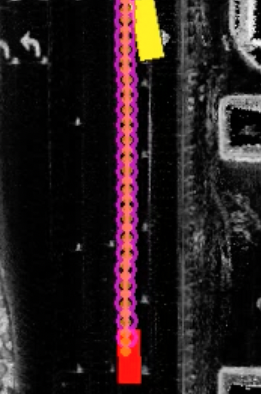}}
    \subfigure{\includegraphics[width=.23\columnwidth]{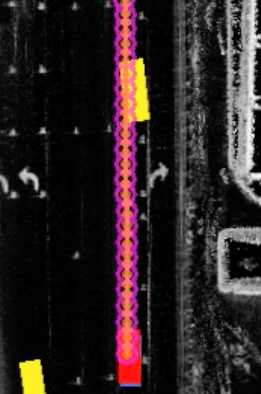}}
    \subfigure{\includegraphics[width=.23\columnwidth]{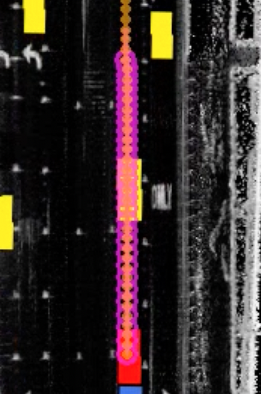}}
    \subfigure{\includegraphics[width=.23\columnwidth]{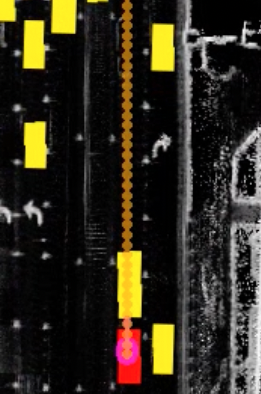}}
    \caption{The ego vehicle handles a moderate cut-in maneuver and stop behind the new lead vehicle. Video clip: sim-cut-ins2.mp4}
    \label{fig:sim-cut-in2}
\end{figure*}

\begin{figure*}
    \centering
    \subfigure{\includegraphics[width=.23\columnwidth]{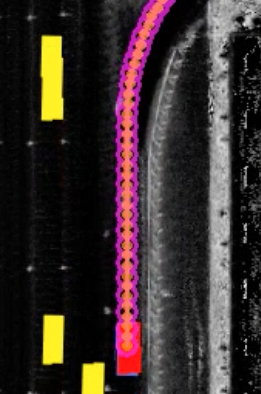}}
    \subfigure{\includegraphics[width=.23\columnwidth]{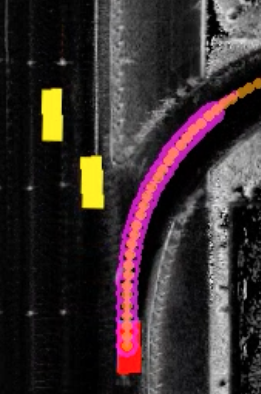}}
    \subfigure{\includegraphics[width=.23\columnwidth]{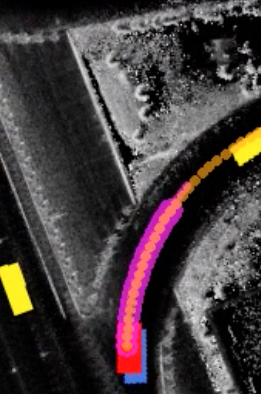}}
    \subfigure{\includegraphics[width=.23\columnwidth]{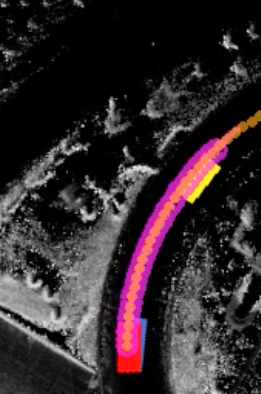}}
    \caption{The ego slows down and makes a right turn. Video clip: sim-turn1.mp4}
    \label{fig:sim-turn1}
\end{figure*}

\begin{figure*}
    \centering
    \subfigure{\includegraphics[width=.23\columnwidth]{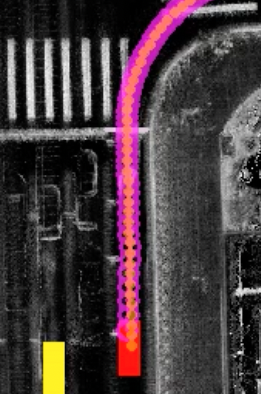}}
    \subfigure{\includegraphics[width=.23\columnwidth]{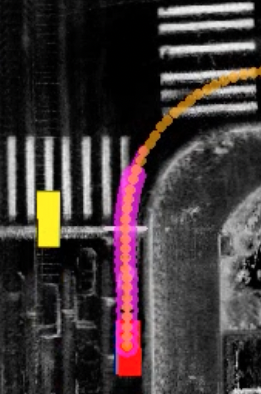}}
    \subfigure{\includegraphics[width=.23\columnwidth]{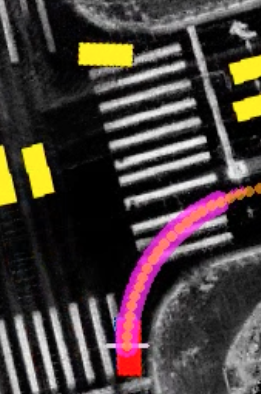}}
    \subfigure{\includegraphics[width=.23\columnwidth]{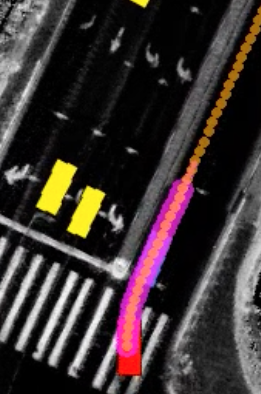}}
    \caption{The ego vehicle makes a $90^{\circ}$ right turn while maintaining a comfortable speed. Video clip: sim-turn2.mp4}
    \label{fig:sim-turn2}
\end{figure*}

Tab.~\ref{tab:sim-clips} presents a curated list of video clips where we use our \textit{Base} model across several simulated scenarios.
The ego is shown as a red rectangle, the replayed expert as the blue rectangle, and other (replayed) agents are yellow.
The planned trajectory of the ego for the current time step is shown in pink.
In particular, Fig.~\ref{fig:sim-start-from-stop1} and Fig.~\ref{fig:sim-start-from-stop2} show maneuvers where the ego vehicle is standing still and starts to move along side the vehicle ahead.
Fig.~\ref{fig:sim-stop-for-vehicle1} and Fig.~\ref{fig:sim-stop-for-vehicle2} show the ego vehicle slowing down and stopping behind a vehicle.
Adaptive Cruise Control (ACC) scenarios where the ego adjusts its speed to follow the lead vehicle are shown in Fig.~\ref{fig:sim-acc1} and Fig.~\ref{fig:sim-acc2}.
Fig~\ref{fig:sim-cut-in1} and Fig.~\ref{fig:sim-cut-in2} depict cut-in scenarios where the ego vehicle reacts and plan a shorter trajectory to avoid collisions, moreover, the ego adjusts its speeds or comes to a complete stop depending on the behavior of the new vehicle ahead.
Finally, Fig.~\ref{fig:sim-turn1} and Fig.~\ref{fig:sim-turn2} illustrate scenarios where the ego vehicle slows down to take a turn while making progress towards the goal.

\begin{table}[ht]
  \caption{Simulation video clips}
  \label{tab:sim-clips}
  \centering
  \begin{tabular}{llc}
    \toprule
    \multicolumn{1}{c}{Video clip} & \multicolumn{1}{c}{Snapshots} & Duration [s] \\
    \midrule
    \multicolumn{3}{l}{\textbf{Starting from a stop}} \\
    sim-start-from-stop1.mp4 & Figure~\ref{fig:sim-start-from-stop1} & 10 \\
    sim-start-from-stop2.mp4 & Figure~\ref{fig:sim-start-from-stop2} & 10 \\
    \multicolumn{3}{l}{\textbf{Stopping behind a vehicle}} \\
    sim-stop-for-vehicle1.mp4 & Figure~\ref{fig:sim-stop-for-vehicle1} & 10 \\
    sim-stop-for-vehicle2.mp4 & Figure~\ref{fig:sim-stop-for-vehicle2} & 10 \\
    \multicolumn{3}{l}{\textbf{Adaptive cruise control}} \\
    sim-acc1.mp4 & Figure~\ref{fig:sim-acc1} & 10 \\
    sim-acc2.mp4 & Figure~\ref{fig:sim-acc2} & 10 \\
    \multicolumn{3}{l}{\textbf{Moderate cut-ins}} \\
    sim-cut-ins1.mp4 & Figure~\ref{fig:sim-cut-in1} & 10 \\
    sim-cut-ins2.mp4 & Figure~\ref{fig:sim-cut-in2} & 10 \\
    \multicolumn{3}{l}{\textbf{Turning}} \\
    sim-turn1.mp4 & Figure~\ref{fig:sim-turn1} & 10 \\
    sim-turn2.mp4 & Figure~\ref{fig:sim-turn2} & 10 \\
    \bottomrule
  \end{tabular}
\end{table}


\subsection{Real-world driving video clips}
\label{app:real-clips}
Tab.~\ref{tab:real-clips} provides a curated list of real-world driving video clips collected during our testing in Las Vegas Strip and nearby areas. 
The description for each video clip is included in the corresponding snapshots figure caption.

\begin{table}
  \caption{Real-world driving video clips}
  \label{tab:real-clips}
  \centering
  \begin{tabular}{llc}
    \toprule
    \multicolumn{1}{c}{Video clip} & \multicolumn{1}{c}{Snapshots} & Duration [s] \\
    \midrule
    \multicolumn{3}{l}{\textbf{Driving with a vehicle far ahead}} \\
    far-ahead-01.mp4 & Figure~\ref{fig:far-ahead-01} & 34 \\
    far-ahead-02.mp4 & Figure~\ref{fig:far-ahead-02} & 25 \\
    far-ahead-03.mp4 & Figure~\ref{fig:far-ahead-03} & 20 \\
    far-ahead-04.mp4 & Figure~\ref{fig:far-ahead-04} & 30 \\
    \multicolumn{3}{l}{\textbf{Driving with a vehicle close ahead}} \\
    close-ahead-01.mp4 & Figure~\ref{fig:close-ahead-01} & 40 \\
    close-ahead-02.mp4 & Figure~\ref{fig:close-ahead-02} & 30 \\
    close-ahead-03.mp4 & Figure~\ref{fig:close-ahead-03} & 30 \\
    \multicolumn{3}{l}{\textbf{Passenger pickup/dropoff zones}} \\
    pudo-01.mp4 & Figure~\ref{fig:pudo-01} & 30 \\
    pudo-02.mp4 & Figure~\ref{fig:pudo-02} & 26 \\
    \multicolumn{3}{l}{\textbf{Stoping behind a vehicle}} \\
    stopping-01.mp4 & Figure~\ref{fig:stopping-01} & 15 \\
    stopping-02.mp4 & Figure~\ref{fig:stopping-02} & 15 \\
    stopping-03.mp4 & Figure~\ref{fig:stopping-03} & 15 \\
    \multicolumn{3}{l}{\textbf{Cut-ins}} \\
    cut-in-01.mp4 & Figure~\ref{fig:cut-in-01} & 15 \\
    cut-in-02.mp4 & Figure~\ref{fig:cut-in-02} & 20 \\
    cut-in-03.mp4 & Figure~\ref{fig:cut-in-03} & 32 \\
    \bottomrule
  \end{tabular}
\end{table}

\begin{figure*}
  \centering
  \subfigure{\includegraphics[width=.325\columnwidth]{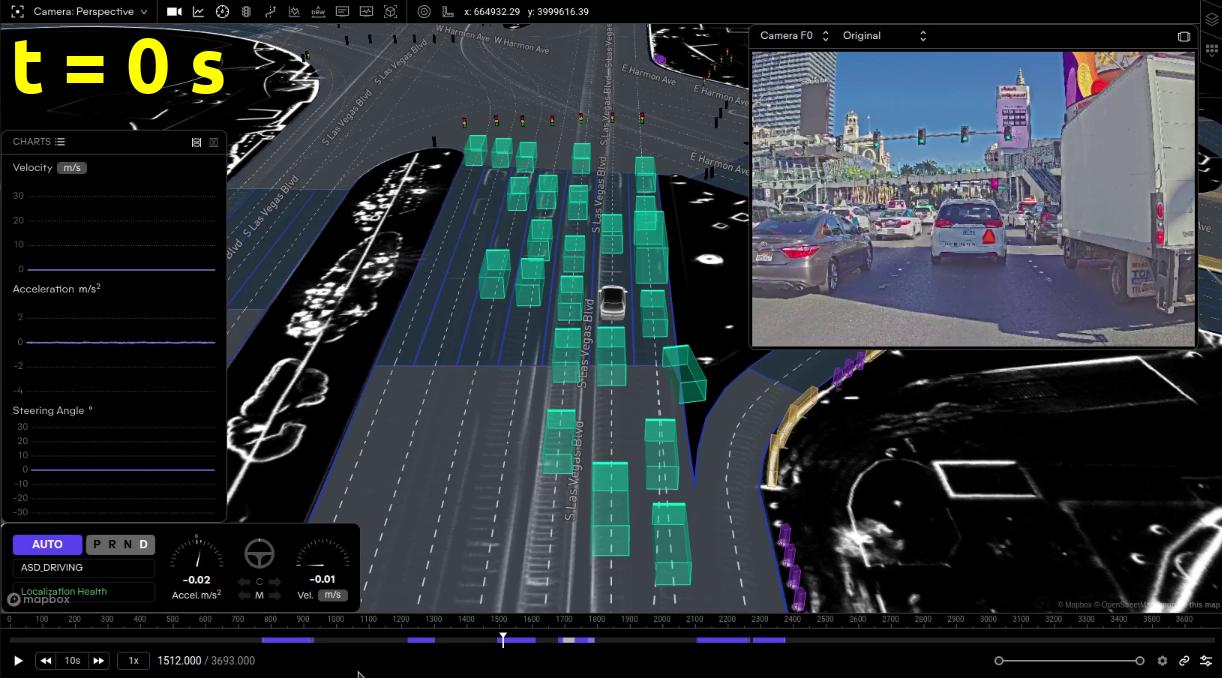}}
  \subfigure{\includegraphics[width=.325\columnwidth]{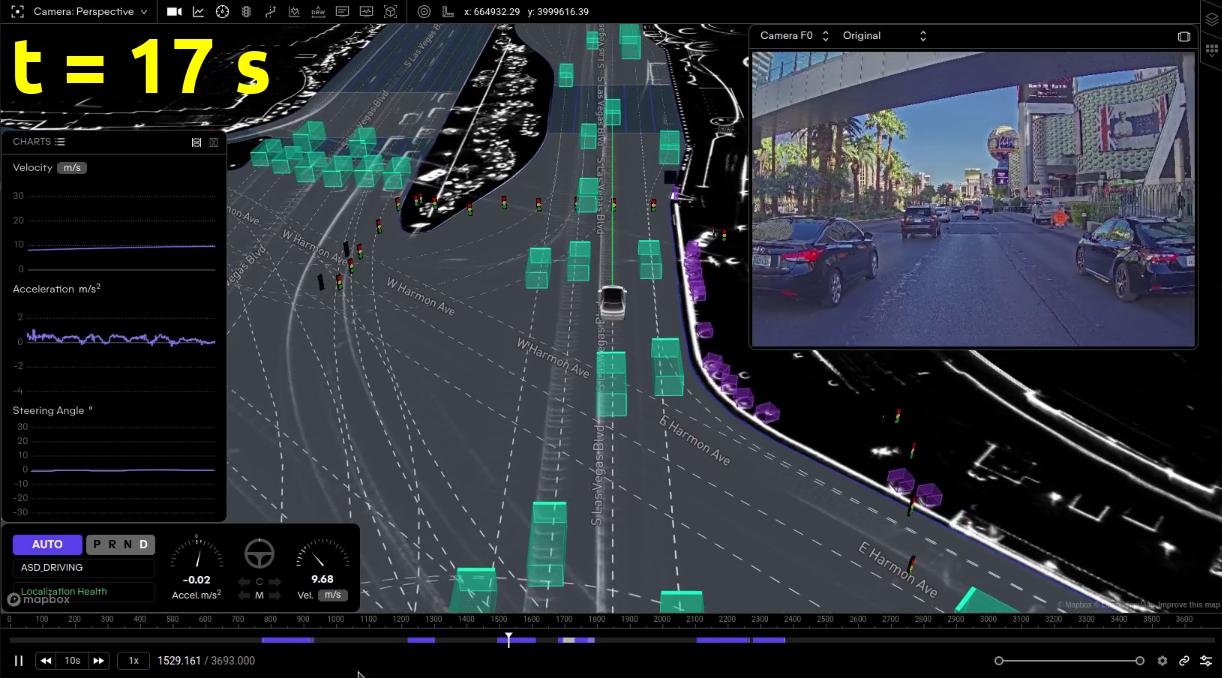}}
  \subfigure{\includegraphics[width=.325\columnwidth]{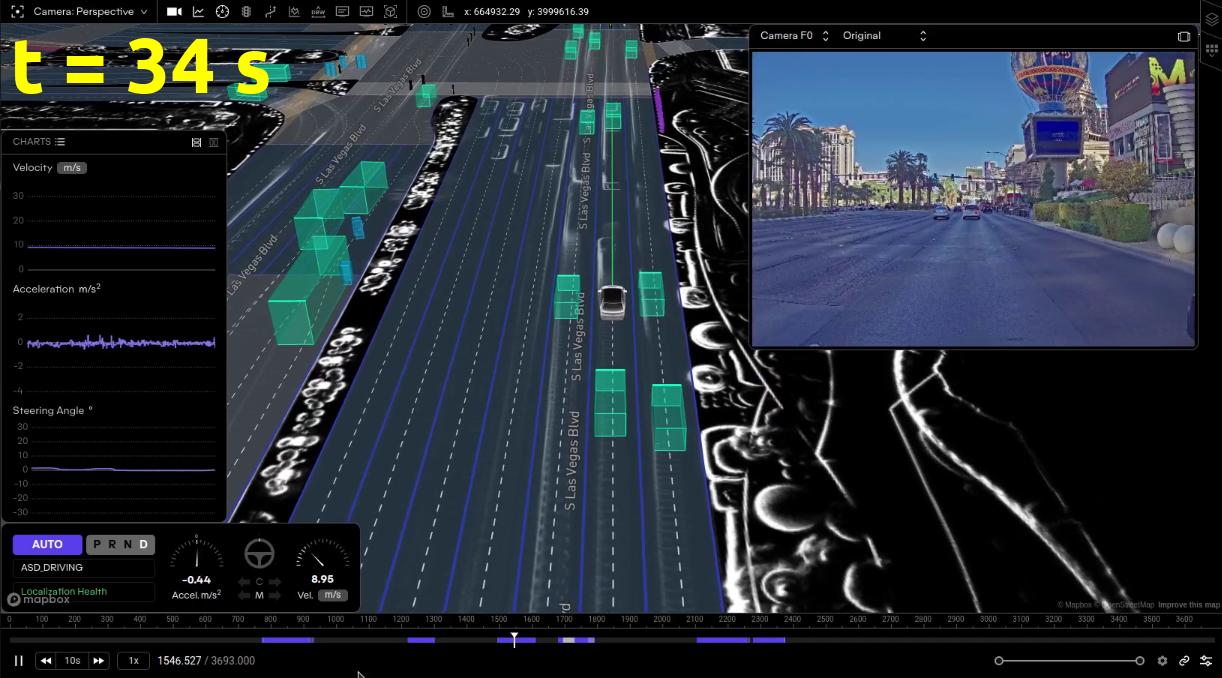}}
  \caption{The ego vehicle follows dense traffic as it starts to flow after a transition from red to green traffic light. Video clip: far-ahead-01.mp4}
  \label{fig:far-ahead-01}
\end{figure*}

\begin{figure*}
  \centering
  \subfigure{\includegraphics[width=.325\columnwidth]{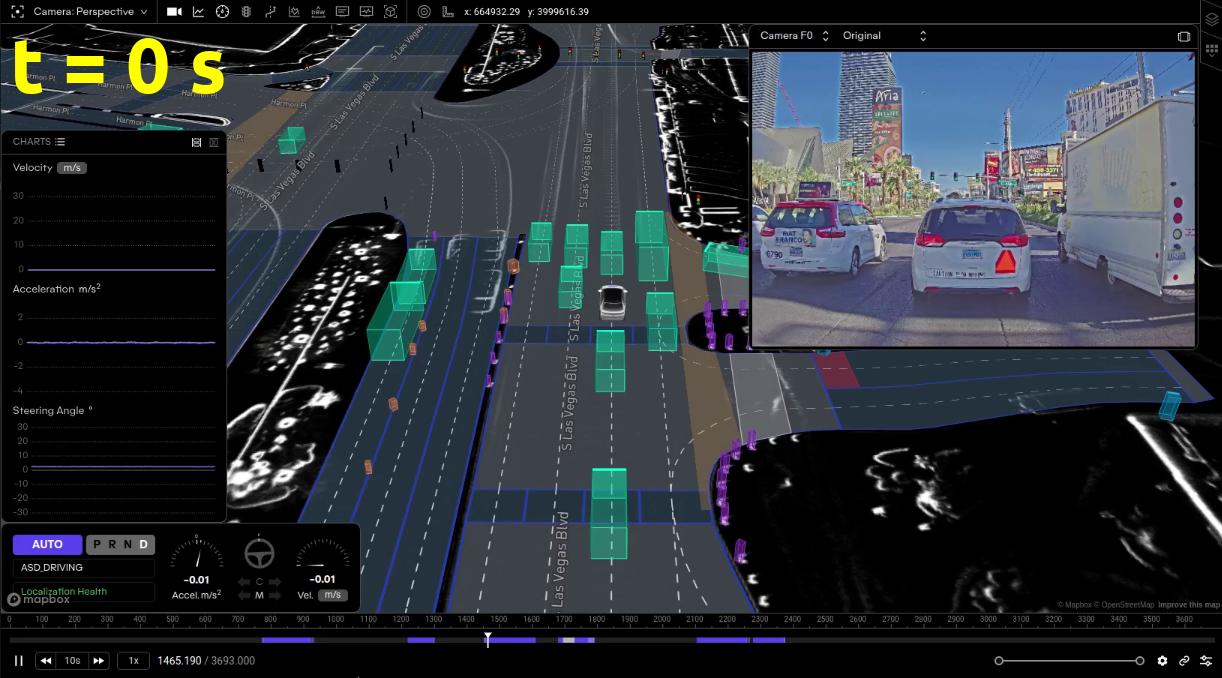}}
  \subfigure{\includegraphics[width=.325\columnwidth]{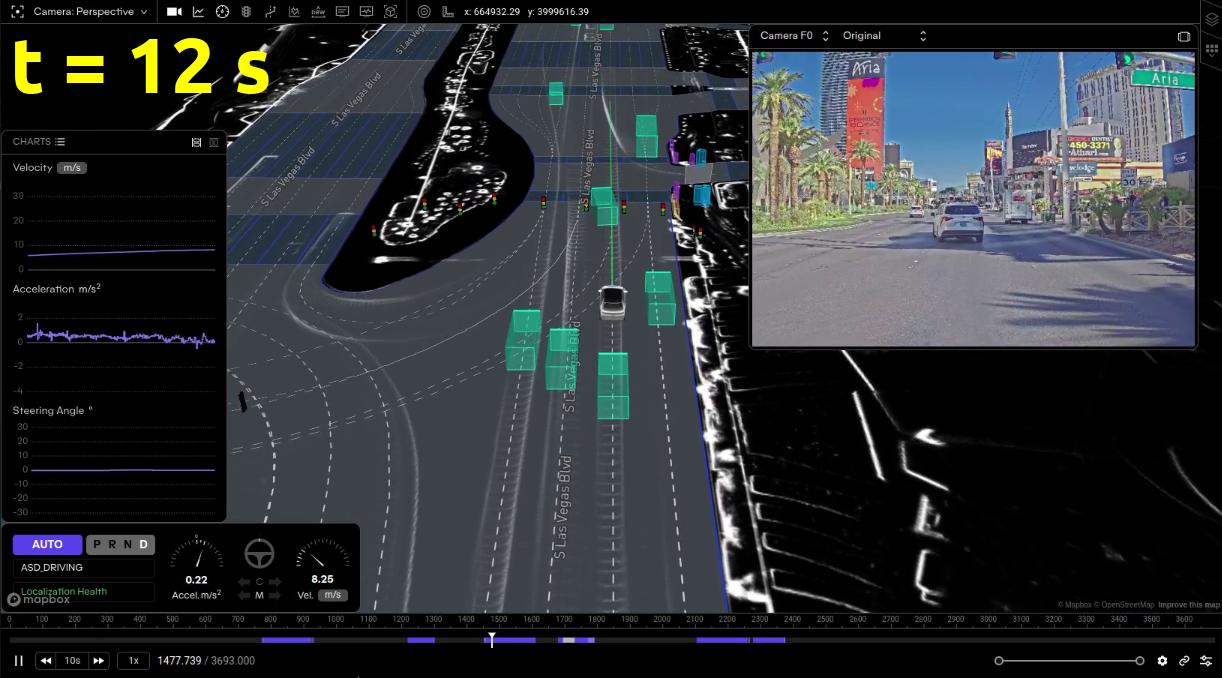}}
  \subfigure{\includegraphics[width=.325\columnwidth]{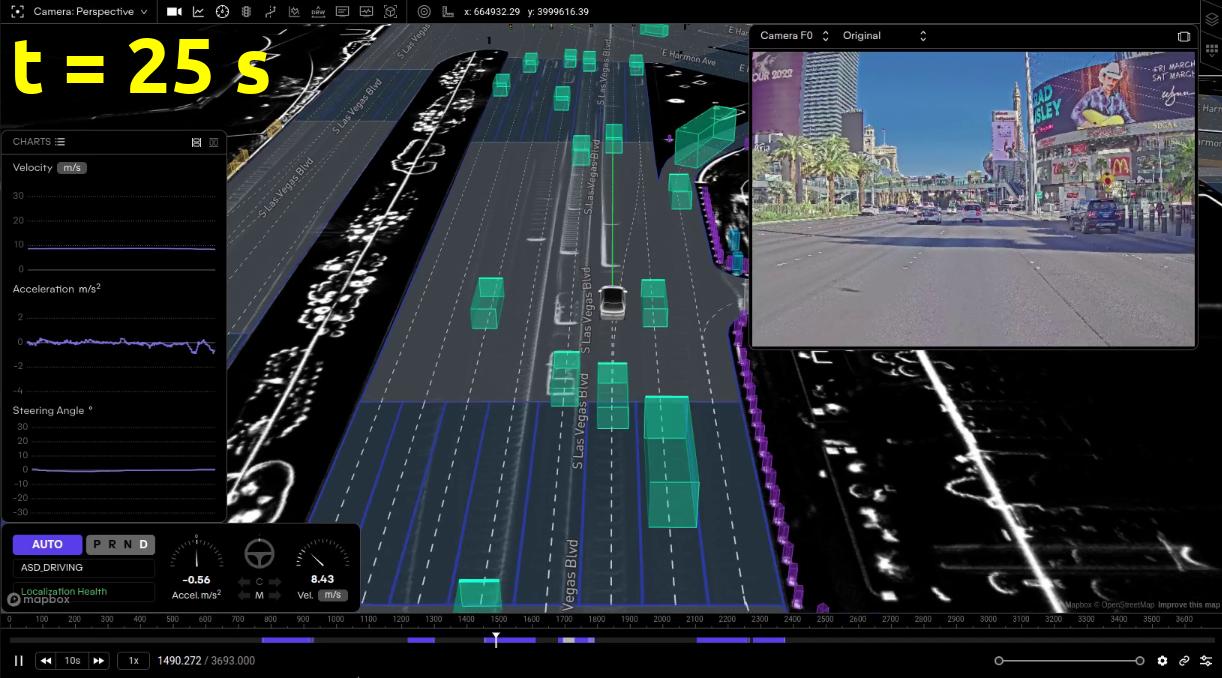}}
  \caption{The ego vehicle follows traffic as it starts to flow after a transition from red to green traffic light. Video clip: far-ahead-02.mp4}
  \label{fig:far-ahead-02}
\end{figure*}

\begin{figure*}
  \centering
  \subfigure{\includegraphics[width=.325\columnwidth]{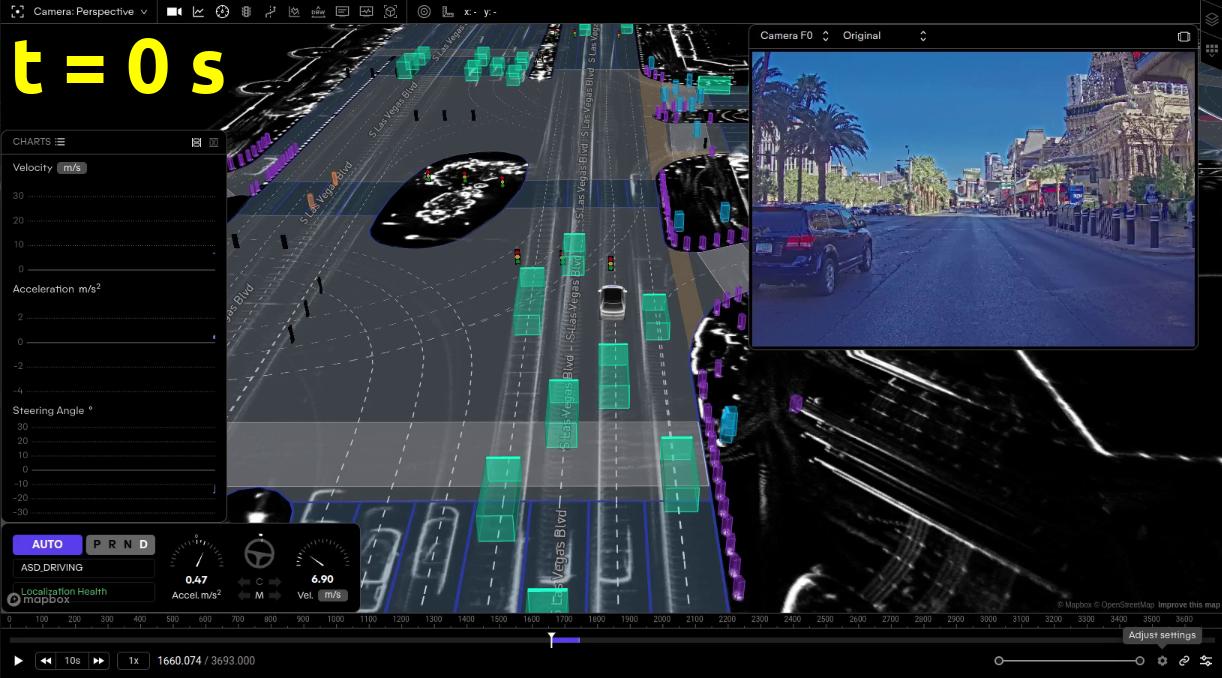}}
  \subfigure{\includegraphics[width=.325\columnwidth]{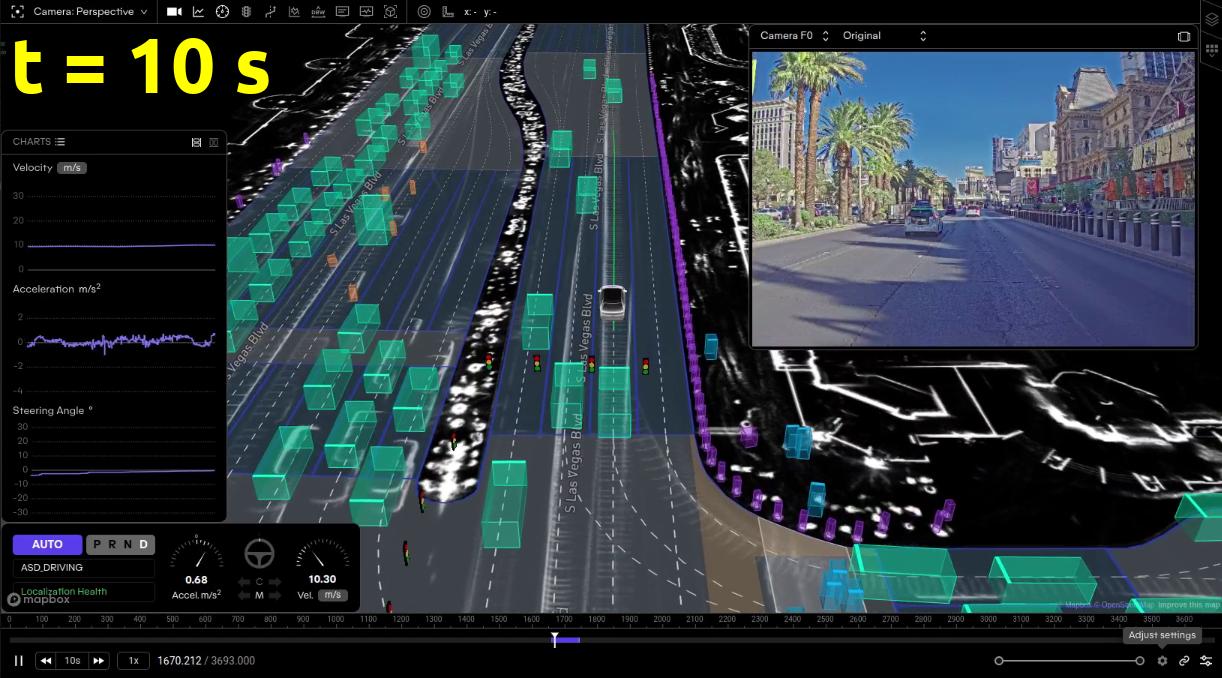}}
  \subfigure{\includegraphics[width=.325\columnwidth]{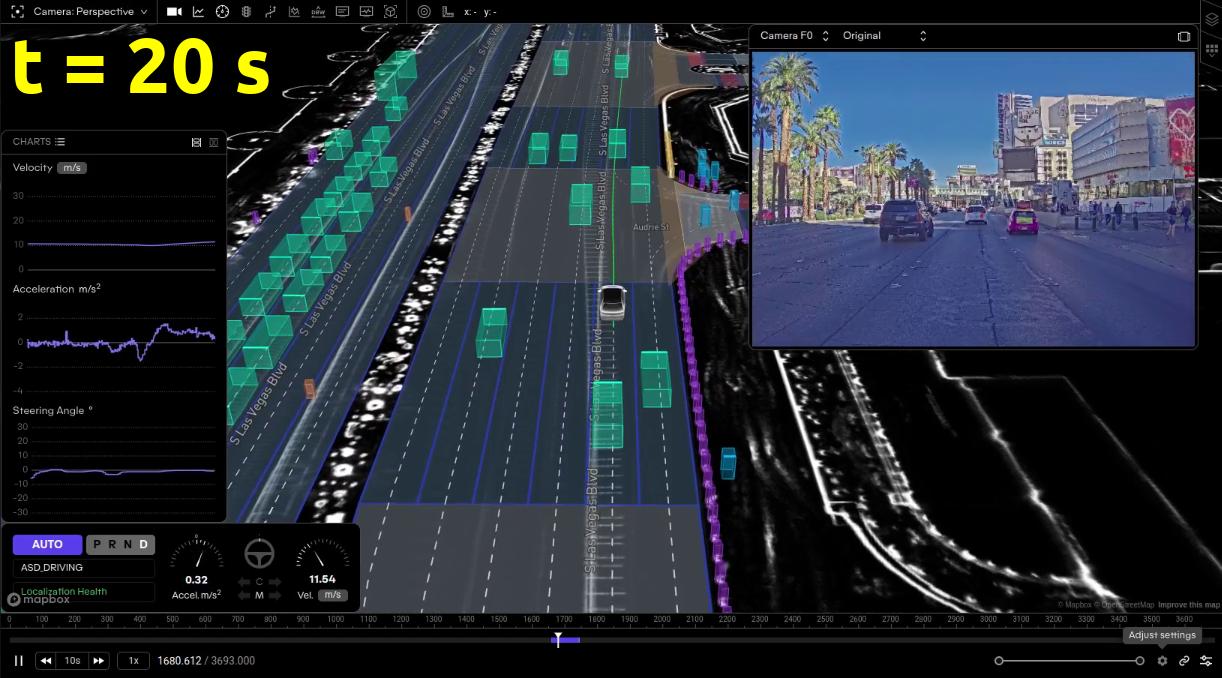}}
  \caption{The ego vehicle drives surrounded by few vehicles. Video clip: far-ahead-03.mp4}
  \label{fig:far-ahead-03}
\end{figure*}

\begin{figure*}
  \centering
  \subfigure{\includegraphics[width=.325\columnwidth]{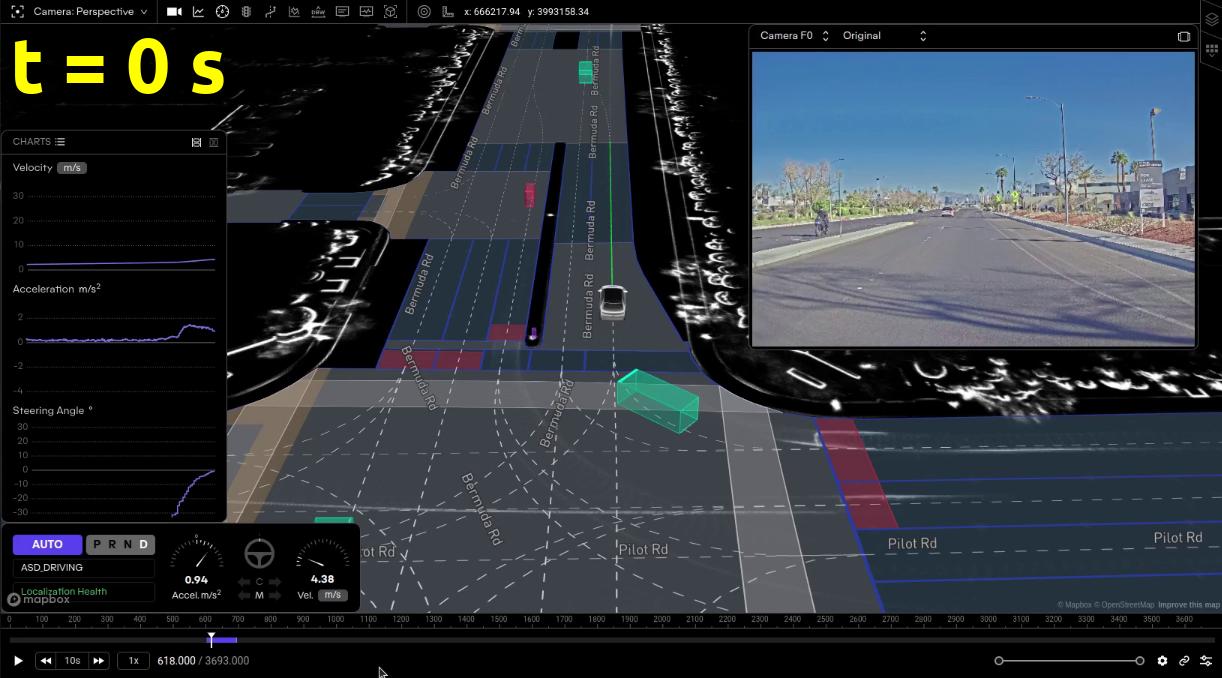}}
  \subfigure{\includegraphics[width=.325\columnwidth]{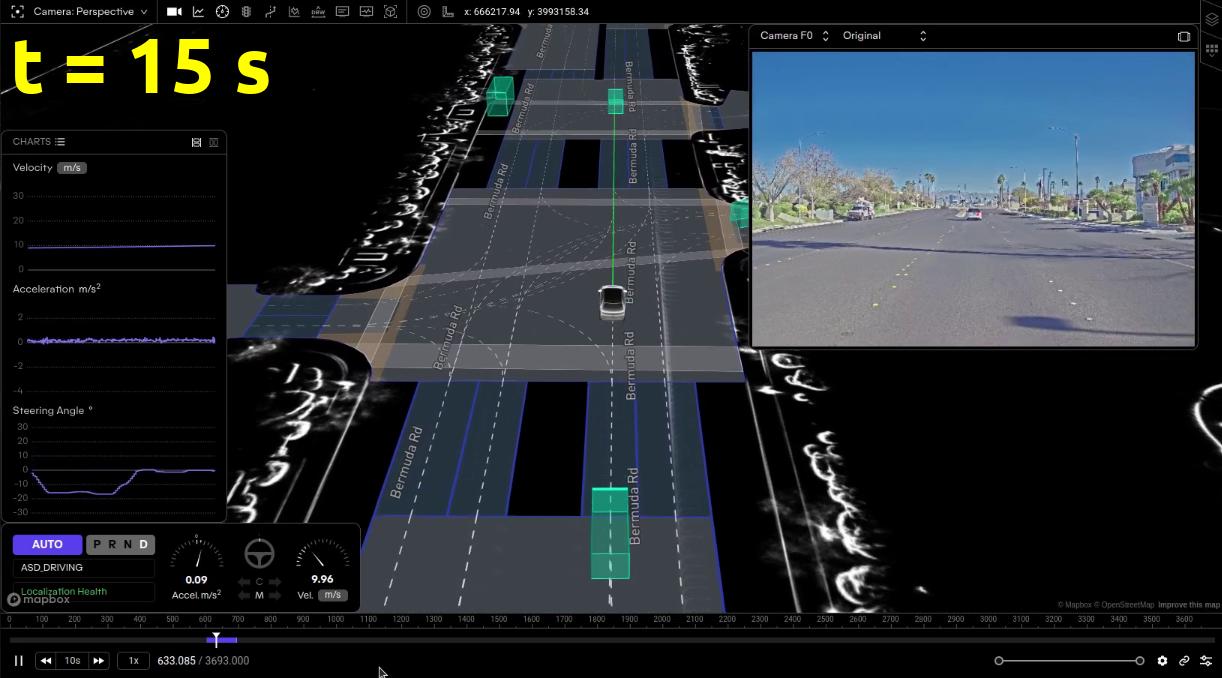}}
  \subfigure{\includegraphics[width=.325\columnwidth]{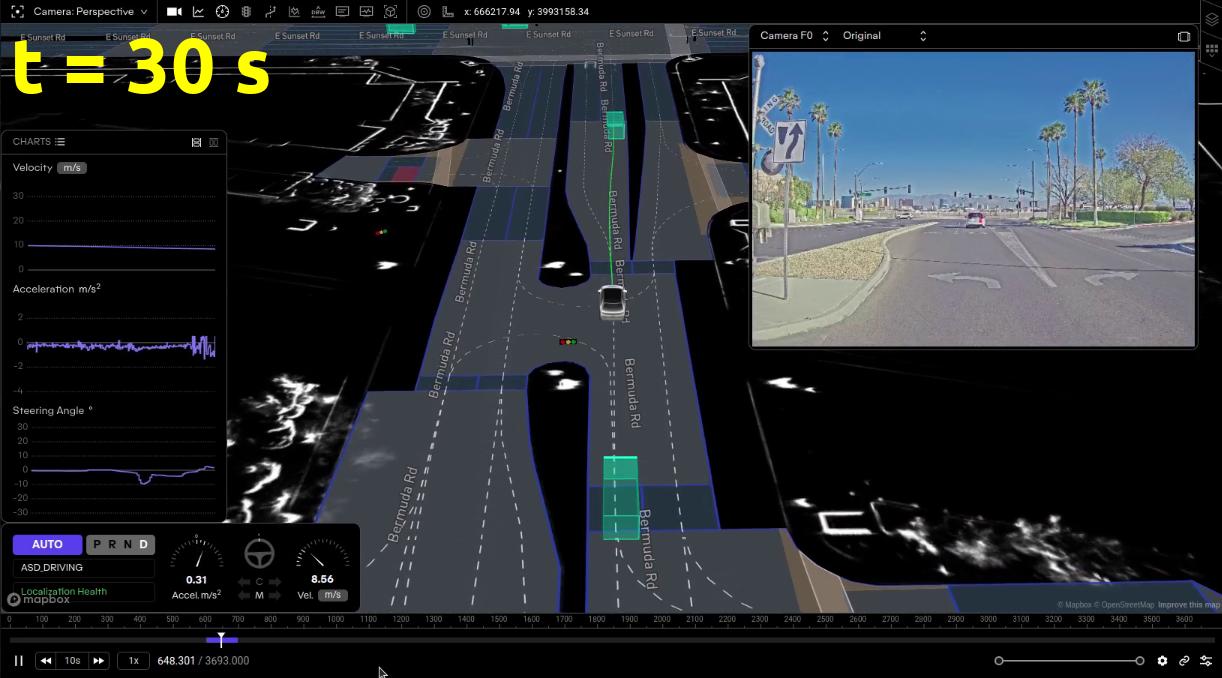}}
  \caption{The ego vehicle performs a smooth lane-change. Video clip: far-ahead-04.mp4}
  \label{fig:far-ahead-04}
\end{figure*}

\begin{figure*}
  \centering
  \subfigure{\includegraphics[width=.325\columnwidth]{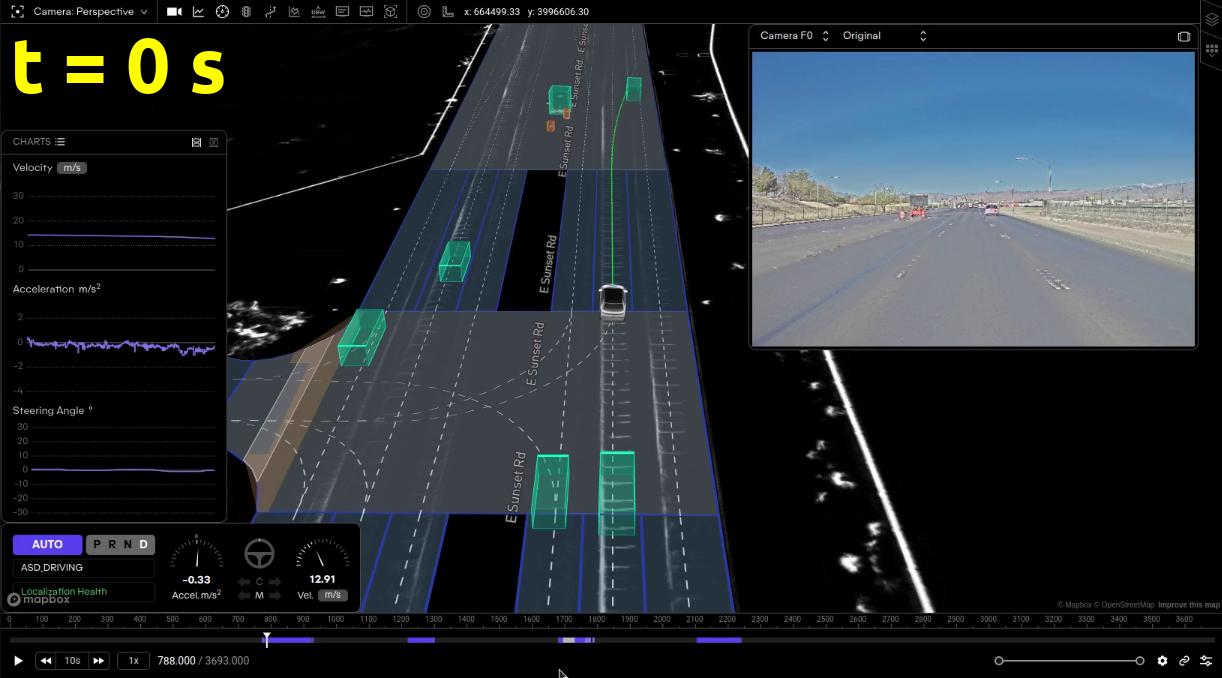}}
  \subfigure{\includegraphics[width=.325\columnwidth]{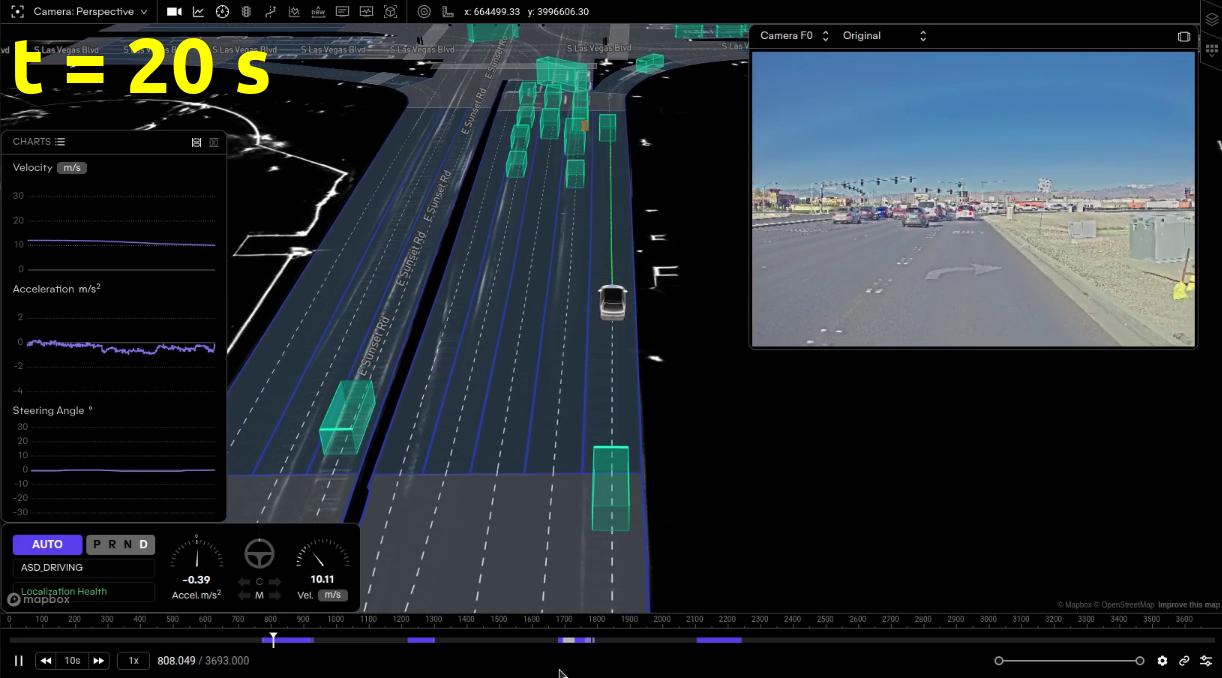}}
  \subfigure{\includegraphics[width=.325\columnwidth]{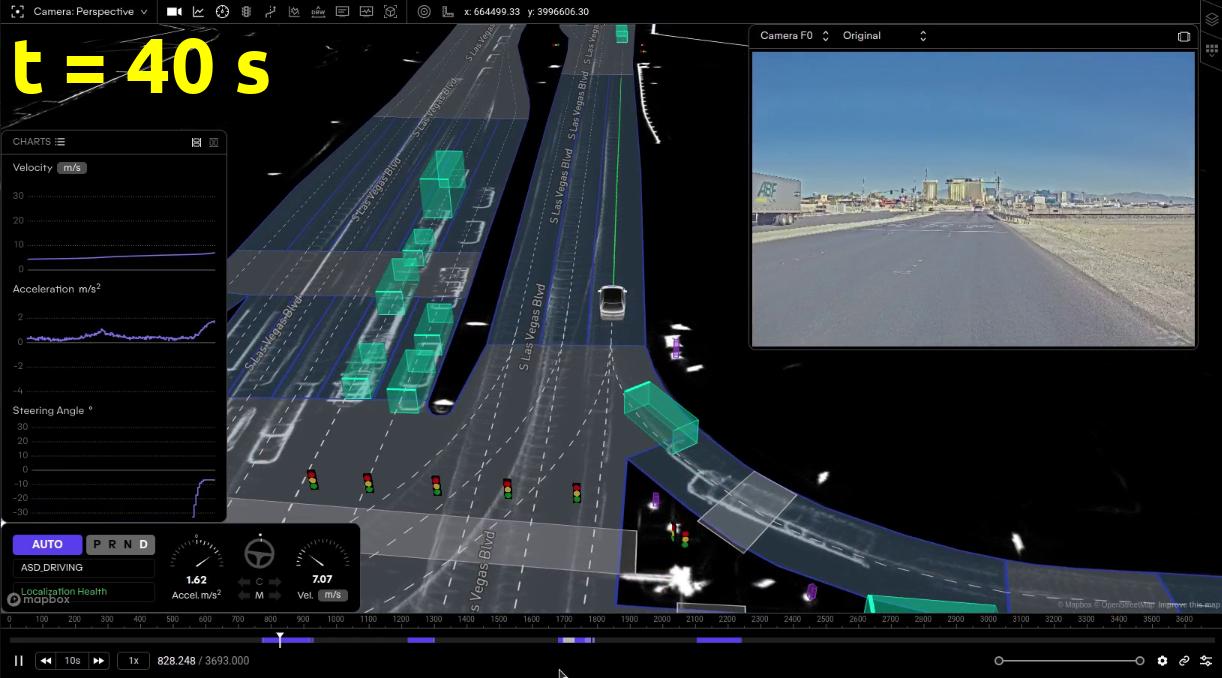}}
  \caption{The ego vehicle drives with a vehicle close ahead, performs a smooth lane-change and merges into a perpendicular road. Video clip: close-ahead-01.mp4}
  \label{fig:close-ahead-01}
\end{figure*}

\begin{figure*}
  \centering
  \subfigure{\includegraphics[width=.325\columnwidth]{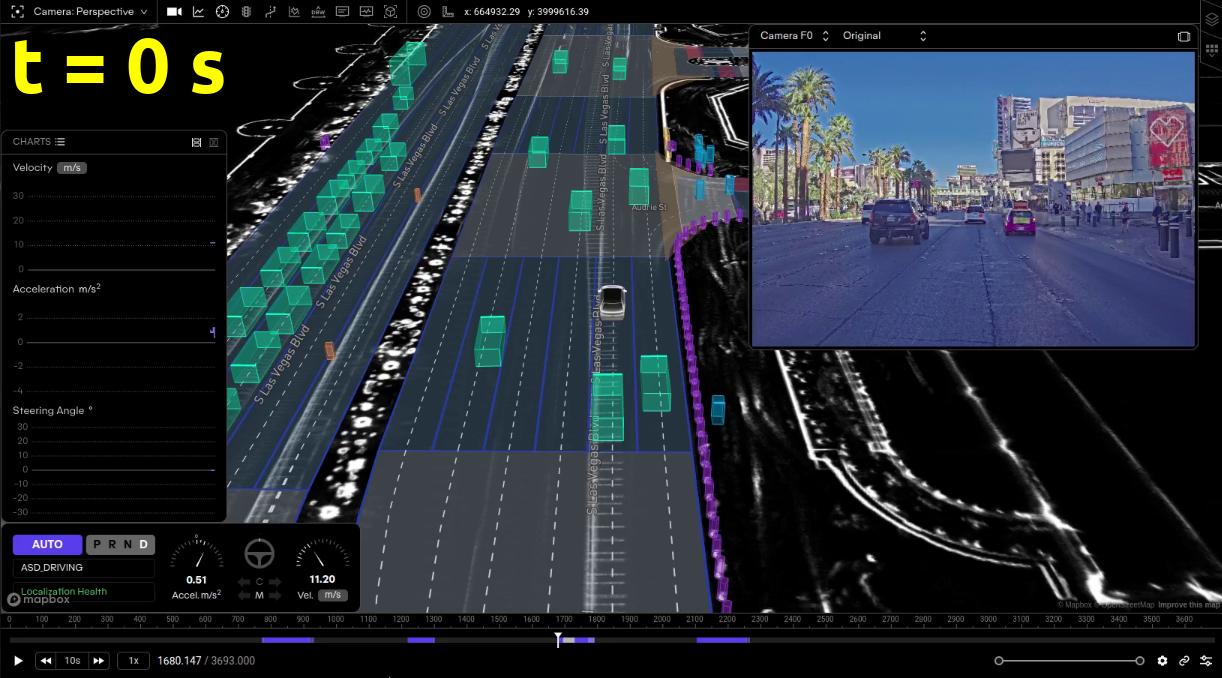}}
  \subfigure{\includegraphics[width=.325\columnwidth]{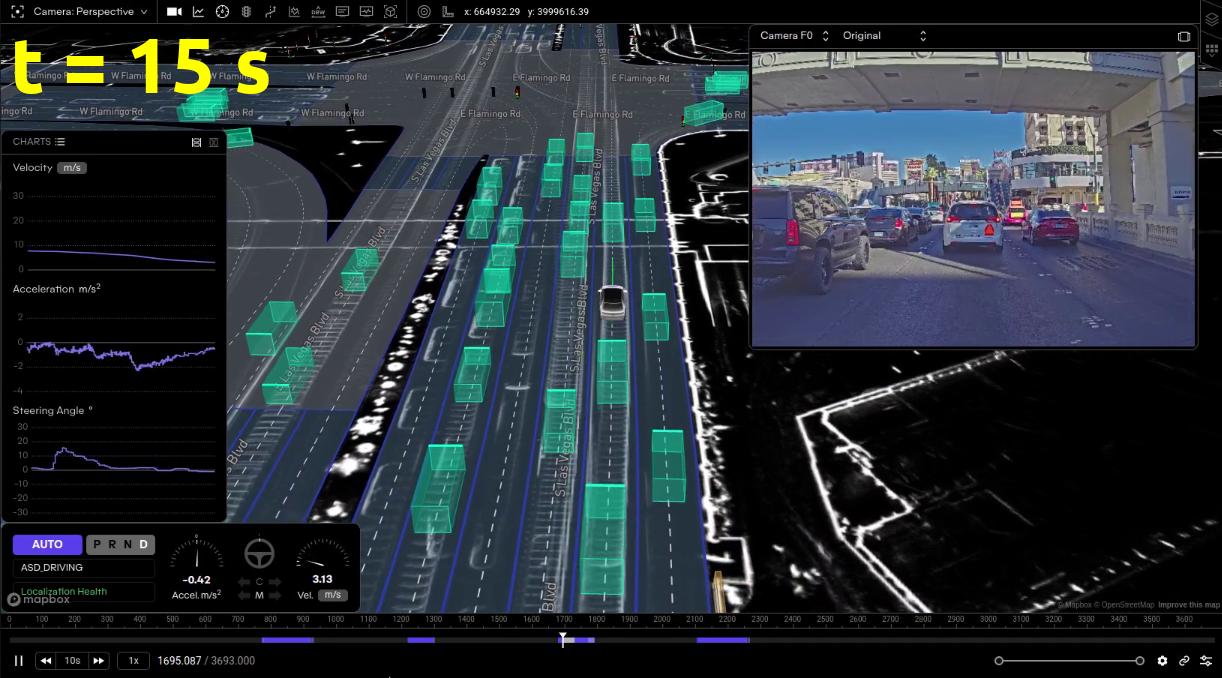}}
  \subfigure{\includegraphics[width=.325\columnwidth]{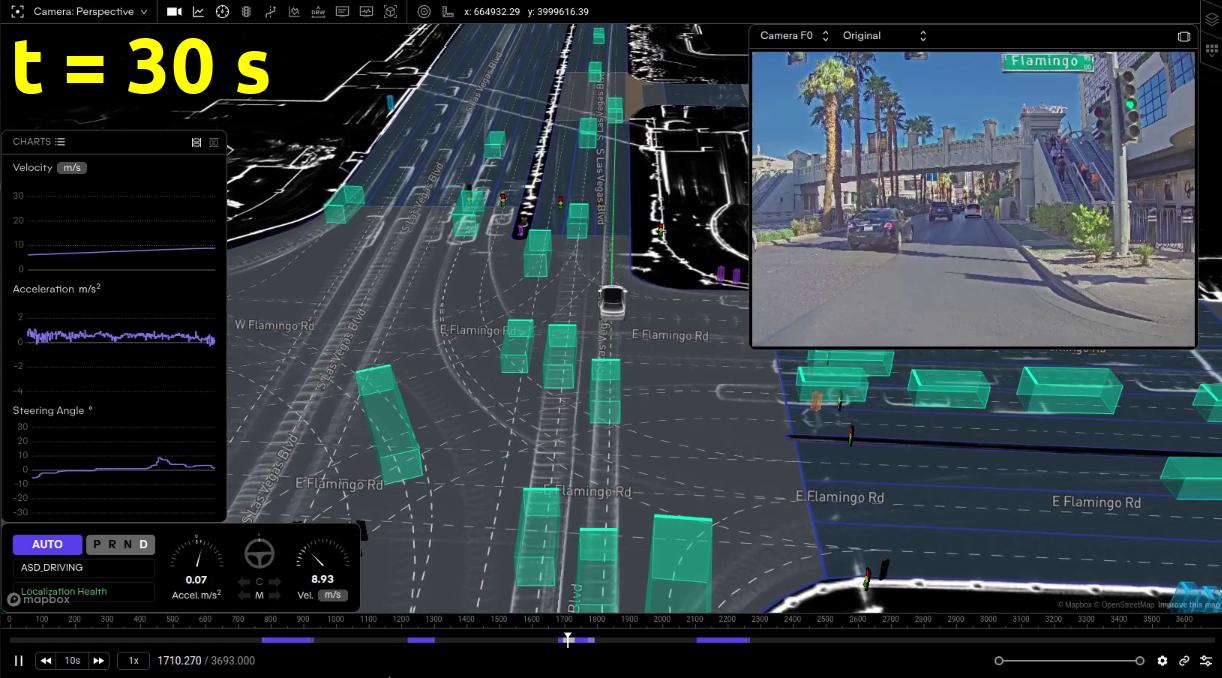}}
  \caption{The ego vehicle slows down and speeds up to match the lead vehicle speed. Video clip: close-ahead-02.mp4}
  \label{fig:close-ahead-02}
\end{figure*}

\begin{figure*}
  \centering
  \subfigure{\includegraphics[width=.325\columnwidth]{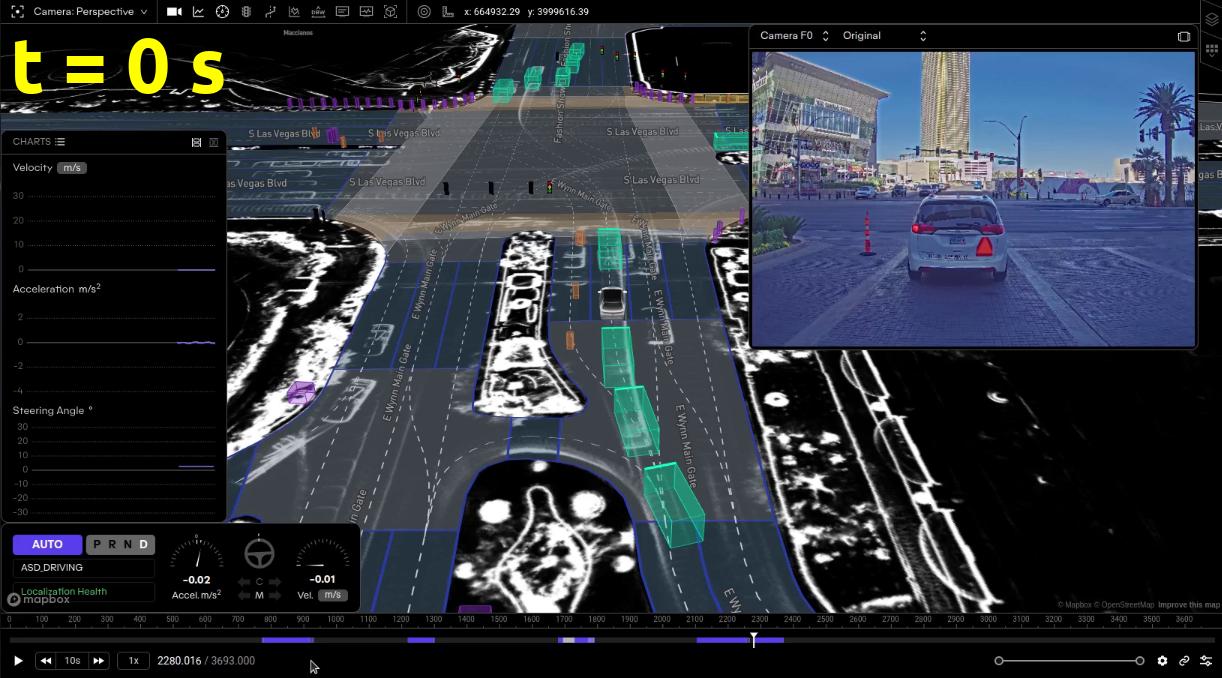}}
  \subfigure{\includegraphics[width=.325\columnwidth]{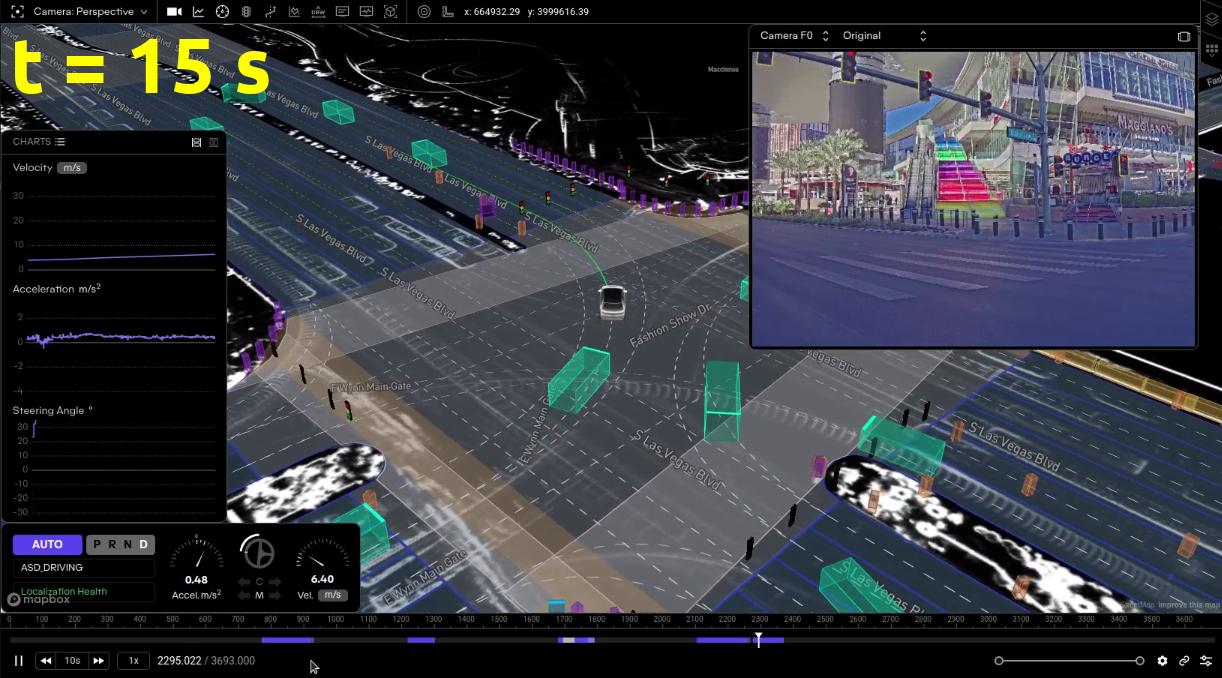}}
  \subfigure{\includegraphics[width=.325\columnwidth]{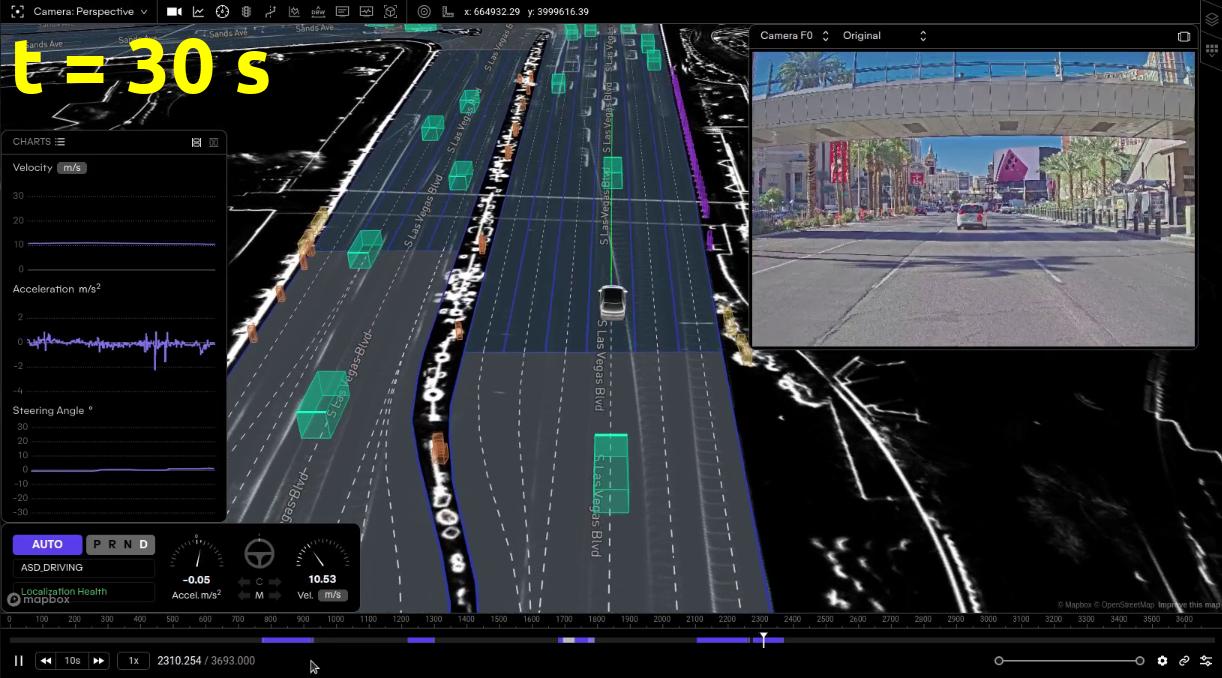}}
  \caption{The ego vehicle makes a protected left turn following the green traffic light. Video clip: close-ahead-03.mp4}
  \label{fig:close-ahead-03}
\end{figure*}

\begin{figure*}
  \centering
  \subfigure{\includegraphics[width=.325\columnwidth]{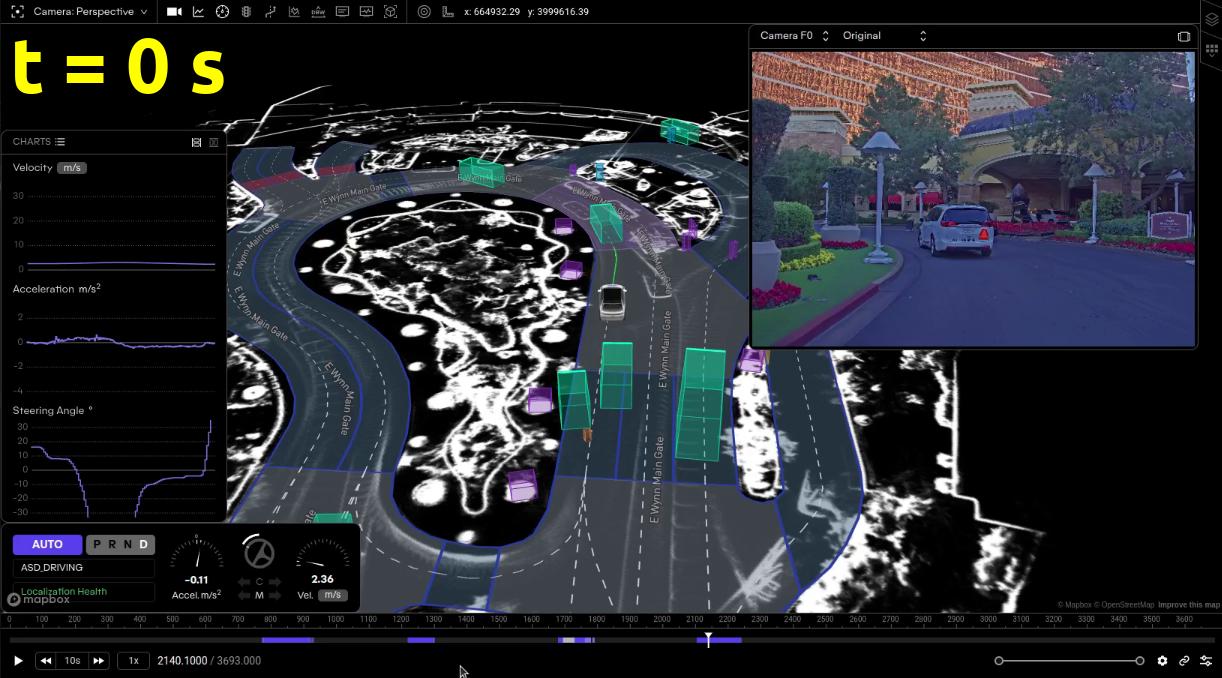}}
  \subfigure{\includegraphics[width=.325\columnwidth]{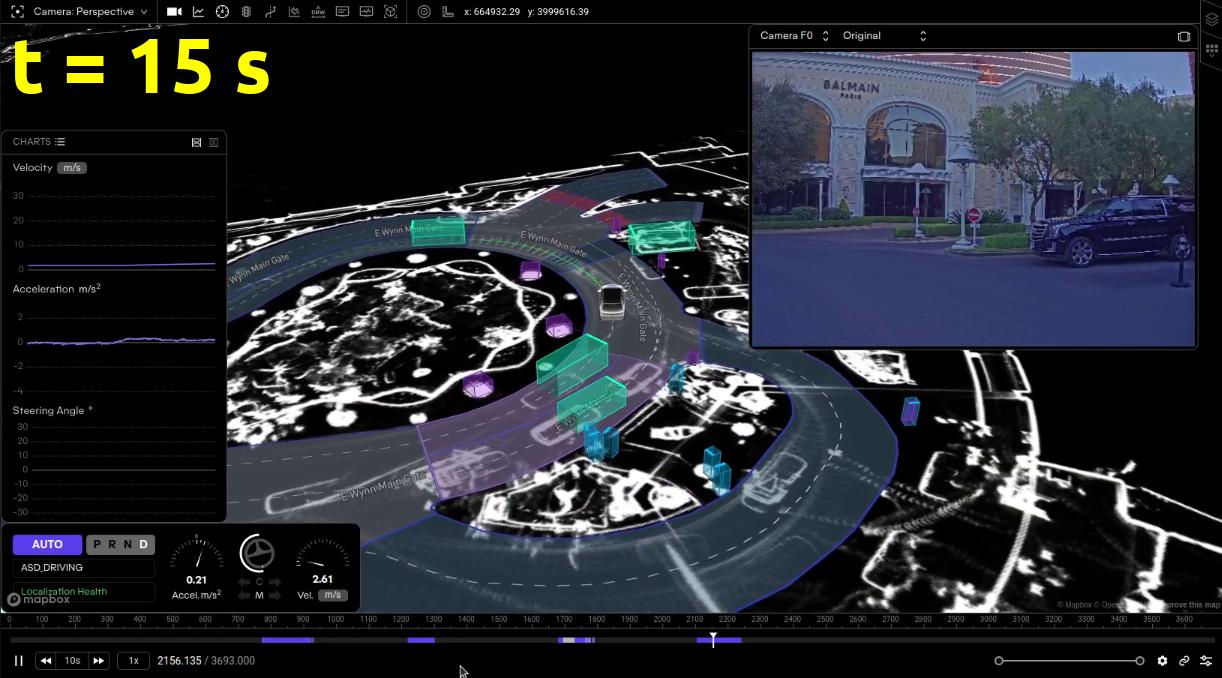}}
  \subfigure{\includegraphics[width=.325\columnwidth]{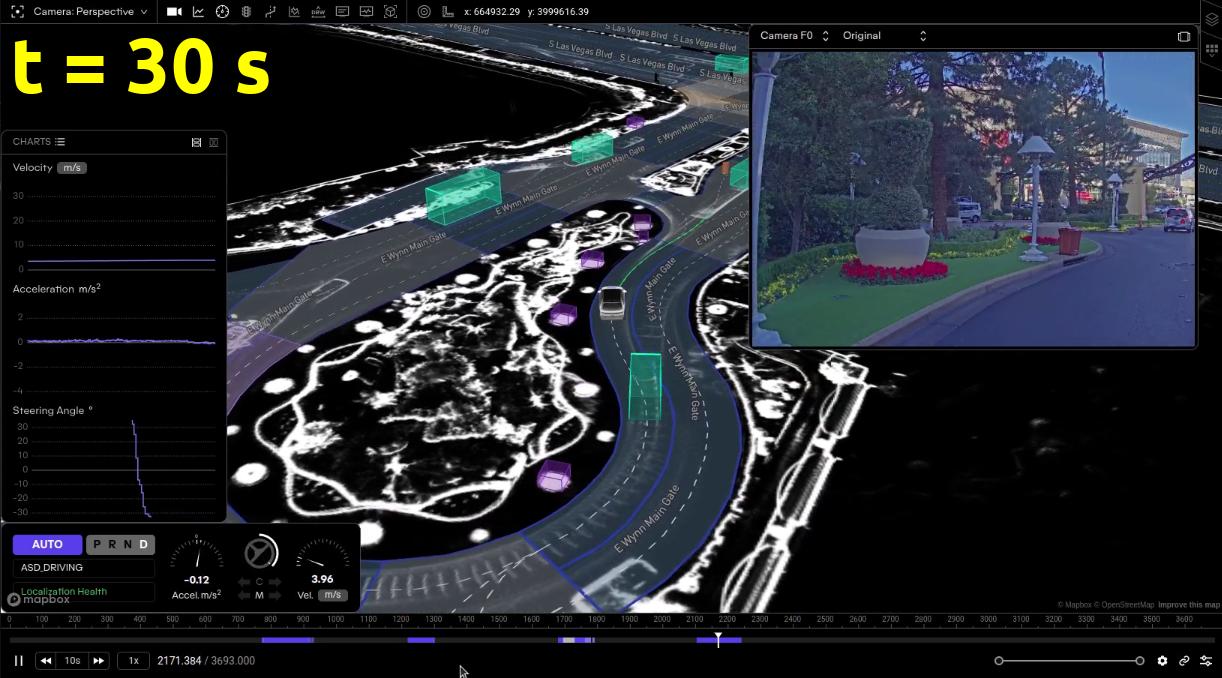}}
  \caption{The ego vehicle drives through a passenger pickup/dropoff zone. Video clip: pudo-01.mp4}
  \label{fig:pudo-01}
\end{figure*}

\begin{figure*}
  \centering
  \subfigure{\includegraphics[width=.325\columnwidth]{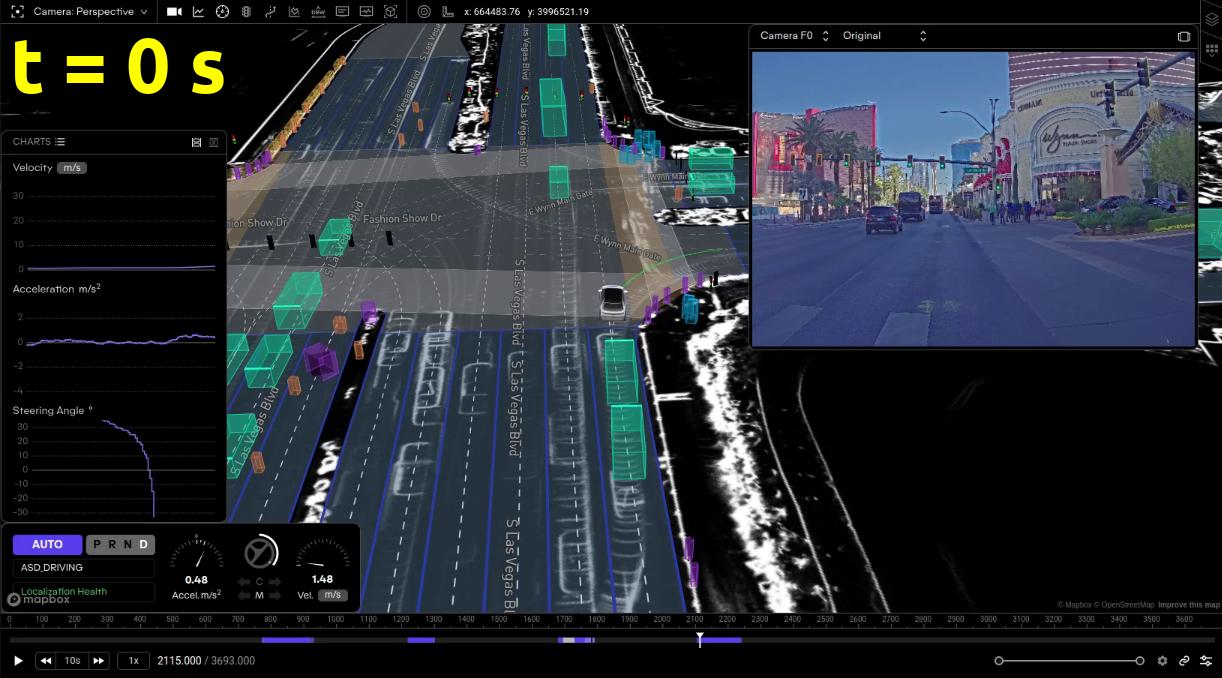}}
  \subfigure{\includegraphics[width=.325\columnwidth]{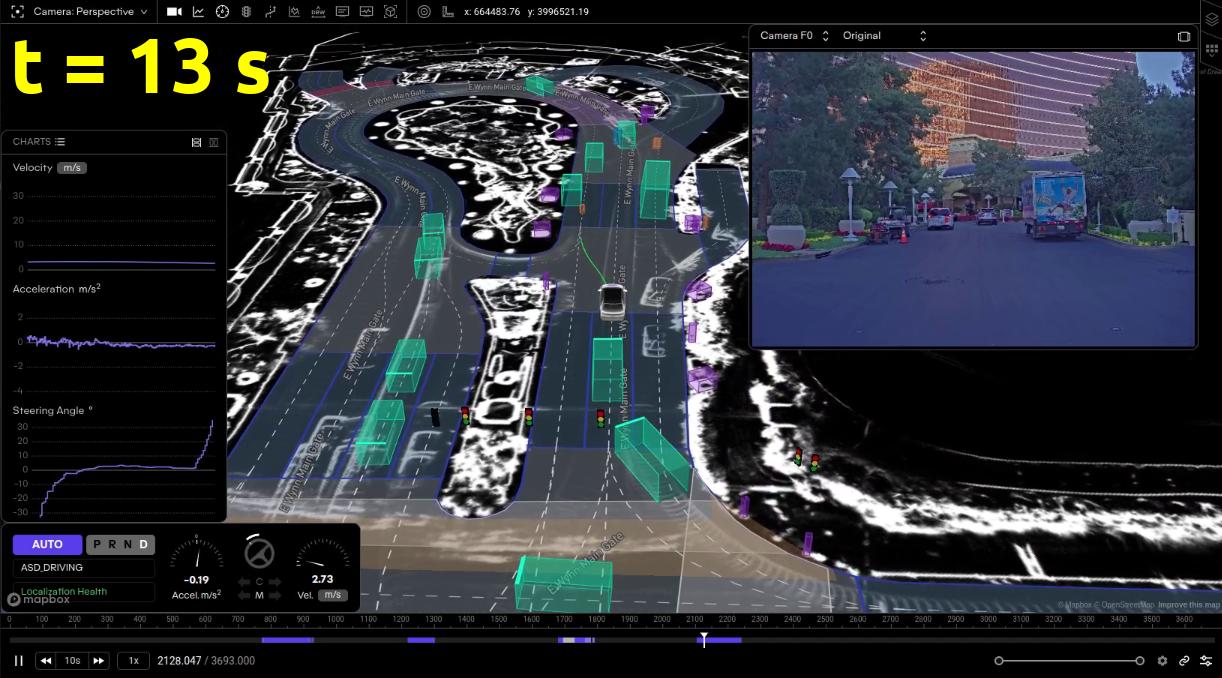}}
  \subfigure{\includegraphics[width=.325\columnwidth]{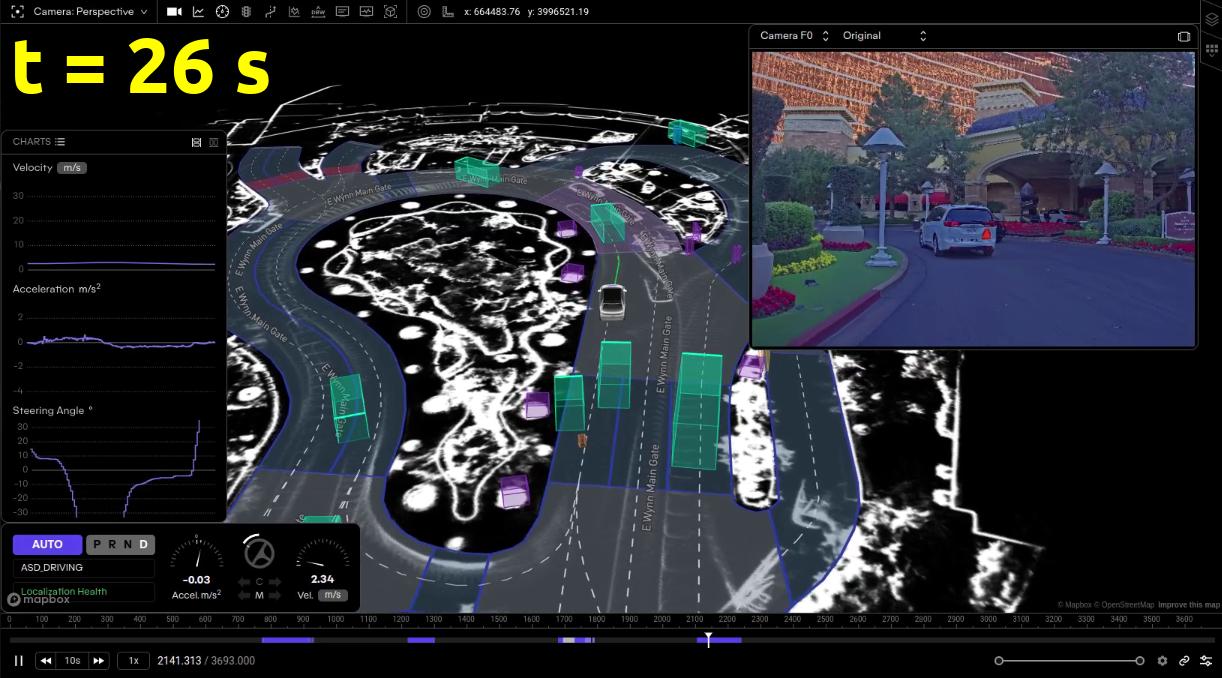}}
  \caption{The ego vehicle makes an unprotected right turn to reach a passenger pickup/dropoff zone. Towards the end of the maneuver, the safety driver takes over to adjust the ego vehicle route. Video clip: pudo-02.mp4}
  \label{fig:pudo-02}
\end{figure*}

\begin{figure*}
  \centering
  \subfigure{\includegraphics[width=.325\columnwidth]{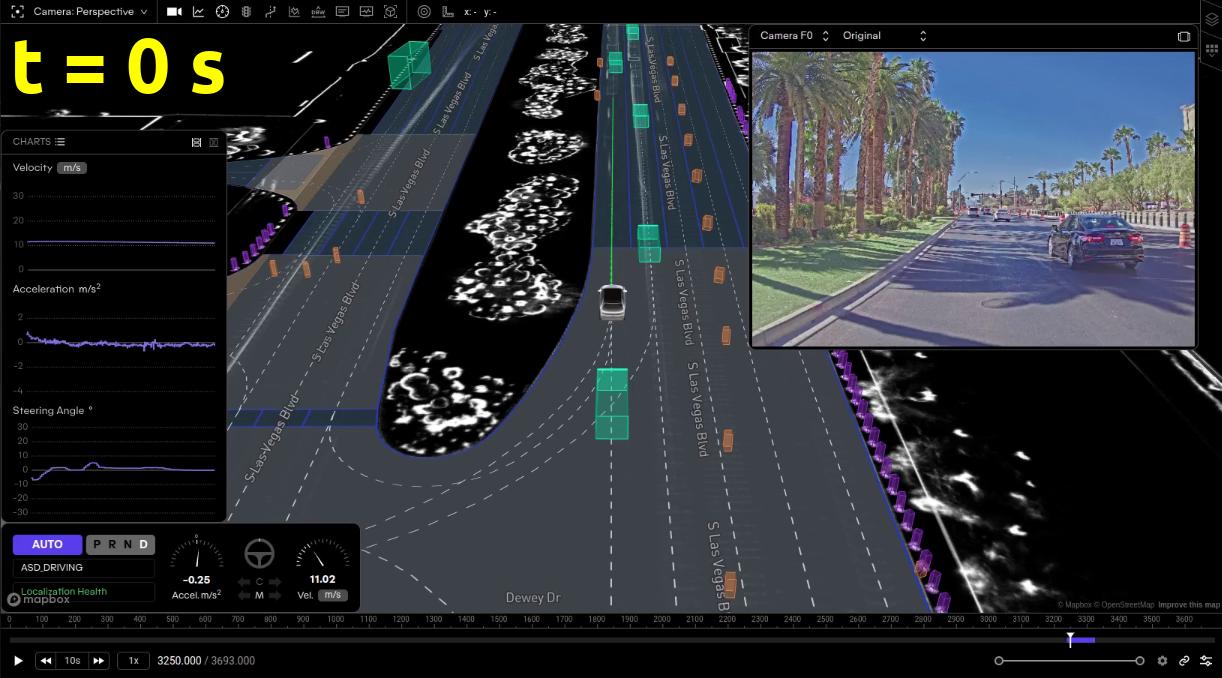}}
  \subfigure{\includegraphics[width=.325\columnwidth]{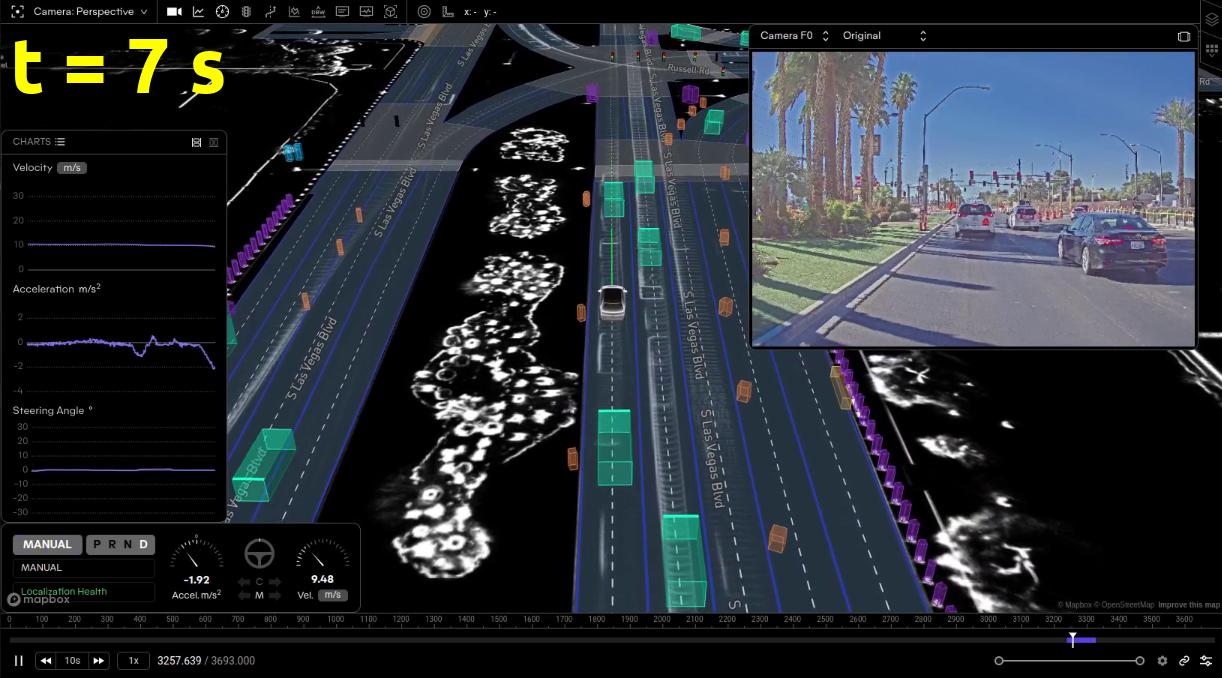}}
  \subfigure{\includegraphics[width=.325\columnwidth]{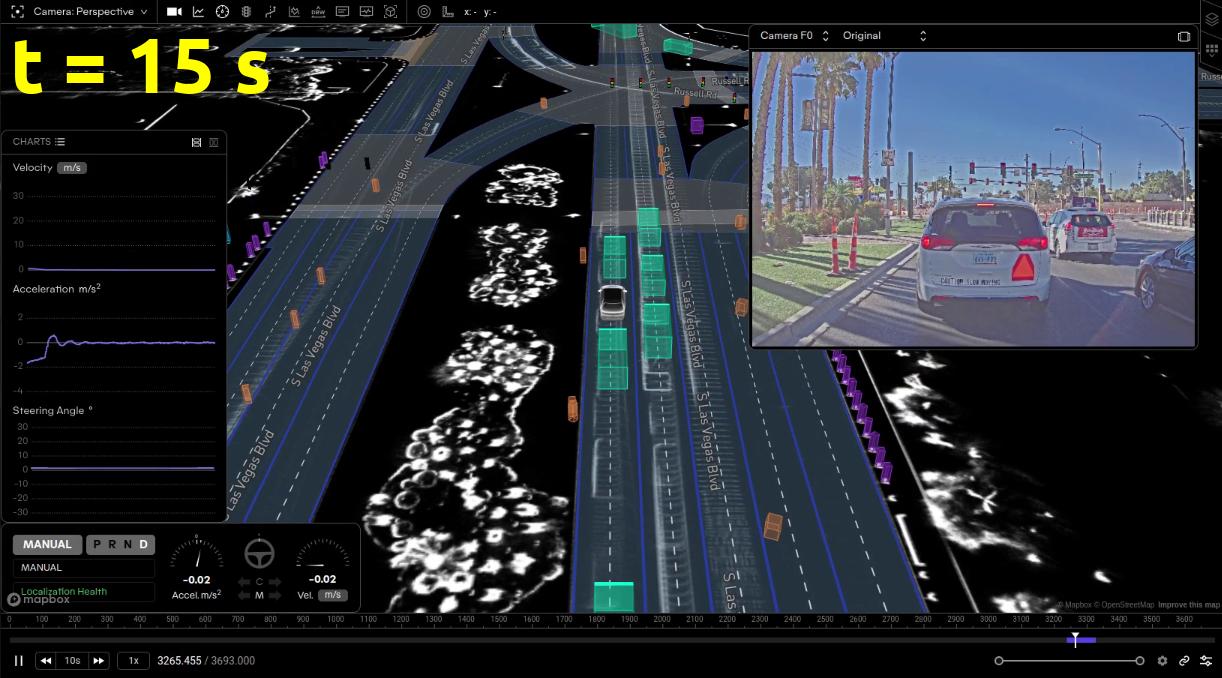}}
  \caption{The ego vehicle makes a hard stop from $~10.5$ m/s behind a vehicle. Towards the end of the maneuver, the safety driver takes over a comfortable stopping distance. Video clip: stopping-01.mp4}
  \label{fig:stopping-01}
\end{figure*}

\begin{figure*}
  \centering
  \subfigure{\includegraphics[width=.325\columnwidth]{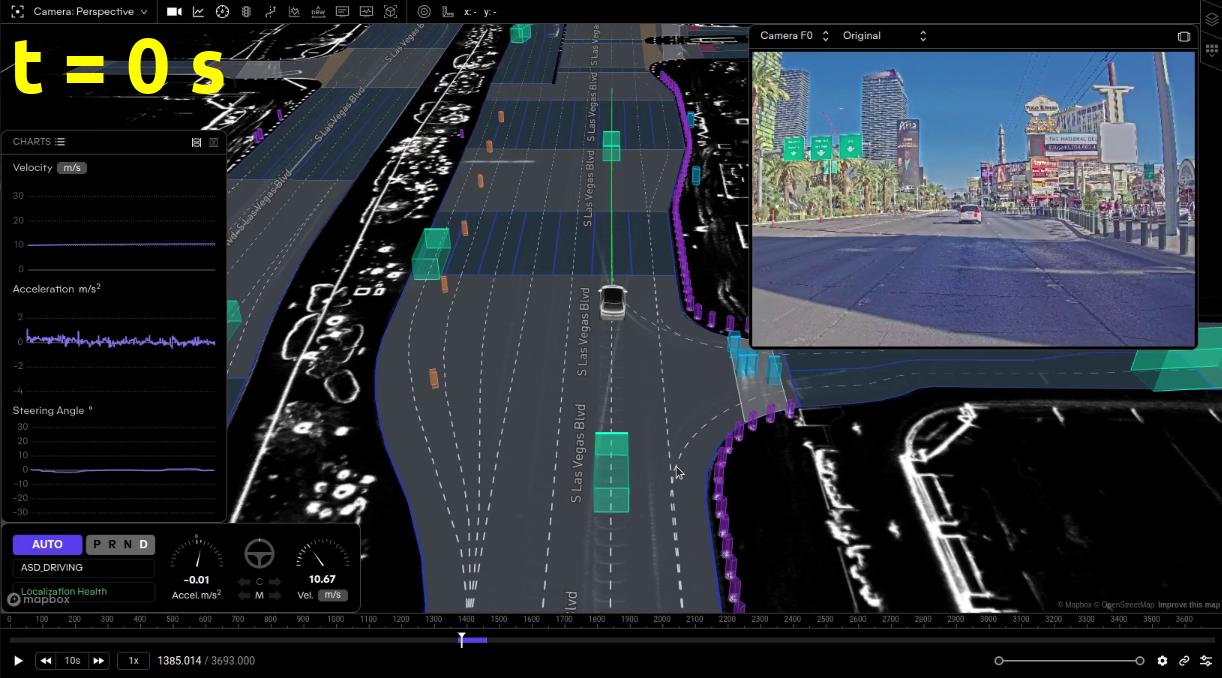}}
  \subfigure{\includegraphics[width=.325\columnwidth]{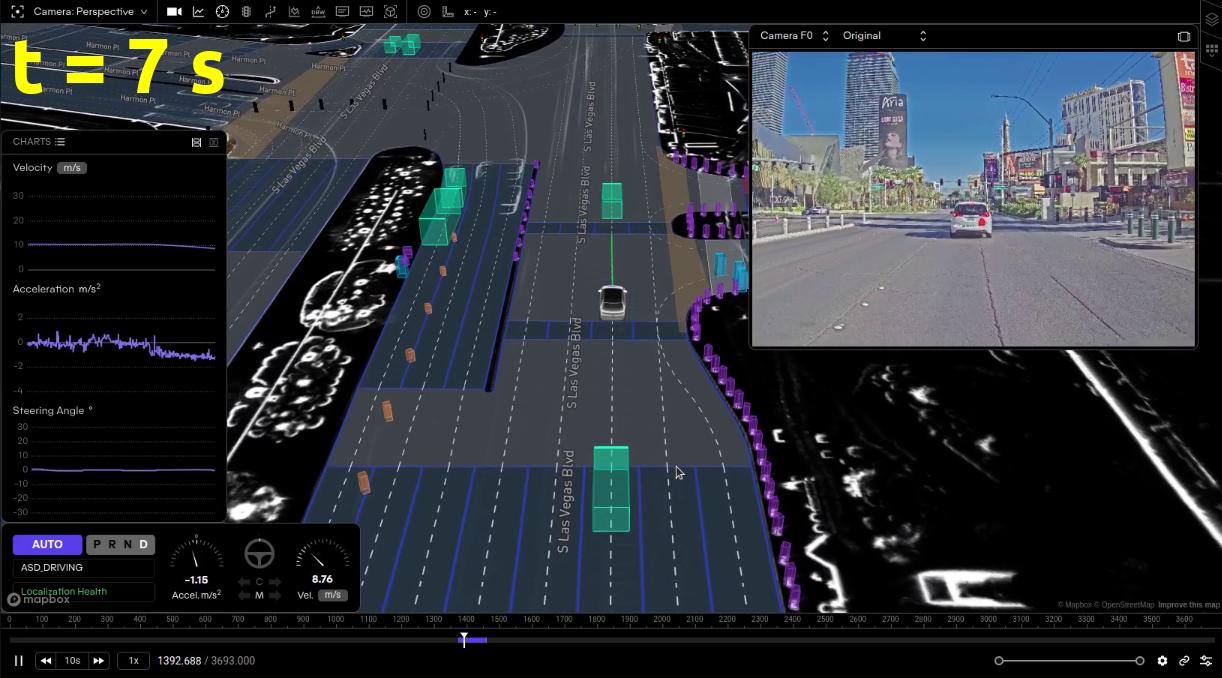}}
  \subfigure{\includegraphics[width=.325\columnwidth]{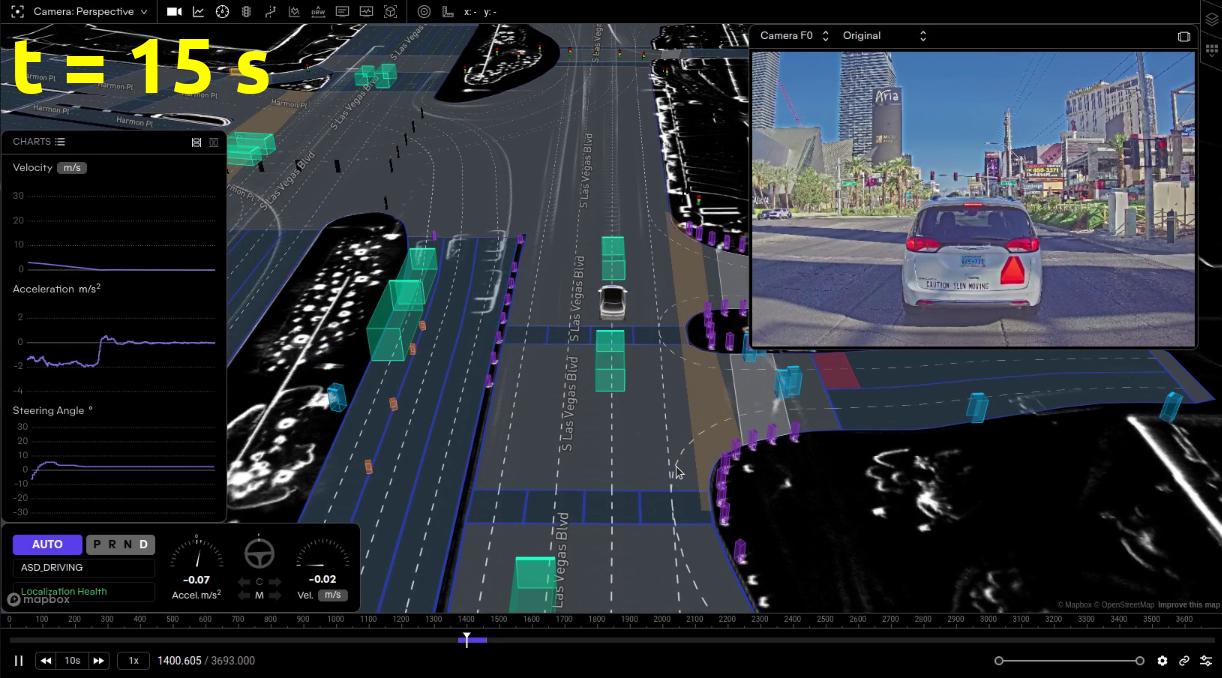}}
  \caption{The ego vehicle smoothly stops from $~10.5$ m/s behind a vehicle. Video clip: stopping-03.mp4}
  \label{fig:stopping-03}
\end{figure*}

\begin{figure*}
  \centering
  \subfigure{\includegraphics[width=.325\columnwidth]{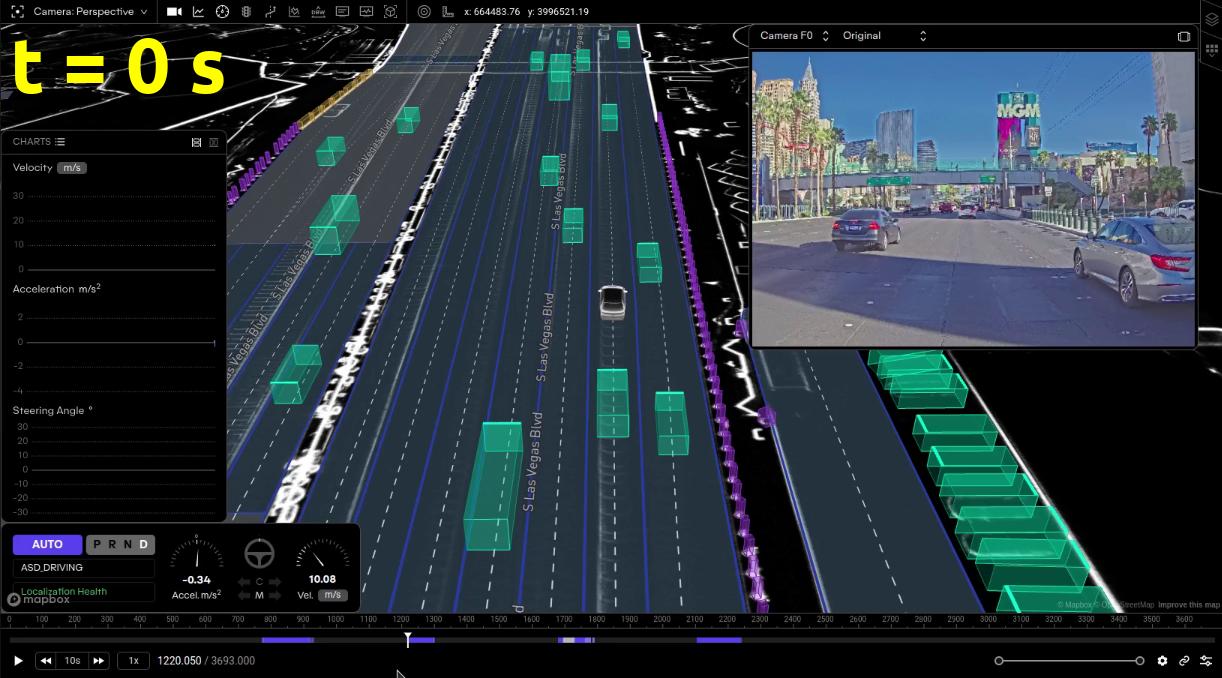}}
  \subfigure{\includegraphics[width=.325\columnwidth]{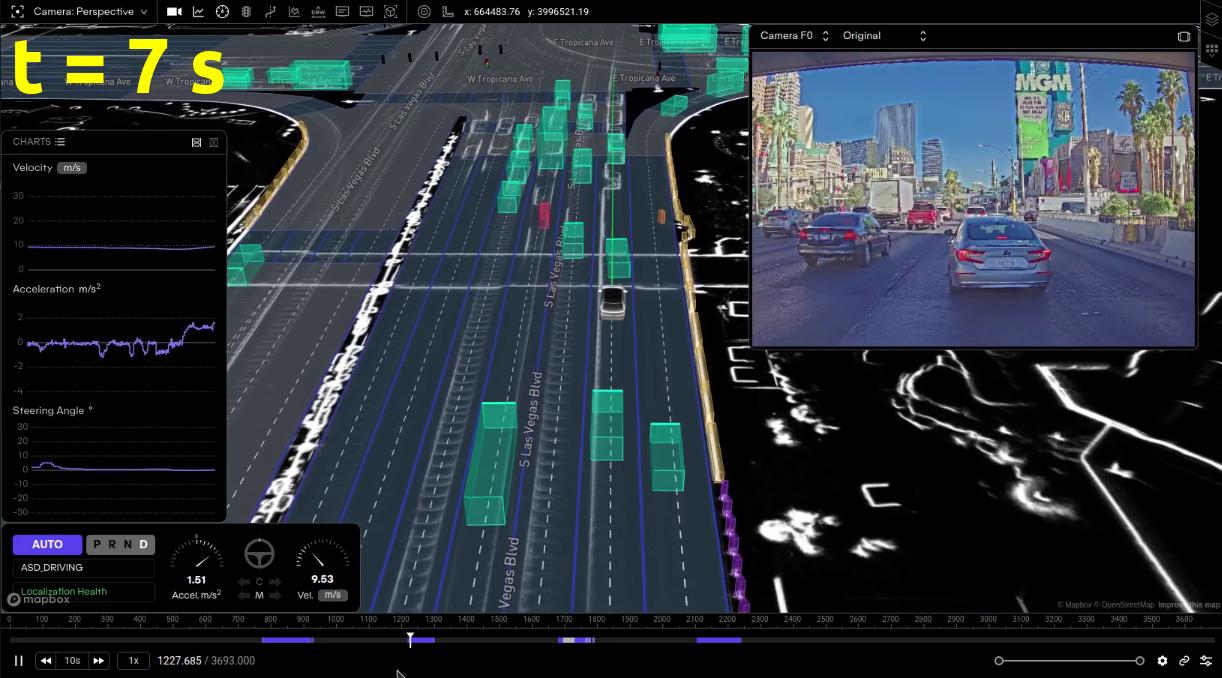}}
  \subfigure{\includegraphics[width=.325\columnwidth]{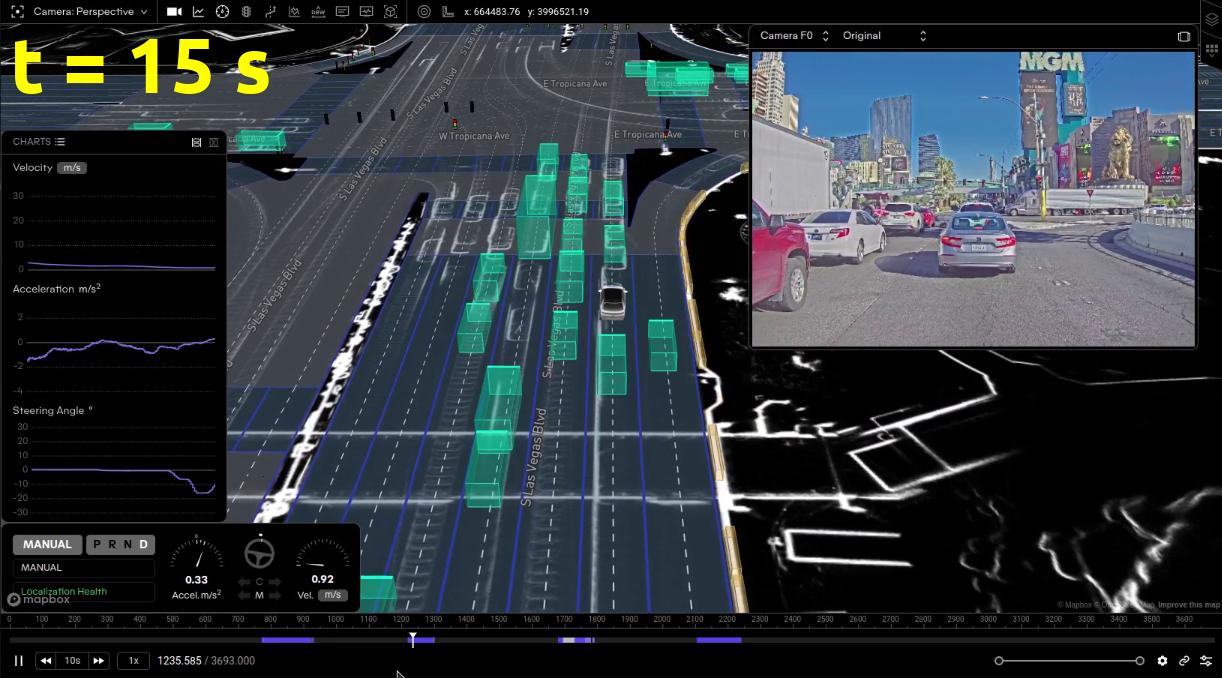}}
  \caption{Edge case where our planner does not handle properly a sharp cut-in and the safety driver has to take over. Video clip: cut-in-01.mp4}
  \label{fig:cut-in-01}
\end{figure*}

\begin{figure*}
  \centering
  \subfigure{\includegraphics[width=.325\columnwidth]{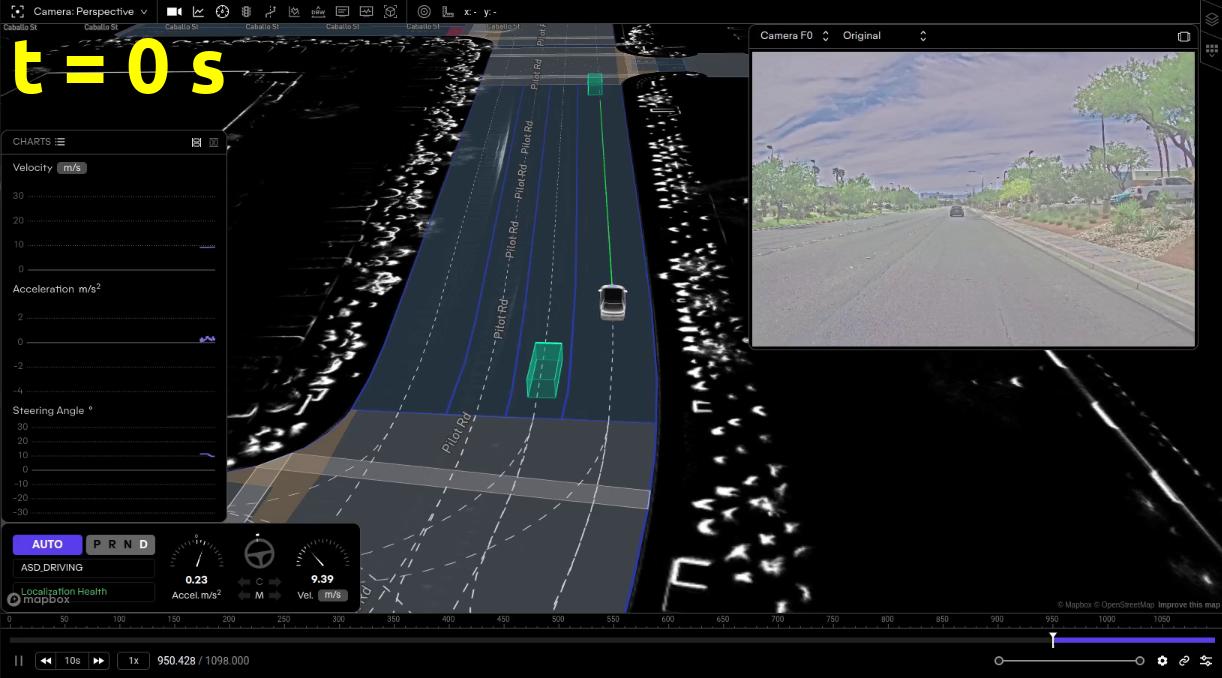}}
  \subfigure{\includegraphics[width=.325\columnwidth]{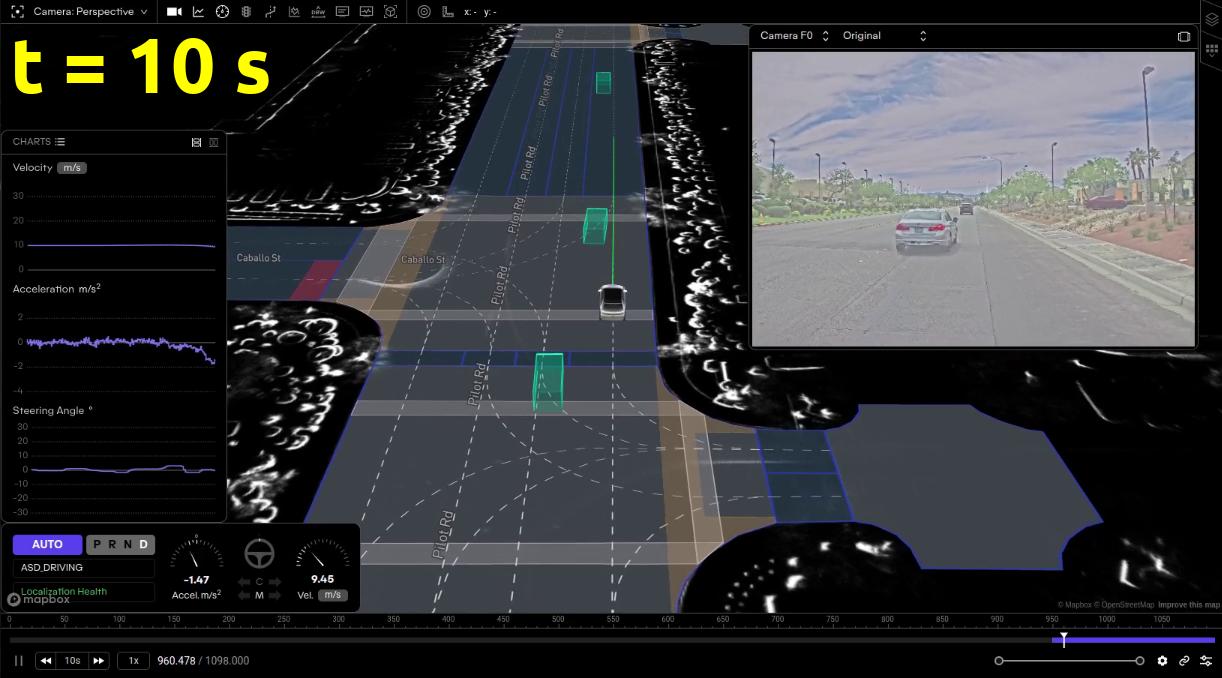}}
  \subfigure{\includegraphics[width=.325\columnwidth]{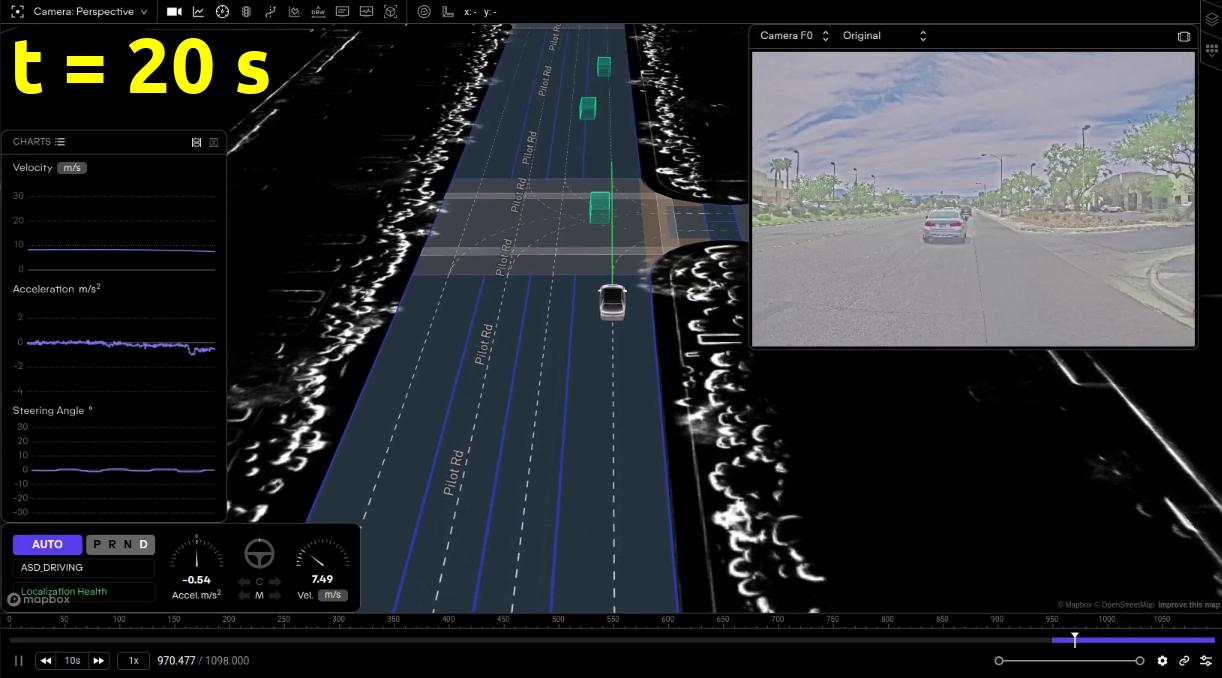}}
  \caption{The ego vehicle reacts properly to a moderate cut-in manuever at $~10.5$ m/s. Video clip: cut-in-02.mp4}
  \label{fig:cut-in-02}
\end{figure*}

\begin{figure*}
  \centering
  \subfigure{\includegraphics[width=.325\columnwidth]{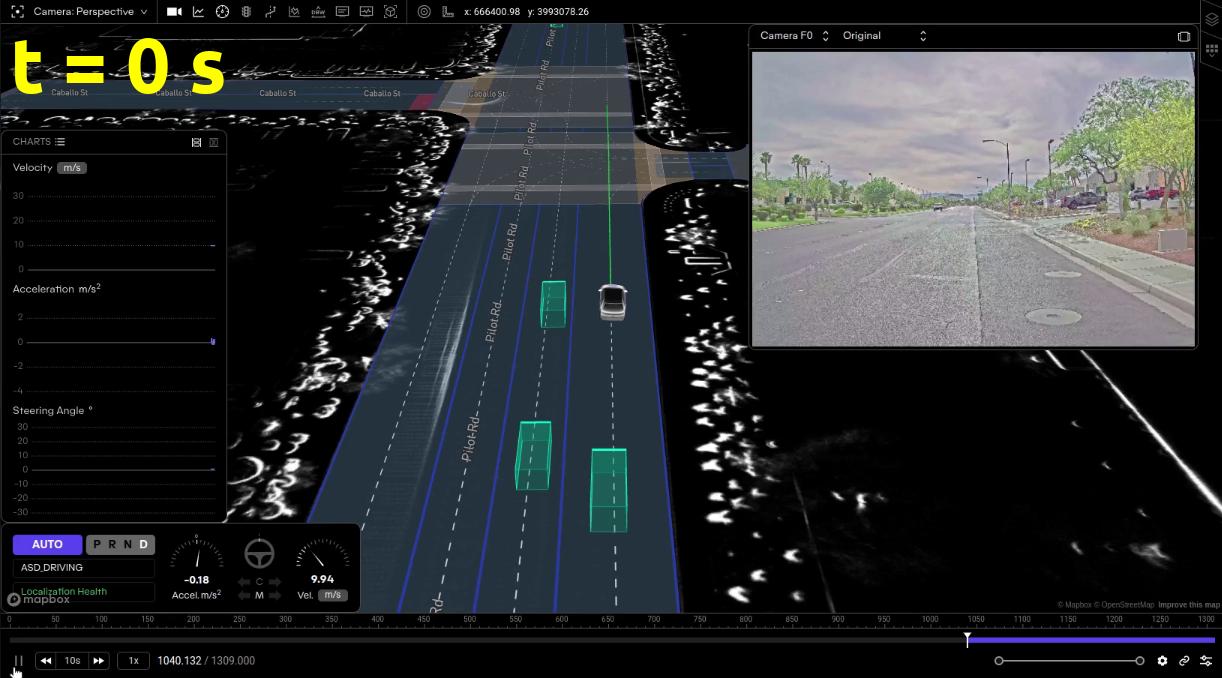}}
  \subfigure{\includegraphics[width=.325\columnwidth]{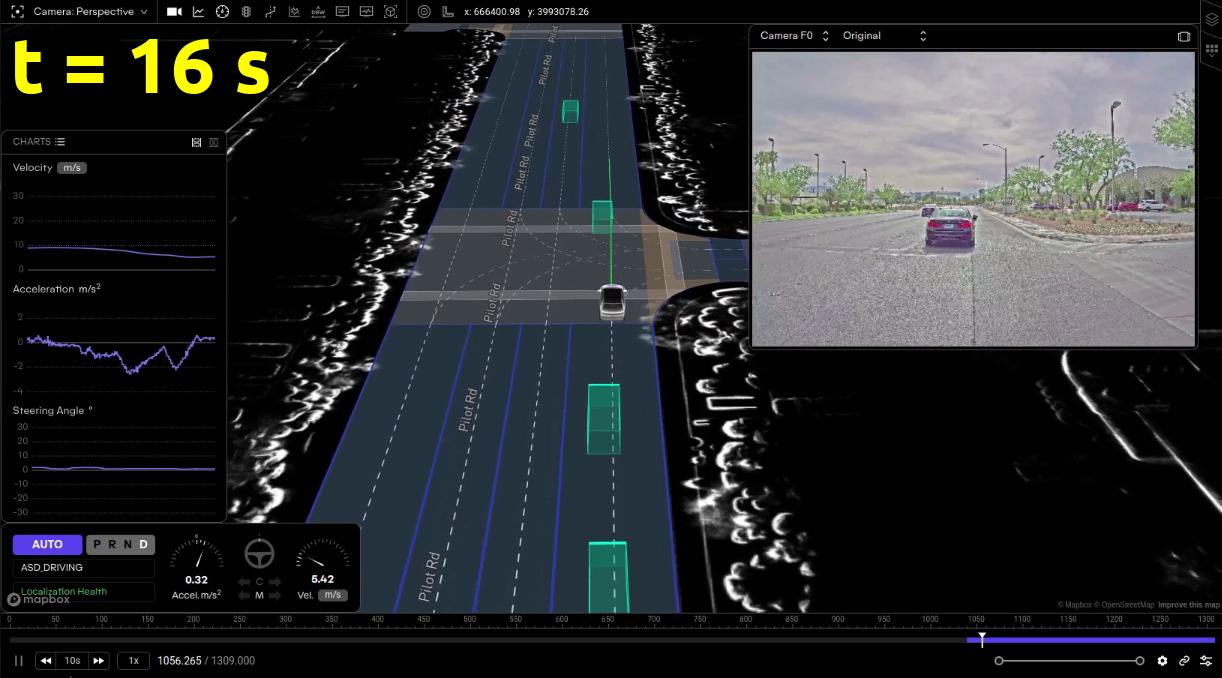}}
  \subfigure{\includegraphics[width=.325\columnwidth]{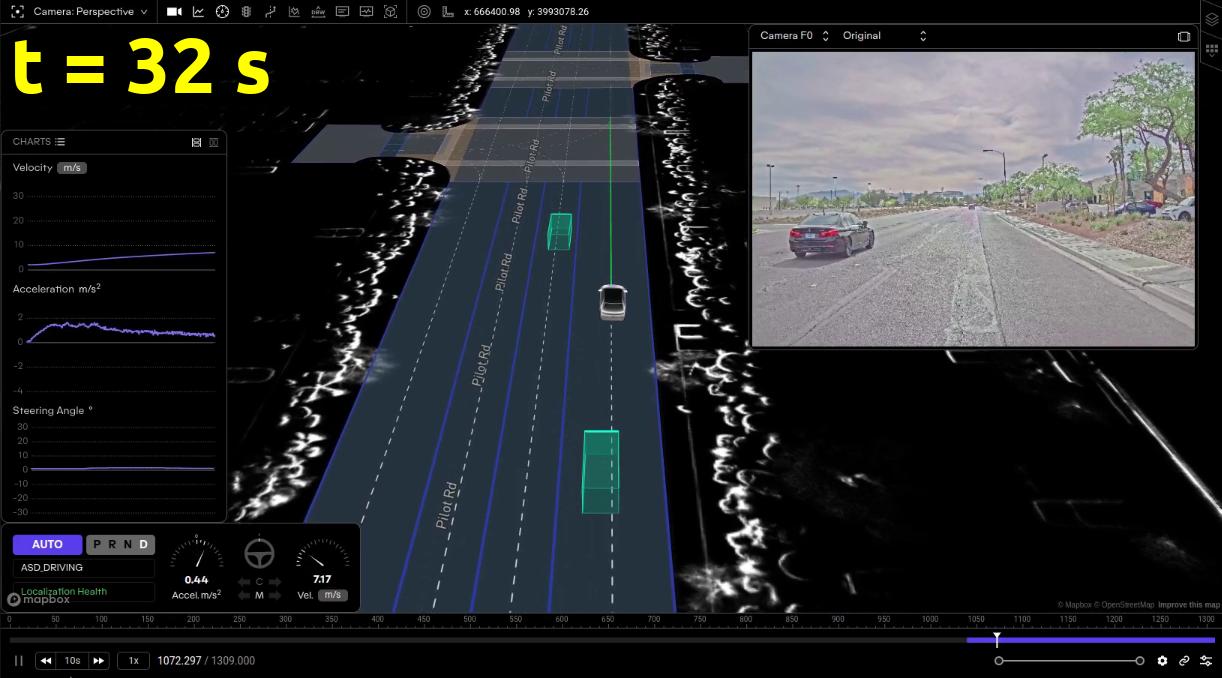}}
  \caption{The ego vehicle reacts properly to a sharp cut-in manuever at $~10.5$ m/s. Video clip: cut-in-03.mp4}
  \label{fig:cut-in-03}
\end{figure*}

\end{document}